\documentclass{article}

\usepackage{microtype}
\usepackage{graphicx}
\usepackage{subfigure}
\usepackage{booktabs} 

\usepackage{hyperref}

\usepackage[accepted]{icml2025}

\usepackage{amsmath}
\usepackage{amssymb}
\usepackage{mathtools}
\usepackage{amsthm}

\usepackage[capitalize,noabbrev]{cleveref}

\theoremstyle{plain}

\theoremstyle{definition}

\theoremstyle{remark}

\usepackage{amsmath} 
\usepackage{array}
\usepackage{tabularx}
\usepackage{multirow}
\usepackage[most]{tcolorbox}
\usepackage{makecell}
\usepackage{colortbl}
\usepackage{balance}
\usepackage{float}
\usepackage{graphicx}
\usepackage{subcaption}
\usepackage{diagbox}

\newcommand{\Rmnum}[1]{\expandafter\@slowromancap\romannumeral #1@}

\DeclareMathAlphabet\mathbfcal{OMS}{cmsy}{b}{n}

\newcommand{\eat}[1]{}
\ifodd 1

\else

\fi

\usepackage{pifont}%

\newenvironment{takeaway}[1][]
  {
 \begin{tcolorbox}
 [%
    boxrule=0.5pt,
    arc=4pt,
    left=2pt,
    right=2pt,
    bottom=2pt,
    top=2pt,
    rounded corners
    ]{}
  \textbf{#1.}
  \small \itshape}
  {
\end{tcolorbox}
}

\usepackage[textsize=tiny]{todonotes}

\icmltitlerunning{Bag of Tricks for  Inference-time Computation of LLM Reasoning }

\begin{document}

\twocolumn[
\icmltitle{Bag of Tricks for  Inference-time Computation of LLM Reasoning}





    \begin{icmlauthorlist}
\icmlauthor{Fan Liu}{to1}
\icmlauthor{Wenshuo Chao}{to1}
\icmlauthor{Naiqiang Tan}{to3}
\icmlauthor{Hao Liu}{to1}
\end{icmlauthorlist}

\icmlaffiliation{to1}{Hong Kong University of Science and Technology (Guangzhou);}
\icmlaffiliation{to3}{Didichuxing Co. Ltd;}

\icmlcorrespondingauthor{Hao Liu}{liuh@ust.hk}

\icmlkeywords{Machine Learning, ICML}

\vskip 0.3in
]



\printAffiliationsAndNotice{} 
\begin{abstract}
With the advancement of large language models (LLMs), solving complex reasoning tasks has gained increasing attention. Inference-time computation methods (e.g., Best-of-N, beam search, et al.) are particularly valuable as they can enhance reasoning performance without modifying model parameters or requiring additional training. However, these techniques come with implementation challenges, and most existing methods remain at the proof-of-concept stage with limited practical adoption due to their computational complexity and varying effectiveness across different tasks. In this paper, we investigate and benchmark diverse inference-time computation strategies across reasoning tasks of varying complexity. Since most current methods rely on a proposer-verifier pipeline that first generates candidate solutions (e.g., reasoning solutions) and then selects the best one based on reward signals (e.g., RLHF rewards, process rewards), our research focuses on optimizing both candidate solution generation (e.g., instructing prompts, hyperparameters such as temperature and top-p) and reward mechanisms (e.g., self-evaluation, reward types). Through extensive experiments (more than 20,000 A100-80G GPU hours with over 1,000 experiments) across a variety of models (e.g., Llama, Qwen, and Mistral families) of various sizes, our ablation studies reveal that previously overlooked strategies can significantly enhance performance (e.g., tuning temperature can improve reasoning task performance by up to 5\%). Furthermore, we establish a standardized benchmark for inference-time computation by systematically evaluating six representative methods across eight reasoning tasks. These findings provide a stronger foundation for future research. The code is available at \url{https://github.com/usail-hkust/benchmark_inference_time_computation_LLM}

\begin{table*}[t]
\centering
\caption{Configuration of inference-time computation methods.
Inference-time computation involves two main steps: generating candidate solutions (e.g., chain-of-thought reasoning) and selecting the optimal solution. These configurations, though significant, often receive less attention and lack standardization.}~\label{tab:config_llm_reasoning}
\vspace{-0.05in}
\resizebox{\textwidth}{!}{
\begin{tabular}{l l c c c c c c l}
\hline
\textbf{Methods} & \textbf{Domain} & \textbf{*N} & \textbf{Prompt} & \textbf{Temperature} & \textbf{Top-p} & \textbf{Trajectory Selection} & \textbf{Verification} & \textbf{Reward Model} \\
\hline
Q*~\cite{wang2024q} & Math/Code & 6 & COT & 0.9/0.2 & * & Best-of-N & Step-level reward (score value) & Policy model  \\
MALT~\cite{motwani2024malt} & Math/CommonsenseQA & 27 & COT & 0.3 & * & Best-of-N & Trajectory-level reward & Verify agent  \\
GenRM~\cite{zhang2024generative} & Math/Algorithmic & 1-32 & COT & * & * & Best-of-N & Step-level reward (yes or no) & Verifier  \\
HiAR-ICL~\cite{wu2024beyond} & Math/CommonsenseQA & 5 & Automatic COT & 0.8 & 0.9 & MCTS & Step-level reward & Self-Verify \\
CPO~\cite{zhang2024chain} & Math/CommonsenseQA/Fact Verification & 5 & COT & 0.9/0.4 & 0.9 & Tree of Search & Step-level reward (score value) & Self-Verify \\
AutoMathCritique~\cite{xi2024enhancing} & Math & 1-128 & COT & 0.7 & * & Step-level refine & Step-level reward ( correct or wrong) & Critique model \\
\hline
\end{tabular}
}
\vspace{-0.00in}
\end{table*}

\end{abstract}

\section{Introduction}~\label{sec:intro}

Large language models (LLMs) have demonstrated remarkable reasoning capabilities, enabling them to tackle increasingly sophisticated tasks in fields such as science, mathematics, and coding~\cite{zhang2024generative, chen2021codex_humaneval}. While scaling model size and expanding high-quality training datasets have significantly driven these advancements, researchers are actively exploring complementary approaches to further enhance model performance. Inspired by human problem-solving behavior—where individuals often dedicate more time deliberating on complex problems to improve their decisions—there is growing interest~\cite{snell2024scaling} in leveraging \textit{inference-time computation} (e.g., utilizing additional computation during testing to enhance the performance of reasoning tasks) to strengthen the reasoning abilities of LLMs.

While inference-time computation holds significant potential for enhancing the reasoning performance of LLMs~\cite{wang2022self_consistency}, existing studies reveal mixed results in inference-time computation (e.g., limited self-correction capabilities~\cite{huang2023large}).  Its effectiveness on broader reasoning tasks (e.g., logical reasoning, code generation, question answering, and fact verification) remains limited, with most research narrowly focused on domains like math problems. Moreover, inference-time methods are sensitive to hyperparameters, such as temperature and top-p sampling, where small adjustments can lead to notable performance differences (e.g., a 5\% improvement in solving math problems by tuning temperature). These challenges underscore the critical role of inference-time techniques (e.g., instructing prompt, sampling strategies, reward models), as shown in Table 1. Despite recent advancements, these gaps indicate that the field remains nascent, with many challenges yet to be addressed.

In this study, we investigate key tricks that influence the effectiveness of inference-time computation methods in LLM reasoning.  Since most current methods rely on a proposer-verifier  pipeline that first generates candidate solutions (e.g., chain-of-thought candidate solutions) and then selects the optimal solution based on specific reward signals (e.g., RLHF rewards, process rewards), our research focuses on strategies for both candidate solution generation (e.g., instructing prompts, hyperparameters such as temperature and top-p) and reward mechanisms (e.g., self-evaluation, reward types) across broader reasoning tasks, including logical reasoning, code generation, fact verification, complex mathematics, and arithmetic. Through ablation studies, we evaluate simple techniques in inference-time computation and empirically analyze their impact on reasoning performance. Our analysis reveals that inference-time computation methods are highly sensitive to the experimental setup. With proper configurations, our results demonstrate that previously overlooked techniques can significantly enhance performance. Furthermore, we emphasize the need for a standardized benchmarking framework for inference-time computation in LLM reasoning.

Our main contributions and observations include the following: 1) \textbf{Evaluation of Key Tricks}: We evaluate the impact of a wide range of key tricks (e.g., prompt type, temperature, top-p, reward model type) on LLMs. Key insights include: Instruction prompts significantly influence LLM reasoning, with self-correction often yielding mixed results compared to  CoT prompting;
Self-evaluation frequently fails to assess solution quality effectively;
Reward models can cause performance inflation in LLM reasoning, with inconsistent effectiveness across tasks due to generalization issues;
2) \textbf{Combination of Techniques}: We further explore the effects of combining selected useful tricks on inference-time computation methods. Our empirical results suggest that improvements are not always additive when combining different techniques, although methods such as prompt design, temperature, and reward models can be effective, as demonstrated in Section~\ref{sec:combination_trick}.
3) \textbf{Comprehensive Benchmarks}: We conduct extensive experiments, evaluating six representative inference-time computation methods across eight reasoning tasks, utilizing over 20,000 A100-80G GPU hours and more than 1,000 individual tests. Our results establish a baseline benchmark for inference-time computation, as shown in Table~\ref{tab:main_result}.

\section{Related Work}~\label{sec:relected_work}
We briefly introduce related work, including reasoning with LLMs, inference-time computation methods for LLM reasoning, and benchmarks of LLM reasoning.

\textbf{Reasoning with LLMs.}
LLMs have demonstrated strong reasoning abilities in complex tasks such as code generation, mathematical problem-solving, and research ideation~\cite{zhou2022least}. Existing methods for enhancing LLM reasoning include: 1) Prompt engineering – Activates latent multi-step reasoning capabilities. For example, Chain of Thought (CoT)~\cite{wei2022chain_cot} guides step-by-step problem-solving but relies heavily on high-quality demonstrations for analogical learning. 2) Post-training techniques\cite{chen2024bootstrapping, chen2024alphamath} – Iteratively enrich training datasets to improve model performance. Self-training methods\cite{chen2024bootstrapping} curate new high-quality examples to enhance reasoning, but these approaches demand significant computational resources. 3)Search-based methods\cite{browne2012survey, feng2023alphazero, liu2023making} – Optimize reasoning paths at inference time using search algorithms. For instance, Tree of Thought\cite{yao2024tree} employs breadth-first search to refine solutions.
This work focuses on test-time computation, leveraging inference-time optimization to enhance LLM reasoning without additional training overhead.

\textbf{Inference-Time Computation of LLM Reasoning.} Scaling inference-time computation has proven more effective than merely increasing model parameters~\cite{snell2024scaling}. Recently, research has focused on optimizing reasoning efficiency during inference rather than solely scaling training-time computation. Best-of-N~\cite{cobbe2021training_best_of_n} enhances LLM reasoning by sampling N candidate solutions, evaluating them with a learned verifier or reward model, and selecting the highest-scoring one. Similarly, MCTS~\cite{tian2024toward} improves inference by actively planning and selecting higher-quality responses. These advancements highlight inference-time optimization as crucial for enhancing LLM reasoning beyond scaling training computation.

\textbf{Benchmarks of LLM Reasoning.}  LLMs have made remarkable progress in solving complex tasks in a zero-shot manner~\cite{hendrycks2021measuringMATH, press2022measuring_bamboogle, liu2024jailjudgecomprehensivejailbreakjudge}, positioning them as a key milestone toward artificial general intelligence. Consequently, benchmarking their reasoning abilities has become a central challenge. Recent studies have evaluated LLM reasoning across various domains, including mathematical reasoning~\cite{hendrycks2021measuringMATH}, code generation~\cite{chen2021codex_humaneval}, and factual QA~\cite{Thorne18Fever}, etc~\cite{liu2024regmix, liu2024adversarial}. While these benchmarks enhance our understanding of LLM reasoning, most research has focused on task performance rather than inference-time computation, leaving key optimization techniques underexplored.

Unique to this paper, we are the first to comprehensively study how LLM reasoning performance changes with the incorporation of previously overlooked key techniques. We hope that our work will provide valuable insights into the role of inference-time computation.

\vspace{-0.2in}
\section{Preliminares}

\textbf{LLMs}. Given an input context $\mathbf{x}$ (e.g., math problem, commonsense QA, etc.), the LLM aims to autoregressively predict the next token~\cite{dubey2024llama},
\begin{equation}
    \pi_{\theta}(\mathbf{y}| \mathbf{x}) = \prod_{t=1}^{n}\pi_{\theta}(\mathbf{y}_{t}|\mathbf{x},y_ {< t} ),
\end{equation}
where $ \pi_{\theta}(\cdot)$ is the LLM parameterized by $\theta$, and  $\mathbf{y} =( y_1, y_2,\cdots , y_n )$ is the output sequence. Here,  $y_{<1} = \emptyset $ and  $y_{<t} = ( y_1, y_2,\cdots , y_{t-1} )$. For a vocabulary size  $M$, the probability of predicting the $t$-th token is determined using a softmax with temperature $\tau$ on logit scores $z$ of all tokens, combined with top-p (nucleus sampling) to control the randomness and diversity of the sampling process.

\textbf{Chain of Thought Prompting.} Chain-of-thought (CoT)~\cite{wei2022chain_cot}  is a method that prompts LLMs to generate a series of reasoning steps leading to the final answer. These intermediate steps, denoted as $ y_1, \dots, y_{n-1} $, connect the input $\mathbf{x}$ to the output $y$ (omit $n$ for simplicity), where $n$ represents the total number of steps. For example, given an instruction $\mathbf{I}$ (e.g., "Let's solve this step by step") along with demonstration examples and the input question $\mathbf{x}$, the final answer is $y$. Each intermediate thought $y_i$ is a part of the reasoning process that leads to the final answer. These thoughts are sequentially generated from the distribution $y_i \sim \pi_{\theta}(\cdot \mid \mathbf{I}, \mathbf{x}, \mathbf{y}_{<i-1})$, and the final output is sampled from: $y \sim \pi_{\theta}(\cdot \mid \mathbf{I}, \mathbf{x}, \mathbf{y}_{<n-1})$.

\textbf{Temperature.} The temperature~\cite{hinton2015distilling}  $\tau$ of LLM controls the level of randomness in the generated outputs, influencing their diversity. Instead of directly calculating the softmax, the logits are scaled by the temperature value. The conditional probability of generating a token in the sequence can be expressed as: $\pi_{\theta}(y_t \mid \mathbf{x}, \mathbf{y}_{<t}) = \frac{\exp(z_t / \tau)}{\sum_{i=1}^M \exp(z_i / \tau)},$ where $z_t$ represents the logit score: $\text{logit}_{\theta}(y_t \mid \mathbf{x}, \mathbf{y}_{<t})$, and $\tau$ is the temperature parameter. A higher temperature $\tau$ results in a smoother probability distribution (introducing more randomness), while a lower temperature makes the distribution sharper, leading to more deterministic behavior.

\textbf{Top-p.} The top-p~\cite{holtzman2019curious} controls the LLM output by augmenting the vocabulary size $M$ as only those tokens are considered for which the cumulative probability ($C_k = \sum_{i=1}^k p_{(i)}$) is greater than the Top-p value (the cumulative probability is calculated by sorting the tokens by their probability in a descending order and then adding them up,  where: probabilities: $\{p_{(1)}, p_{(2)}, \dots, p_{(M)}\}$, where:
$p_{(1)} \geq p_{(2)} \geq \dots \geq p_{(M)}$). After the tokens are selected, it would be re-calculated their softmax with reduced vocab size. The  truncated probability distribution can be defined as:
$\pi_{\theta}(y_t \mid \mathbf{x}, \mathbf{y}_{<t}) = 
\begin{cases} 
\frac{p_{(i)}}{\sum_{j=1}^k p_{(j)}}, & \text{if } i \leq k, \\
0, & \text{if } i > k.
\end{cases}$.

\begin{figure*}[htb]
    \centering
    \includegraphics[width=0.85\textwidth]{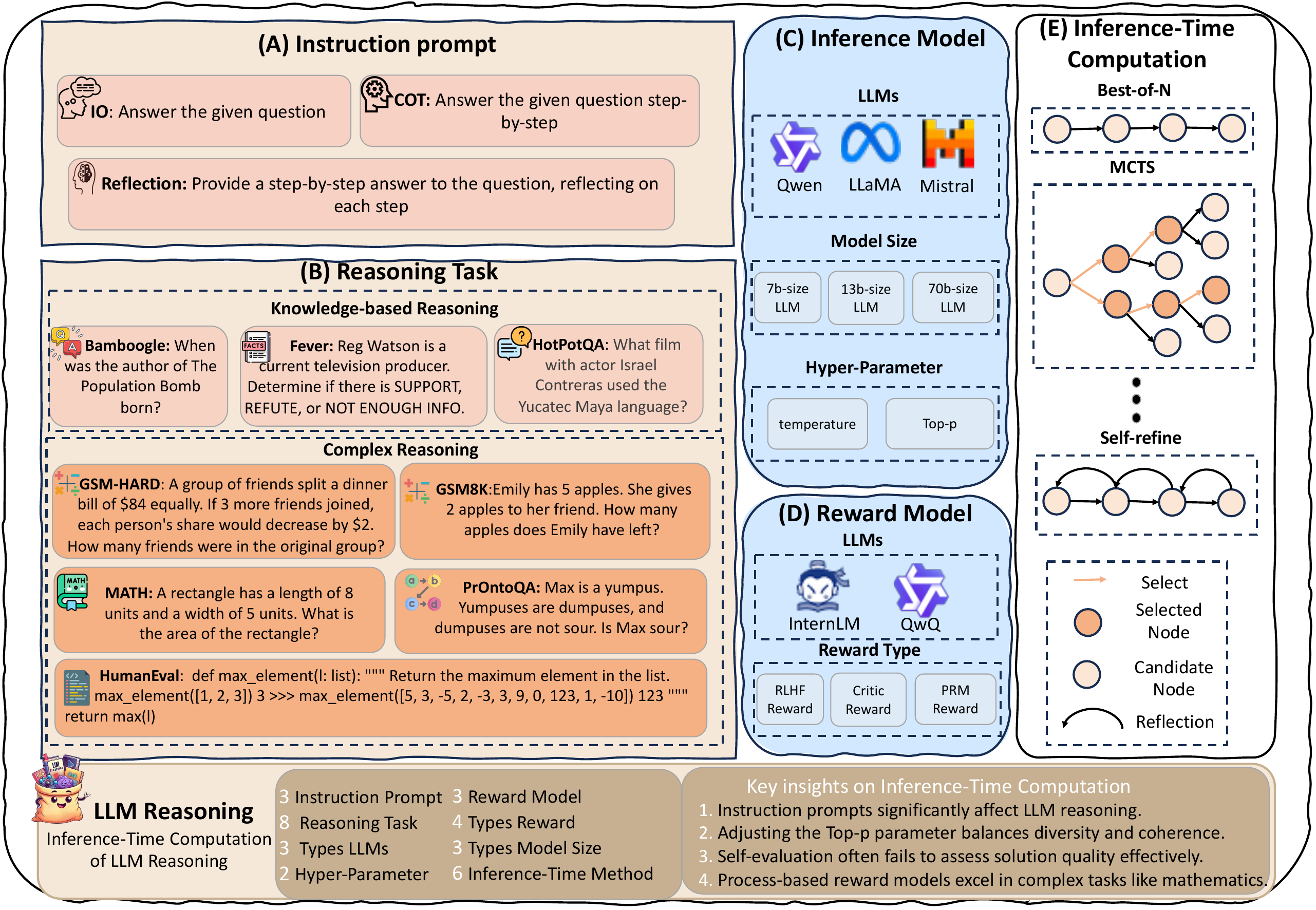}
    \vspace{-0.01in}
    \caption{Overview of Decoding Inference-Time Computation for LLM Reasoning.
\textbf{(A) Instruction Prompt}: Includes IO, Chain-of-Thought (CoT), and reflection-based CoT prompts.
\textbf{(B) Reasoning Task}: Evaluates models on eight datasets: Arithmetic (GSM8K, GSM-Hard), Complex Math (MATH), Logical (PrOntoQA), Code Generation (HumanEval), Question Answering (Bamboogle), Fact Verification (FEVER), and Common Sense (HotpotQA).
\textbf{(C) Inference Model}: Analyzes LLMs (LLaMA, Qwen, Mistral) of varying sizes and architectures, with performance assessed via temperature and top-p hyperparameters.
\textbf{(D) Reward Model}: Explores reward types like RLHF, critic, and process-based models to enhance inference performance.
\textbf{(E) Inference-Time Computation}: Investigates methods like Best-of-N Sampling, Step-Level Best-of-N, Self-Consistency, Monte Carlo Tree Search (MCTS), and Self-Refinement to optimize reasoning.}
    \label{fig:main_framework}
    \vspace{-0.2in}
\end{figure*}

\textbf{Inference-Time Computation Methods for LLM Reasoning.}
Inference-time computation methods~\cite{ott2018analyzing_beam_search} typically follows a pipeline comprising two main steps: generating candidate solutions (e.g., chain-of-thought reasoning candidates) and selecting the optimal solution based on specific reward signals (e.g., numerical reward, self-consistency, process reward, or binary feedback such as "Yes" or "No"). Formally, given a problem $\mathbf{x}$, the inference-time computation methods  sample $K$ candidate solutions:
$y^{(k)} \sim \pi_{\theta}(y \mid \mathbf{I}, \mathbf{x}, \mathbf{y}_{<n}),
 $for $k = 1, 2, \dots, K$, where $y^{(k)}$ represents the $k$-th candidate solution. After sampling, each candidate is evaluated using a reward model to produce a reward signal:$ \mathbf{r}^{(k)} = \text{reward}(\mathbf{I}, \mathbf{x}, \mathbf{y}_{<n-1}, y^{(k)}),$ where the reward model can take various forms. For example, it may be a general LLM that evaluates solutions using instructions $\mathbf{I}$ (e.g., "Let's verify the step-by-step reasoning. Is the answer correct (Yes/No)?"). Alternatively, the reward model could be specifically trained to output a scalar value between 0 and 1, with higher values indicating better solutions. The final solution $\hat{y}$ is then selected based on the reward signals. For numerical rewards, the solution with the highest reward is chosen:$ \hat{y} = \arg\max_{y_k} r_k.$

\vspace{-0.2in}
\section{Decoding Inference-Time Computation of LLM Reasoning}
\vspace{-0.1in}
In this section, we decode the inference-time computation of LLM reasoning. First, we introduce the experimental setup, followed by the main bag of tricks for improving inference-time computation of LLM reasoning. Finally, we benchmark various inference-time computation methods. The overall framework is illustrated in Figure~\ref{fig:main_framework}.
\subsection{Experiments Setup}

\textbf{Models.} \textit{Inference Model.} In our experiments, we evaluated several widely studied LLMs of varying sizes and configurations: 1) LLaMA 3.3~\cite{dubey2024llama}: Meta AI's latest iteration in the LLaMA series, available in 8B and 70B parameters. It is known for its open-source accessibility and strong benchmark performance.
2) Qwen 2.5~\cite{yang2024qwen2}: Developed by Alibaba Cloud, this model offers 7B and 72B parameter configurations, showcasing diverse LLM architectures and training methods. 3) Mistral 7B Instruct v0.3~\cite{jiang2023mistral}: A 7B parameter model from Mistral AI, recognized for its efficiency and performance rivaling larger models.
These models exhibit diverse reasoning strengths, providing insights into the impact of different architectures and training approaches. \textit{Reward Model.} We employ four types of reward models: (1) Process Reward~\cite{processbench}: Evaluates each reasoning step step-by-step. (2) Result Reward: Assesses only the final answer's correctness. (3) RLHF Reward~\cite{cai2024internlm2}: Derived from preference samples (both human-annotated and AI-generated). (4) Proof-Critical Reward: Applied in formal mathematical proof across multiple benchmarks. The details of the setting can be seen in Appendix~\ref{sec:exp_setup}.

\textbf{Tasks.}  Our research focuses on the following reasoning tasks:
1) Arithmetic Reasoning: Evaluating models on GSM8K~\cite{cobbe2021training_gsm8k} and GSM-Hard~\cite{gao2022pal_gsm_hard} datasets to test their arithmetic calculation skills.  
2) Complex mathematical reasoning: Using the MATH~\cite{hendrycks2021measuringMATH}  to assess proficiency in solving advanced mathematical problems.  
3) Logical Reasoning: measuring logical deduction and inference abilities with the ProntoQA~\cite{  PrOntoQA} dataset.  
4) Code Generation: Testing code generation skills on the HumanEval~\cite{chen2021codex_humaneval} dataset.  
5) Question Answering: Evaluating performance in answering diverse questions using the Bamboogle~\cite{press2022measuring_bamboogle}.  
6) Fact Verification: Assessing factual verification using the FEVER~\cite{Thorne18Fever} dataset.  
7) Common Sense Reasoning: Testing understanding of common sense knowledge and reasoning with the HotpotQA~\cite{yang2018hotpotqa} dataset.  

 \textbf{Inference-Time Computation Methods.} This study examines common inference-time computation methods: 1) Best-of-N~\cite{cobbe2021training_best_of_n}: Generates multiple outputs (N samples) for a given input and selects the optimal one based on reward model. 2) Step-Level Best-of-N Sampling~\cite{cobbe2021training_best_of_n}: Applies Best-of-N sampling at each generation step, selecting the most promising thoughts. 3) Self-Consistency~\cite{wang2022self_consistency}: Produces multiple reasoning paths or answers and selects the most consistent one. 4) Beam Search~\cite{ott2018analyzing_beam_search}: Explores outputs level by level, expanding all nodes at the current depth before proceeding to the next. 5) Monte Carlo Tree Search (MCTS)~\cite{feng2023alphazero_mcts}: Uses random sampling to build a search tree and identify the most promising outputs. 6) Self-Refine~\cite{madaan2024self_refine}: Allows LLMs to iteratively refine outputs during inference.


\textbf{Evaluation Metrics.}
For most reasoning tasks, accuracy serves as the primary evaluation metric. For code generation, we use the pass@k metric~\cite{chen2021codex_humaneval}, which considers a problem solved if at least one of k generated code samples passes all test cases. In our evaluation, we focus on pass@1 by generating multiple samples and selecting the best one using the reward model.

\subsection{Bag of Tricks}

Our goal is to investigate how previously overlooked tricks can critically affect the performance of inference-time computation methods, which typically consist of two main steps: generating candidate solutions (e.g., prompt type, temperature, top-p, etc.) and selecting the optimal solution based on specific reward signals (e.g., self-evaluation, reward type, reward process). In our default setup, we primarily adopt the Best-of-N inference-time computation with the number of candidates $N = 32$, the temperature $\tau = 0.7$, and top-p set to $0.9$. Additionally, the instruction prompt type is set to Chain-of-Thought (CoT). Without further modifications, we conduct ablation studies, varying only the specific tricks under investigation. We focus primarily on complex reasoning tasks, including math problems, and code generation tasks, etc, while additional tasks are detailed in the Appendix~\ref{sec:other_exp}.   All additional details regarding the experimental implementation settings can be found in the Appendix~\ref{sec:exp_setup}.

Note that our empirical observations and conclusions may not generalize to all datasets and models. However, we emphasize the necessity of using consistent implementation details to ensure fair comparisons among different inference-time computation methods.
\begin{figure}[t]
\vspace{-0.1cm}
\centering
\includegraphics[width=0.36\textwidth]{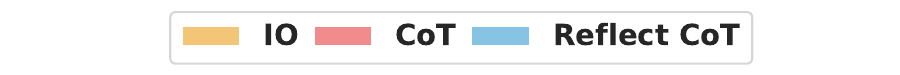}
\vspace{-0.2cm}
\subfigure{
\includegraphics[width=0.225\textwidth]{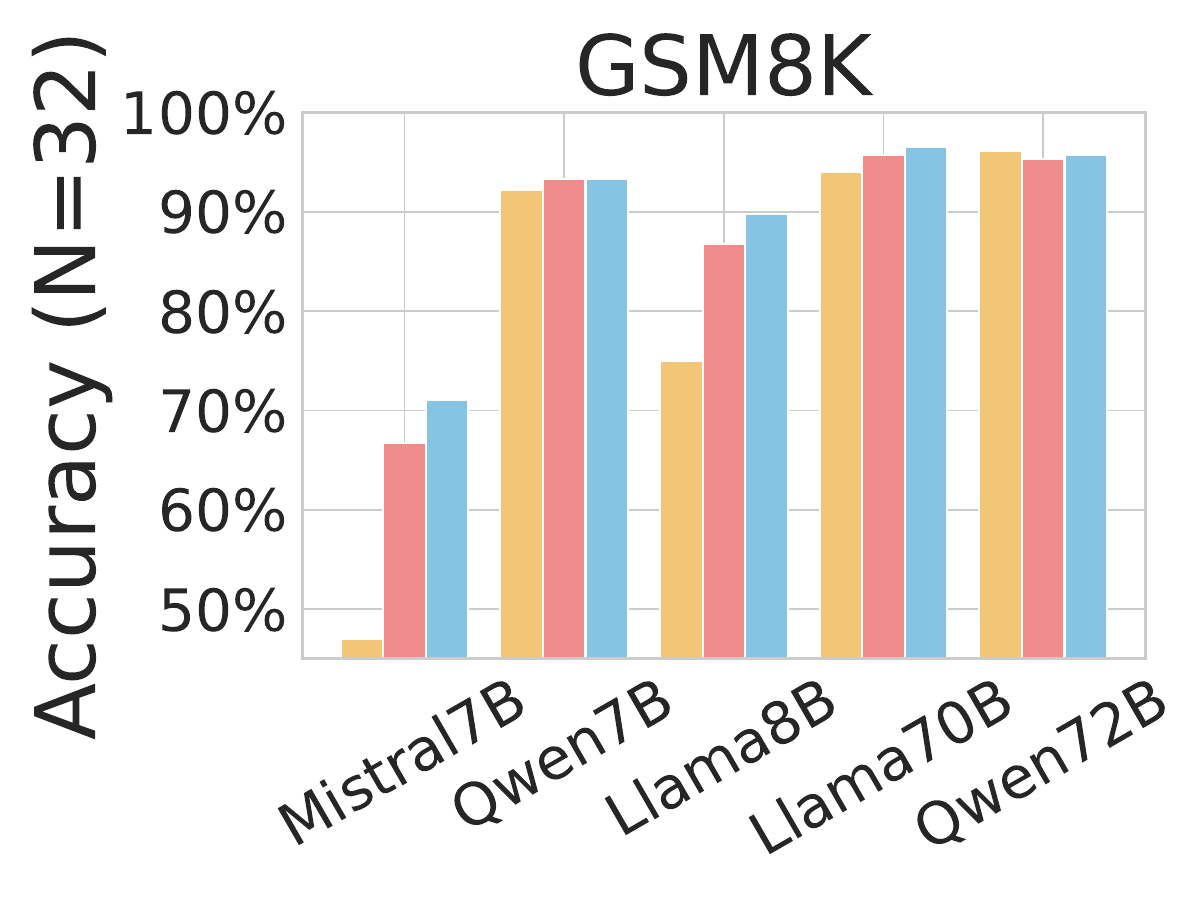}
}
\subfigure{
\includegraphics[width=0.225\textwidth]{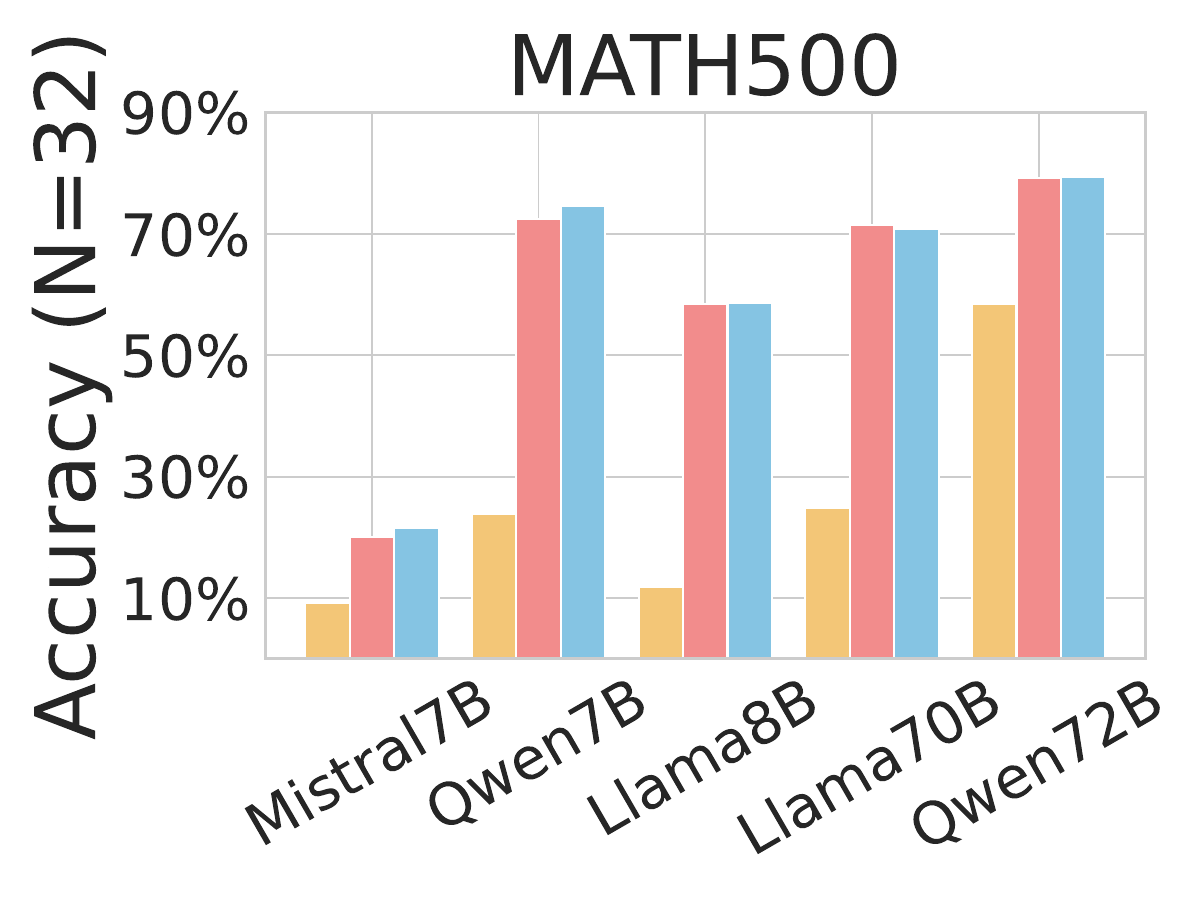}
}
\vspace{-0.2cm} \\  
\subfigure{
\includegraphics[width=0.225\textwidth]{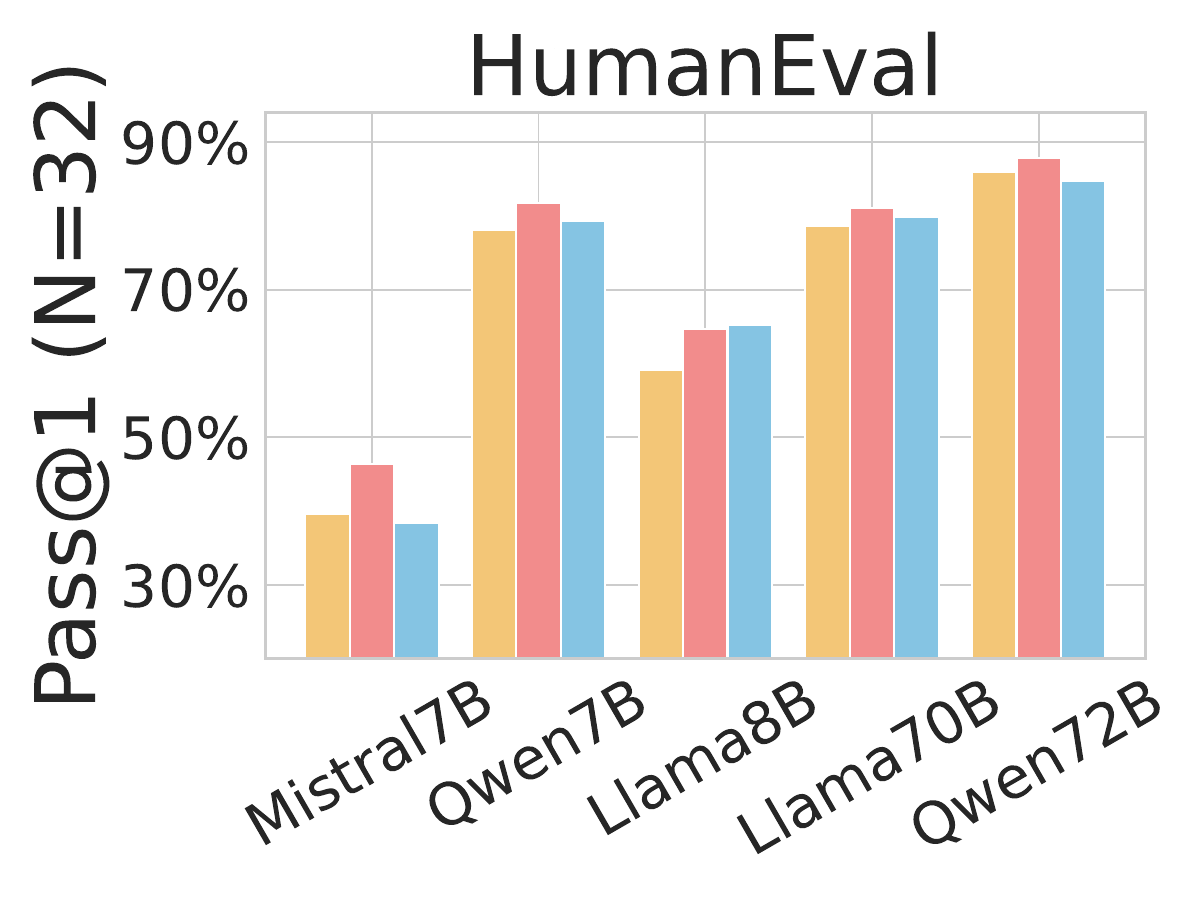}
}
\subfigure{
\includegraphics[width=0.225\textwidth]{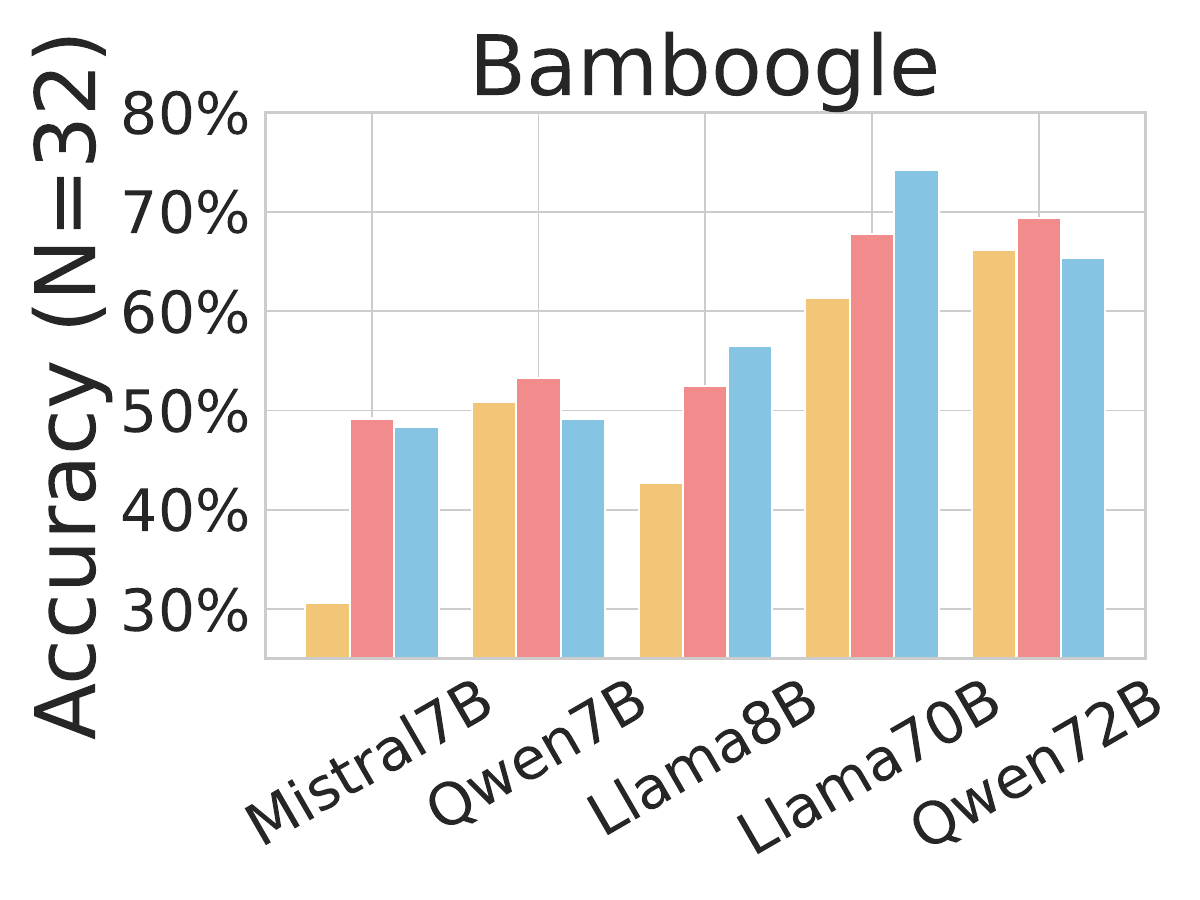}
}
\vspace{-0.2cm}
\caption{Accuracy (\%) across benchmark tasks under different instruction prompts. The results underscore the substantial influence of prompt design on inference-time computation methods. Notably, self-correction mechanisms produced mixed outcomes.
}~\label{fig:trick_prompt}
\vspace{-0.4in}
\label{fig:hist}
\end{figure}

\subsubsection{Generating Candidate Solutions }
Generating candidate solutions is a critical step in inference-time computation for LLM reasoning, but the inherent randomness in this process significantly influences diversity. Hyperparameters such as temperature and top-p, along with strategies like instruction prompts, play a vital role in shaping and guiding the solution trajectory. For example, temperature, as a sampling strategy in token generation, increases diversity at higher values. Therefore, this study focuses on the candidate solution generation process, including instruction prompt types, temperature, and top-p sampling.

\begin{figure}[t]
\vspace{-0.1cm}
\centering
\includegraphics[width=0.48\textwidth]{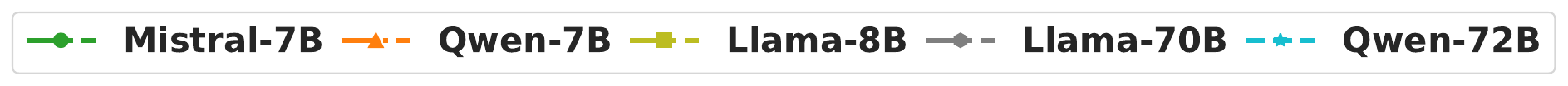}
\subfigure{
\includegraphics[width=0.225\textwidth]{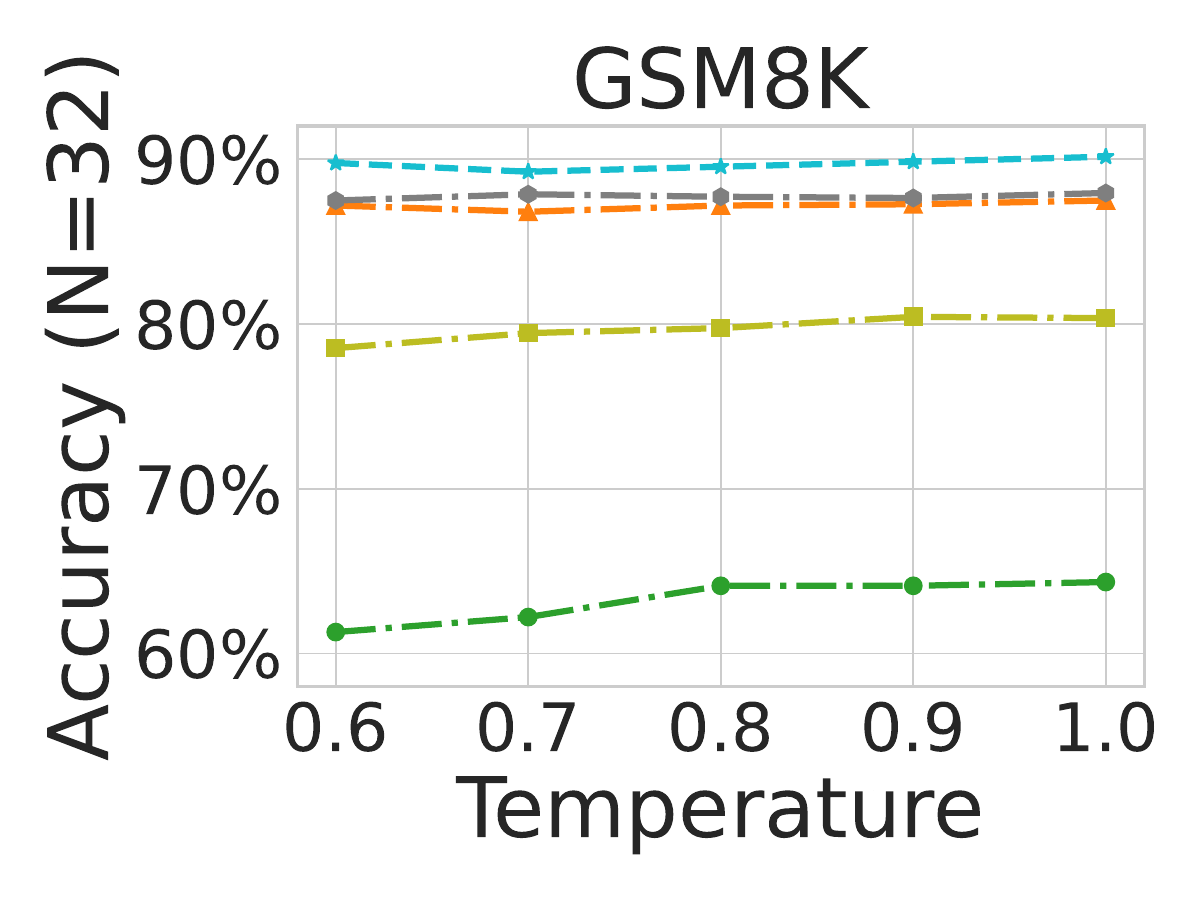}
}
\subfigure{
\includegraphics[width=0.225\textwidth]{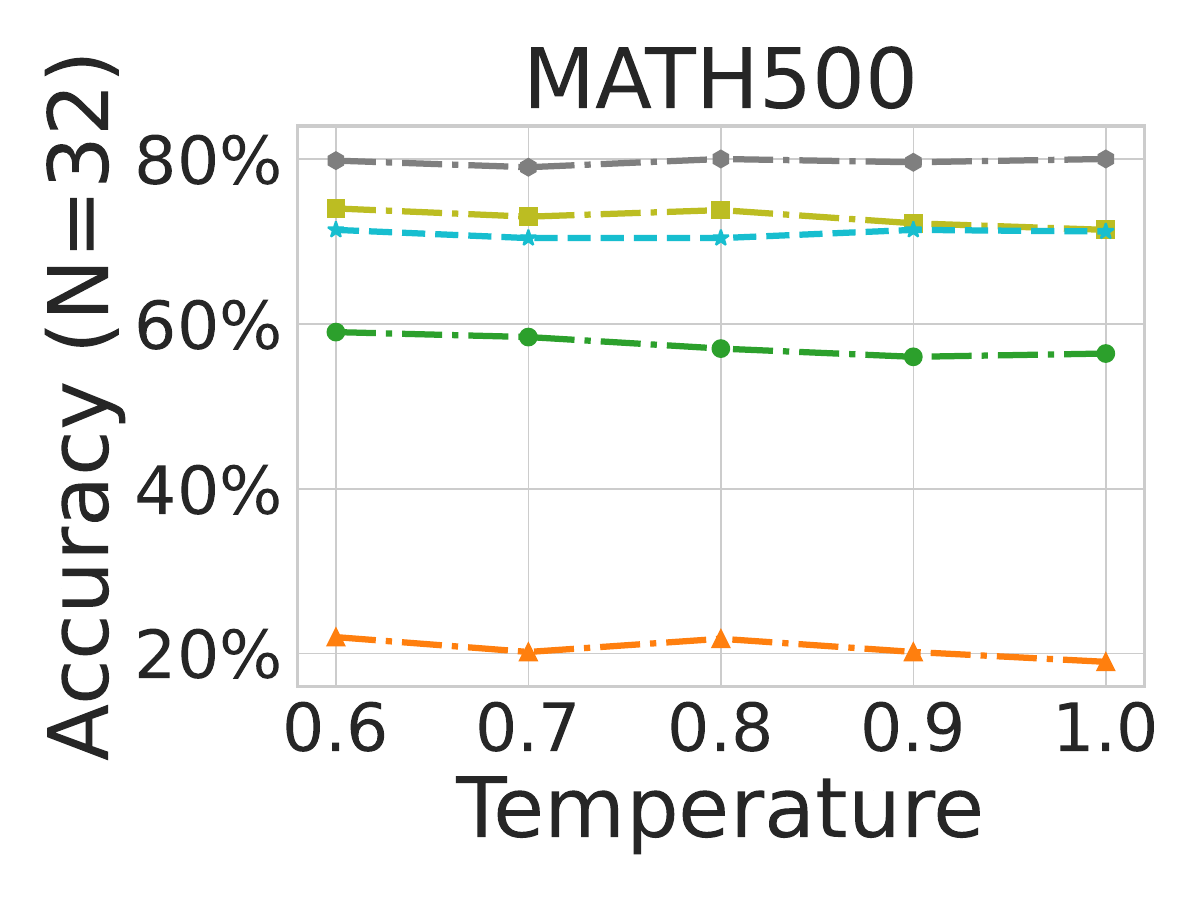}
}
\vspace{-0.4cm} \\  
\subfigure{
\includegraphics[width=0.225\textwidth]{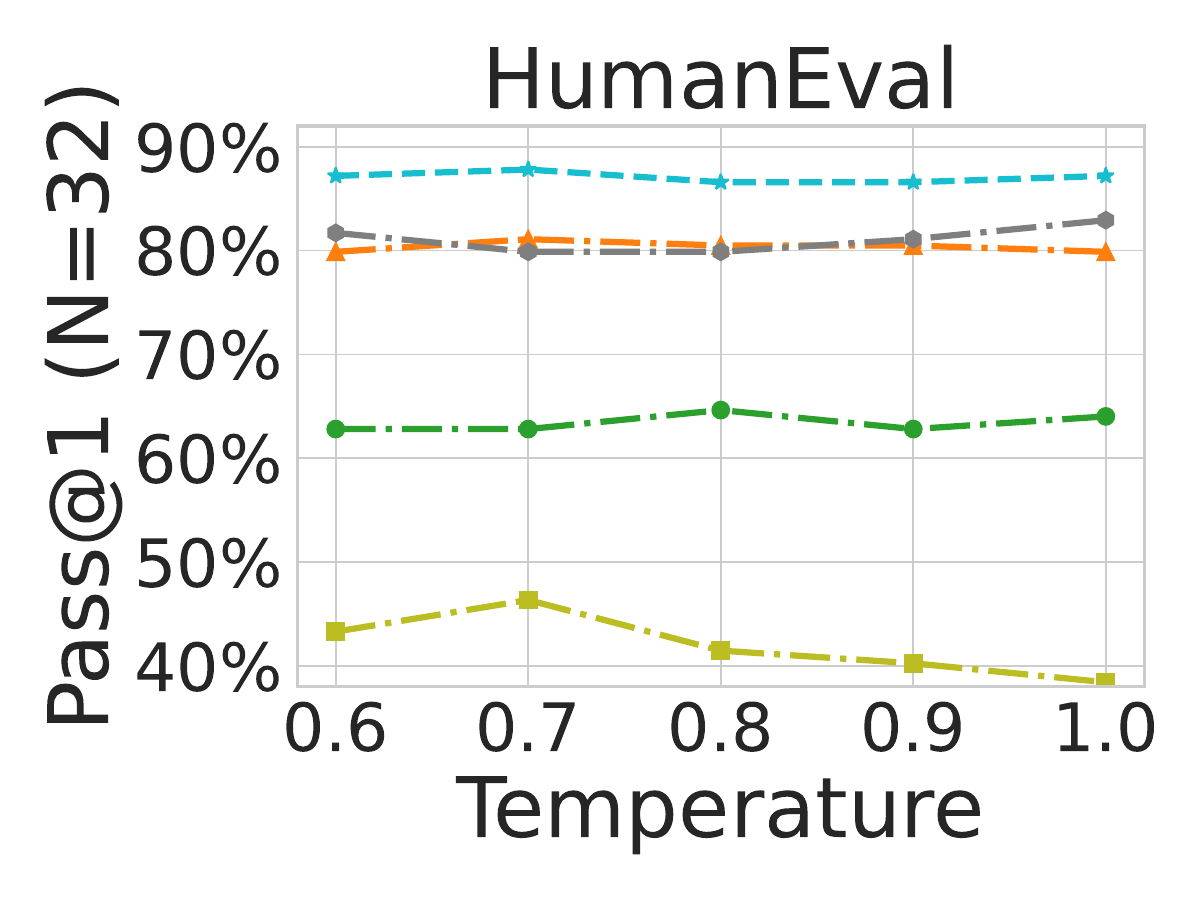}
}
\subfigure{
\includegraphics[width=0.225\textwidth]{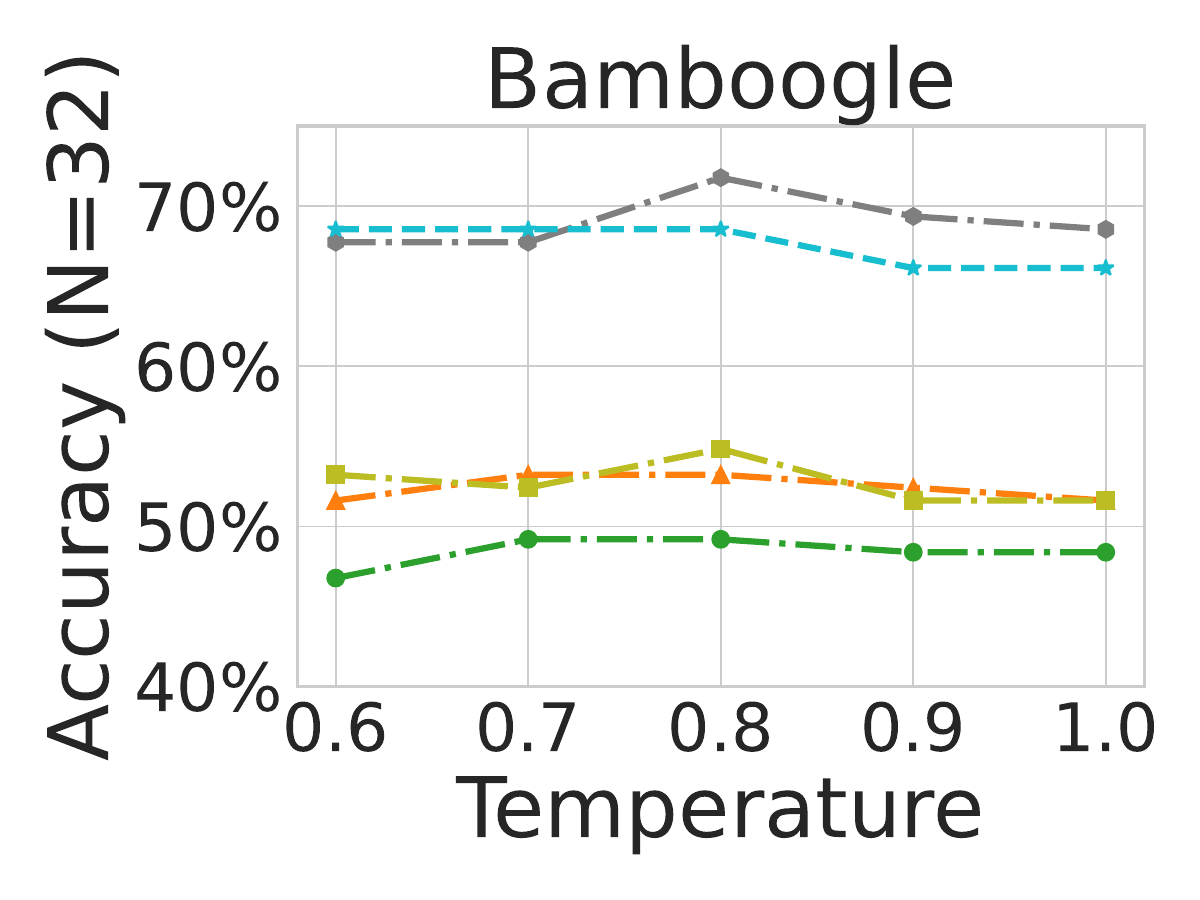}
}
\caption{Accuracy (\%) versus temperature during inference. Temperature varies from 0.6 to 1.0, highlighting its impact on LLM reasoning, with performance stabilizing around the recommended default of 0.8.  
}
\vspace{-0.2in}
\label{fig:trick_temp}
\end{figure}

\textbf{Instruction Prompt Type.} Different instruction prompts can guide an LLM to generate distinct reasoning paths. Specifically, Input-Output (IO) prompts directly provide the answer, whereas Chain-of-Thought (CoT) prompts encourage the LLM to reason step by step. Recent research \cite{huang2023large} suggests that self-correction or self-reflection mechanisms are often ineffective when LLMs operate without external feedback under certain prompt types. We further explores the impact of various prompt types, including IO prompts, standard Chain-of-Thought (CoT) prompts, and reflection-based CoT. As shown in Figure~\ref{fig:trick_prompt}, the results illustrate that different prompts have a significant effect on LLM reasoning performance. Specifically, we observe that while CoT prompts generally improve reasoning accuracy, Reflection-based CoT prompts show more mixed results across the datasets. These findings are consistent with the observations in \cite{huang2023large}, where self-correction mechanisms failed to show a consistent improvement, with the outcomes varying across different tasks.

\textbf{Temperature.}  Temperature~\cite{hinton2015distilling} $\tau$ regulates the diversity of candidate solutions in LLMs. A higher  $\tau$ decreases prediction confidence but increases output variability. We revisit the previous inference-time computation settings in Table~\ref{tab:config_llm_reasoning}, where the temperature is set differently for each case. Figure~\ref{fig:trick_temp} illustrates its effect. In most reasoning scenarios, the LLM's performance is optimized at $\tau$= 0.8, yielding an improvement of approximately 2.32\% to 4.83\% across four datasets. In most cases, both larger and smaller values of 
$\tau$ result in decreased performance.

\begin{figure}[t]
\centering
\includegraphics[width=0.48\textwidth]{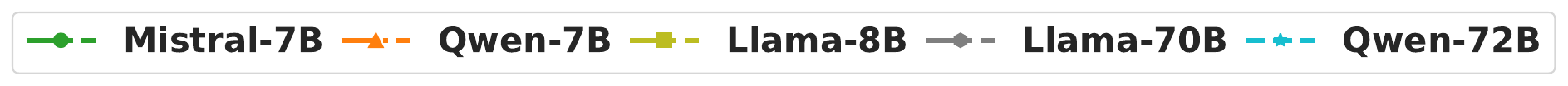}
\subfigure{
\includegraphics[width=0.225\textwidth]{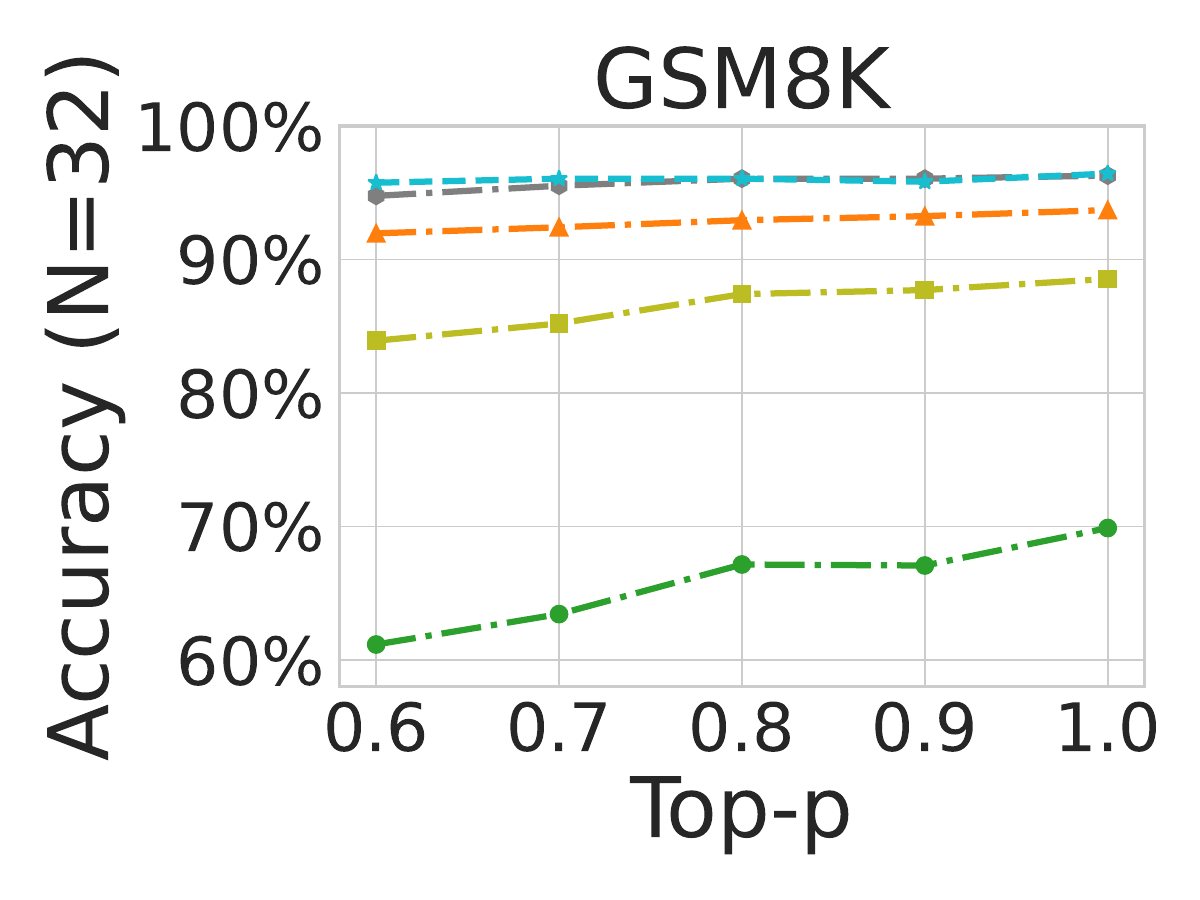}
}
\subfigure{
\includegraphics[width=0.225\textwidth]{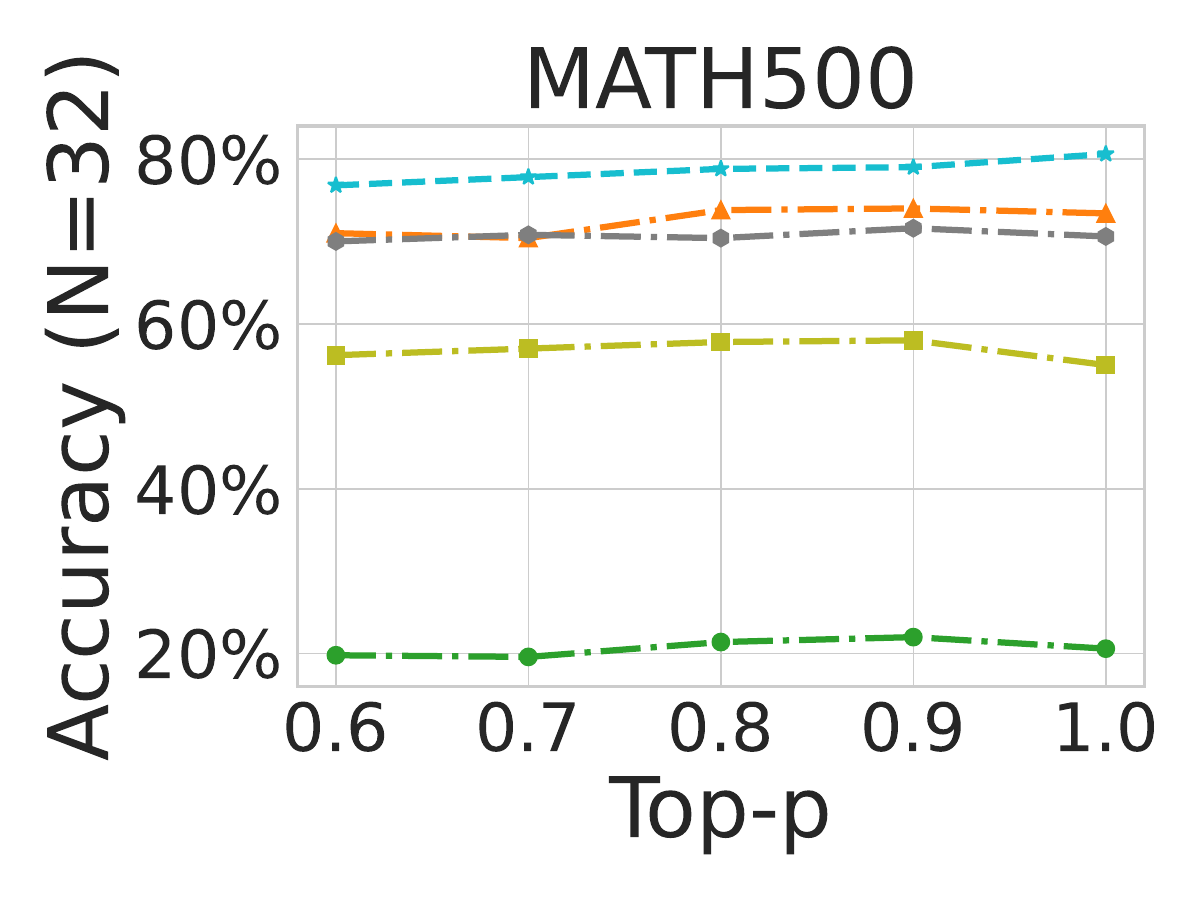}
} 
\vspace{-0.2in}\\  
\subfigure{
\includegraphics[width=0.225\textwidth]{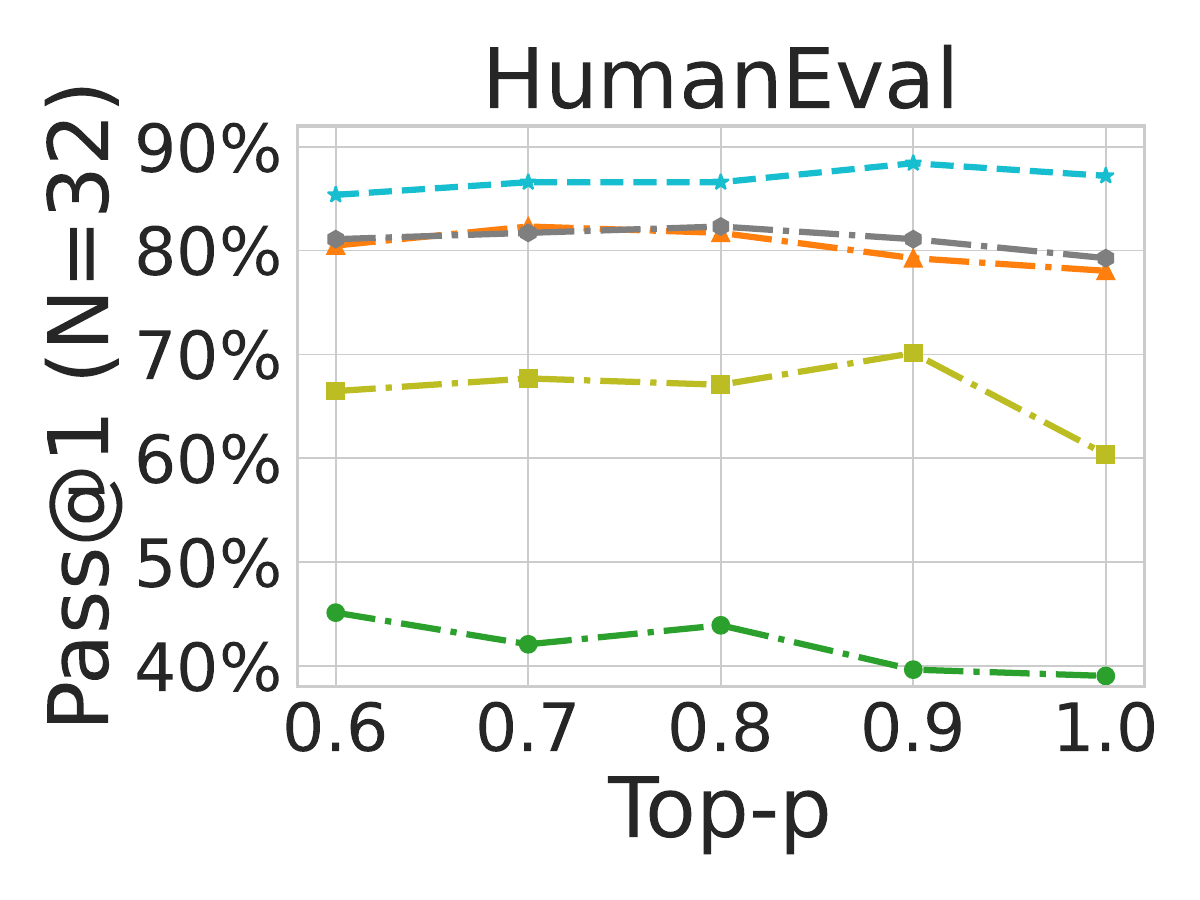}
}
\subfigure{
\includegraphics[width=0.225\textwidth]{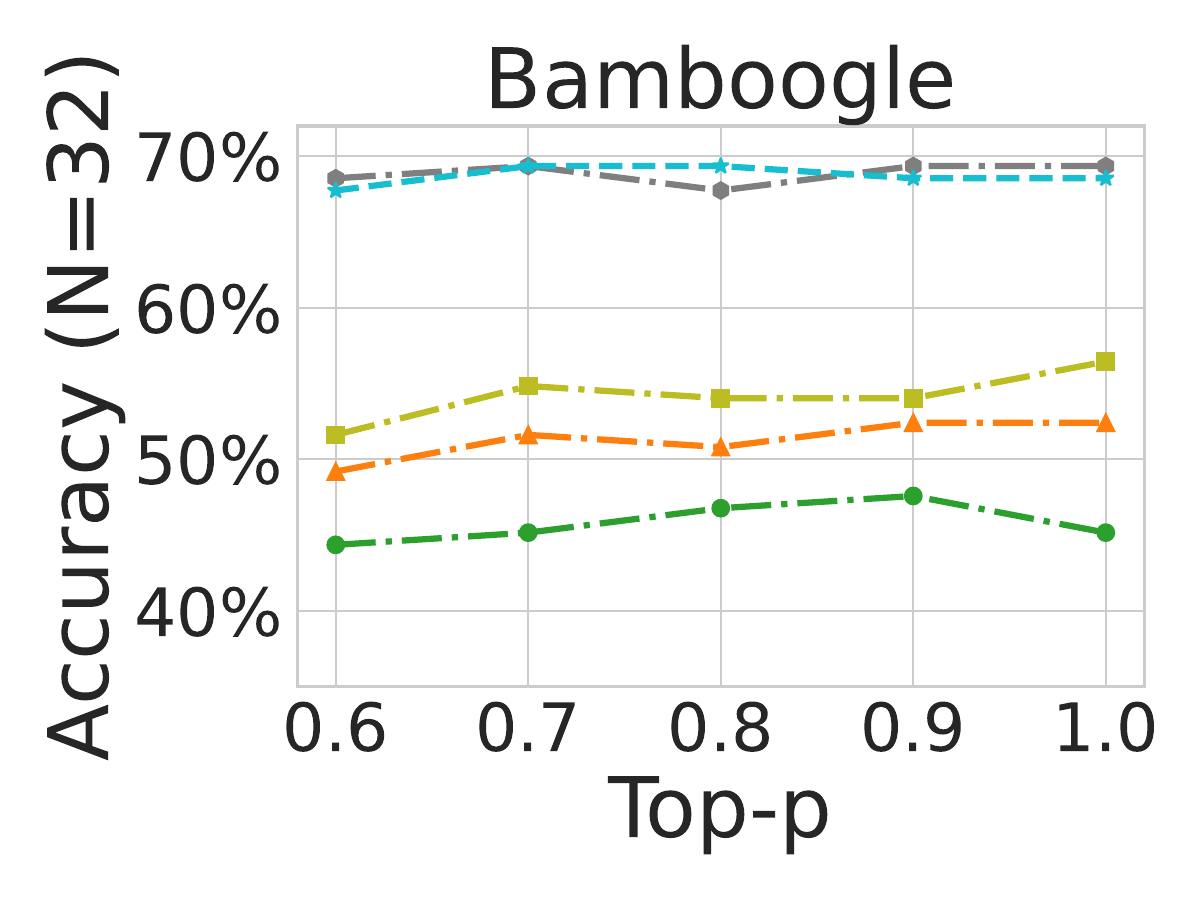}
}
\caption{Accuracy (\%) versus Top-p values during inference. Top-p varies from 0.6 to 1.0, highlighting its impact on LLM reasoning, with performance stabilizing around the recommended default of 0.9.  
}
\vspace{-0.2in}
\label{fig:trick_top_p}
\end{figure}

\textbf{Top-p.} The top-p parameter regulates the output of an LLM by modifying the effective vocabulary size  $M$ , considering only those tokens whose cumulative probability exceeds a predefined threshold. In general, top-p strikes a balance between diversity and quality in the model's generated output. As  $p$  decreases, the model becomes increasingly constrained to a smaller set of high-probability tokens, leading to more focused and deterministic outputs. Conversely, higher values of  $p$  allow for a broader selection of tokens, which increases diversity but may also result in less coherent outputs. Figure~\ref{fig:trick_top_p} demonstrates that the impact of top-p on inference-time computation in LLM reasoning is significant, with an optimal value of  top-p = 0.9 , resulting in an overall improvement of 2.32\%–5.88\%.

\subsubsection{Selecting Optimal Solutions}
Selecting optimal solutions is a critical step in the inference-time computation of LLM reasoning. This process typically involves either selection by the inference model itself (e.g., voting or prompt-based selection) or the use of external reward models (e.g., RLHF, Proof-Critical, or process reward models). A key question is whether LLMs can effectively evaluate their own solutions. However, self-evaluation methods often fall short, as LLMs struggle to correct errors without external guidance. Moreover, reward models frequently fail to distinguish truly correct answers from superficially correct ones, leading to inflated performance evaluations. This challenge underscores the need for more reliable evaluation mechanisms. To address these gaps, we study the selection process, focusing on self-evaluation, reward types, and investigating  generalization of improved reward models.

\begin{figure}[t]
\centering
\includegraphics[width=0.48\textwidth]{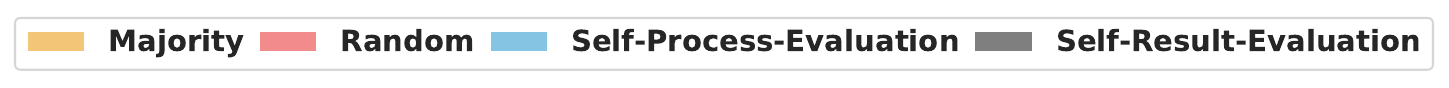}
\subfigure{
\includegraphics[width=0.225\textwidth]{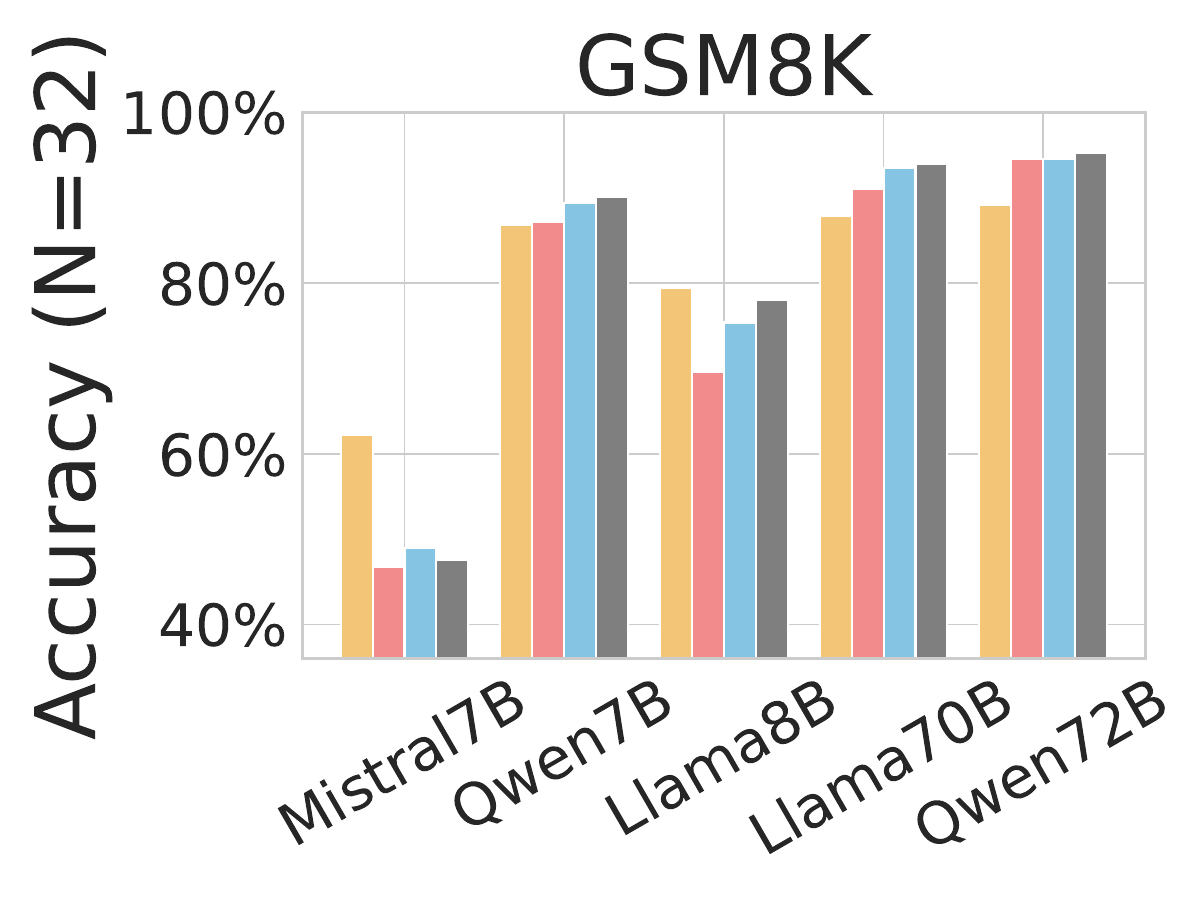}
}
\subfigure{
\includegraphics[width=0.225\textwidth]{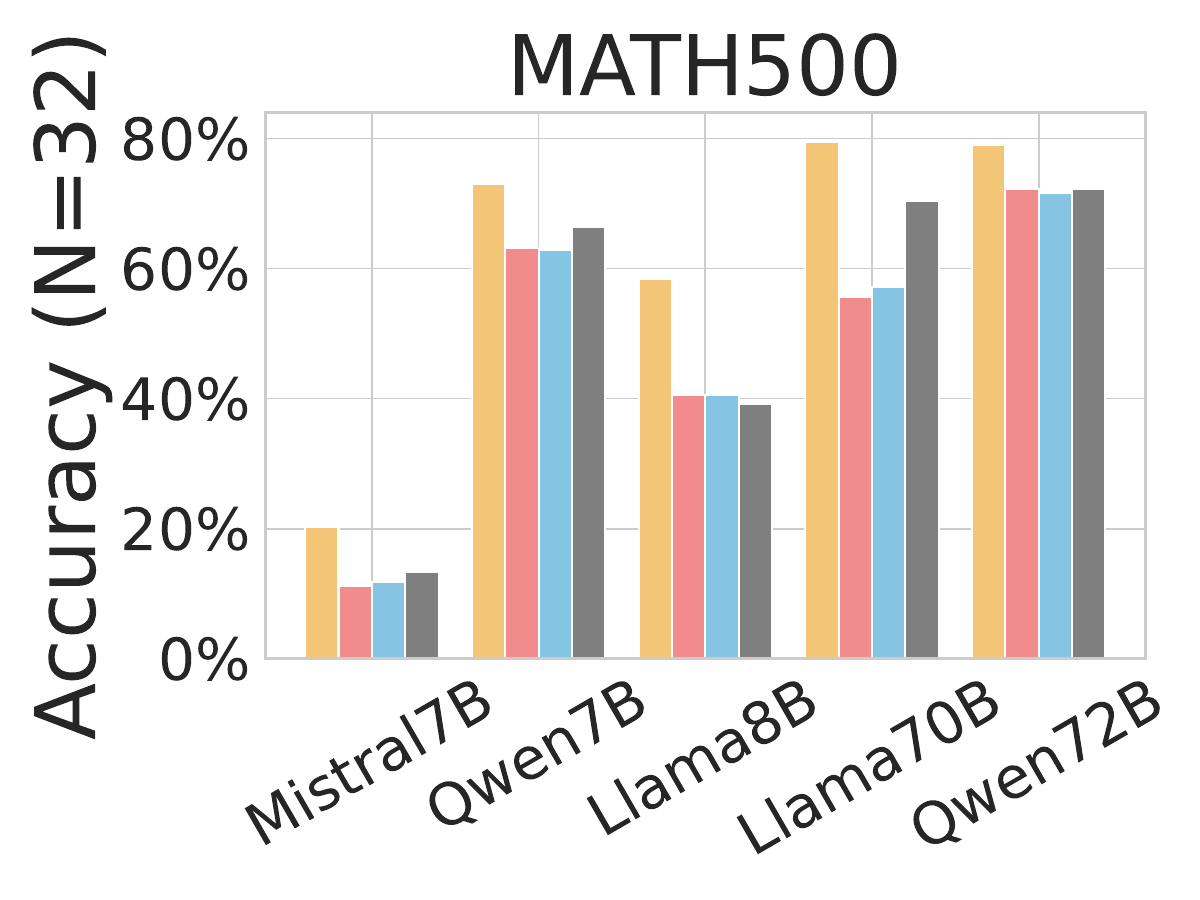}
} \vspace{-0.15in}\\  
\subfigure{
\includegraphics[width=0.225\textwidth]{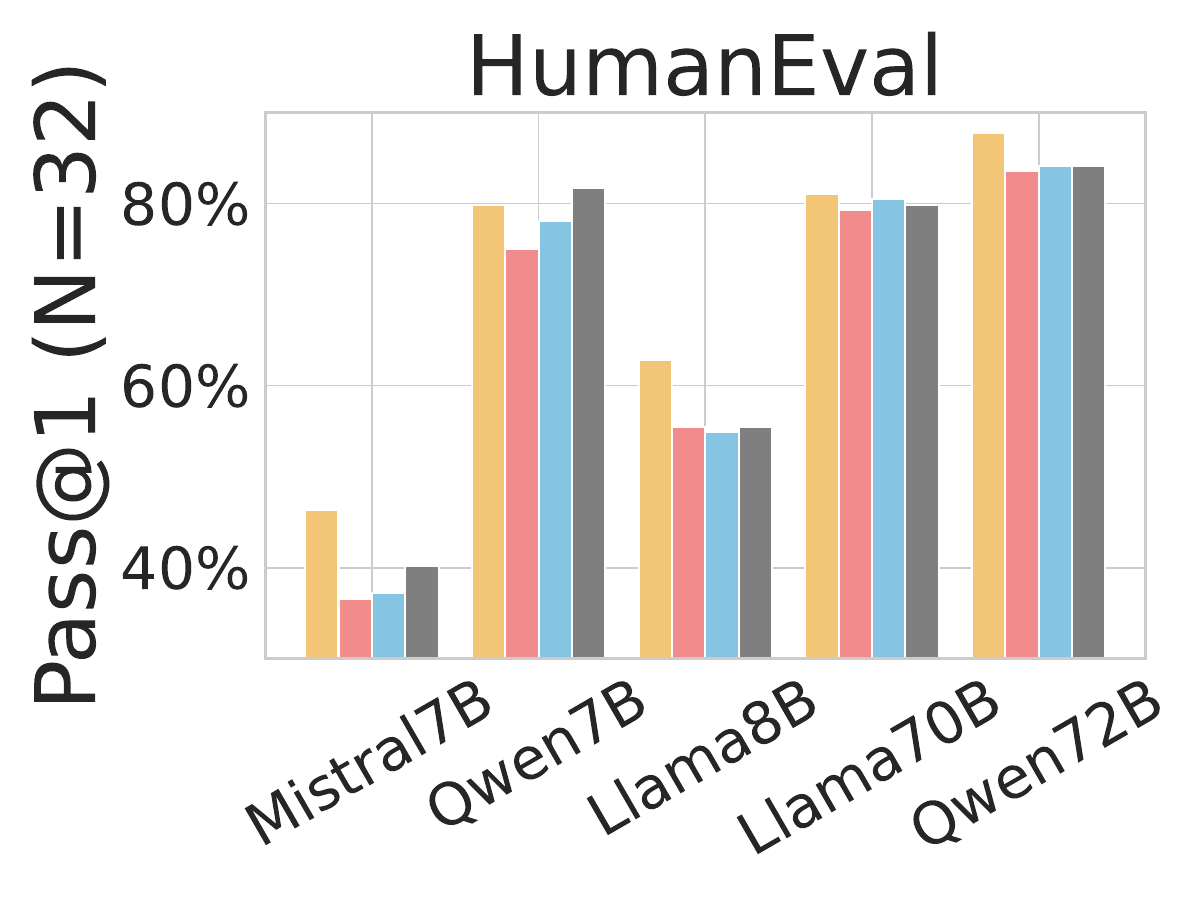}
}
\subfigure{
\includegraphics[width=0.225\textwidth]{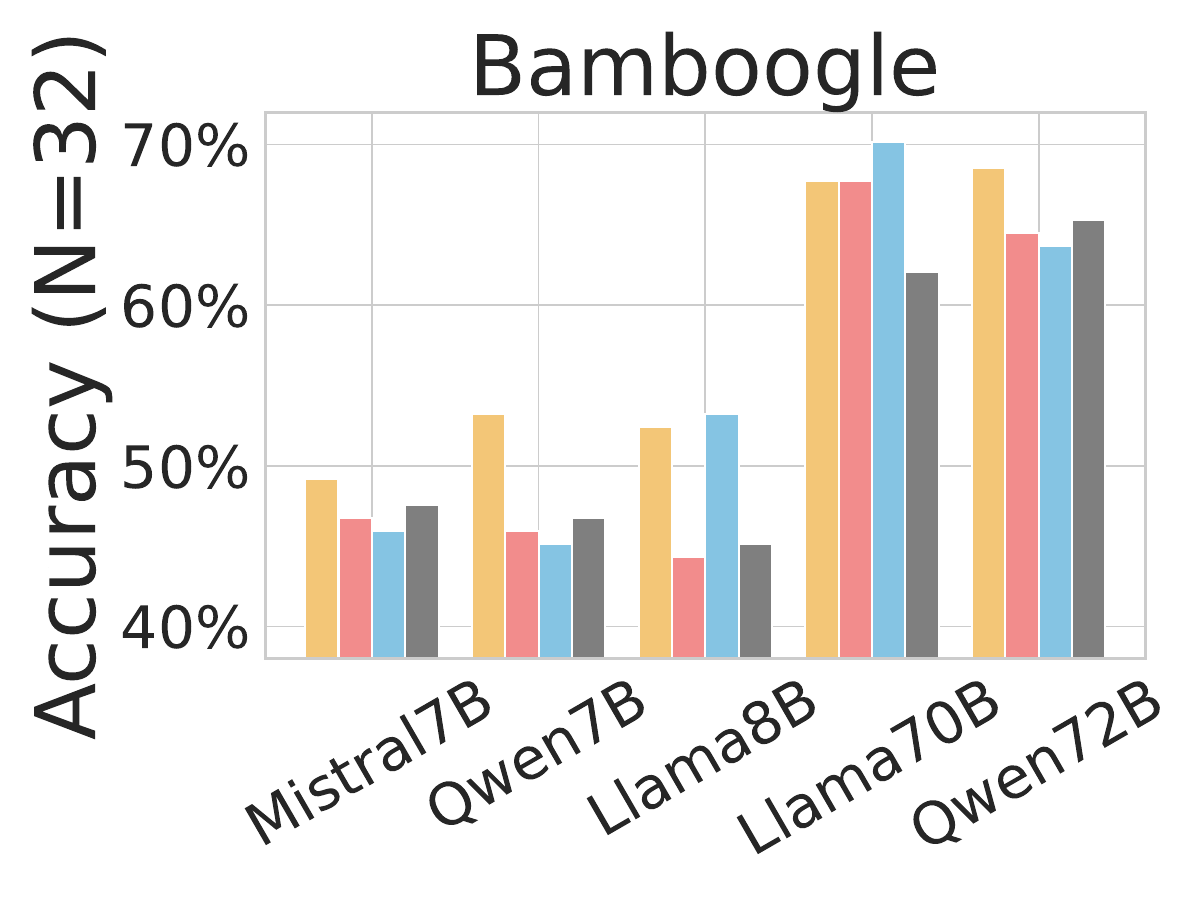}
}
\vspace{-0.15in}
\caption{ Accuracy (\%) across four benchmark tasks  with different evaluation strategies.  The results shows that Self-evaluation often fails to assess solution quality effectively.
}~\label{fig:trick_self_evalaution}
\end{figure}

\textbf{Self-Evaluation.}
Previous studies~\cite{zhang2024chain, yao2024tree} have explored self-evaluation methods for selecting the optimal candidate solution in the inference phase, where the model assesses its own generated solutions using fuzzy judgments (e.g., categorizing solutions as "impossible," "likely," or "sure" to solve a problem). These fuzzy evaluations are then translated into probabilities, such as mapping "impossible" to 0.01 and "sure" to 1. We examine the effectiveness of self-evaluation approaches, including self-process evaluation and self-result evaluation, in comparison to random selection and majority voting. Figure~\ref{fig:trick_self_evalaution} shows that self-evaluation does not consistently improve performance; in fact, in some worst-case scenarios, it performs worse than random selection.

\begin{figure}[t]
\centering
\includegraphics[width=0.48\textwidth]{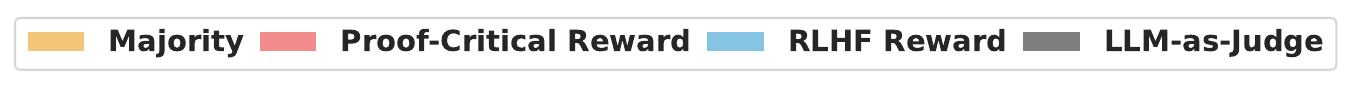}
\subfigure{
\includegraphics[width=0.225\textwidth]{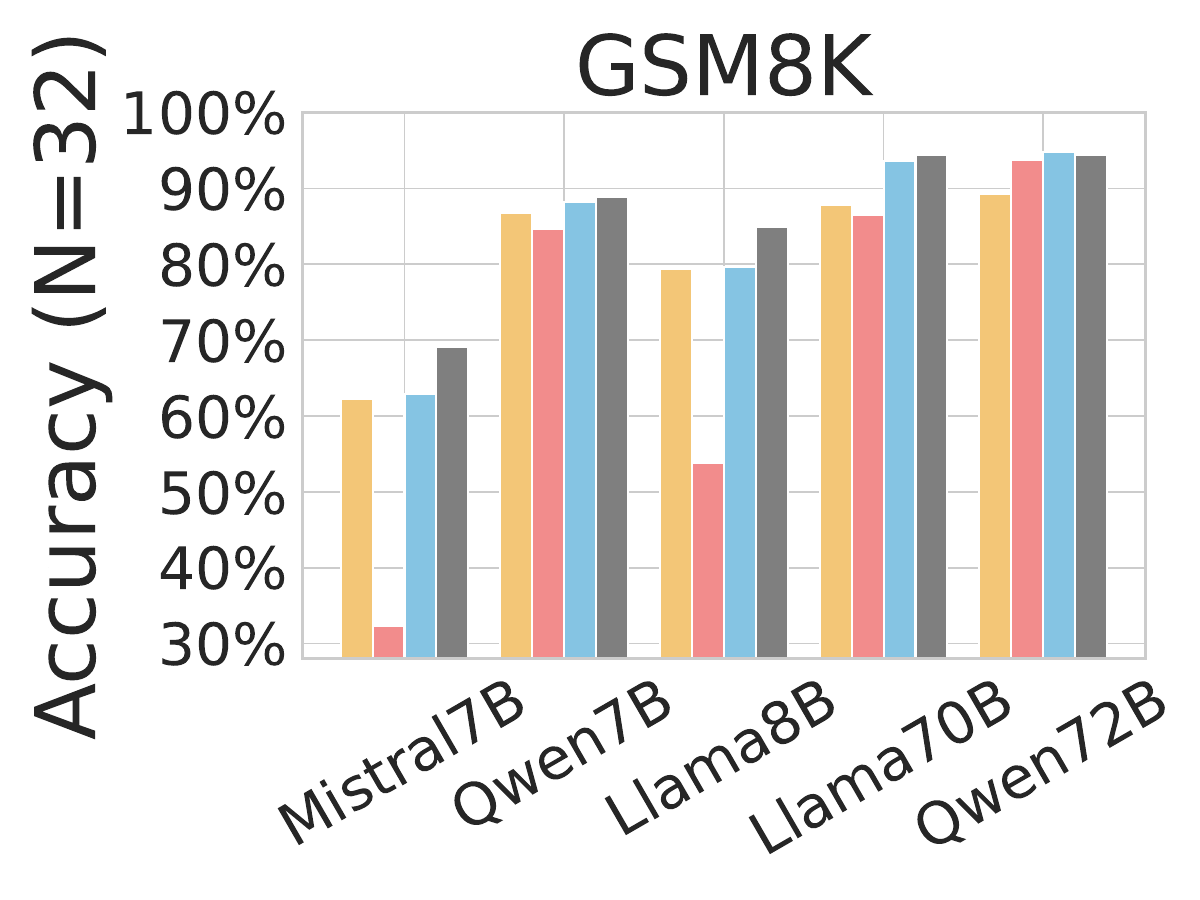}
}
\subfigure{
\includegraphics[width=0.225\textwidth]{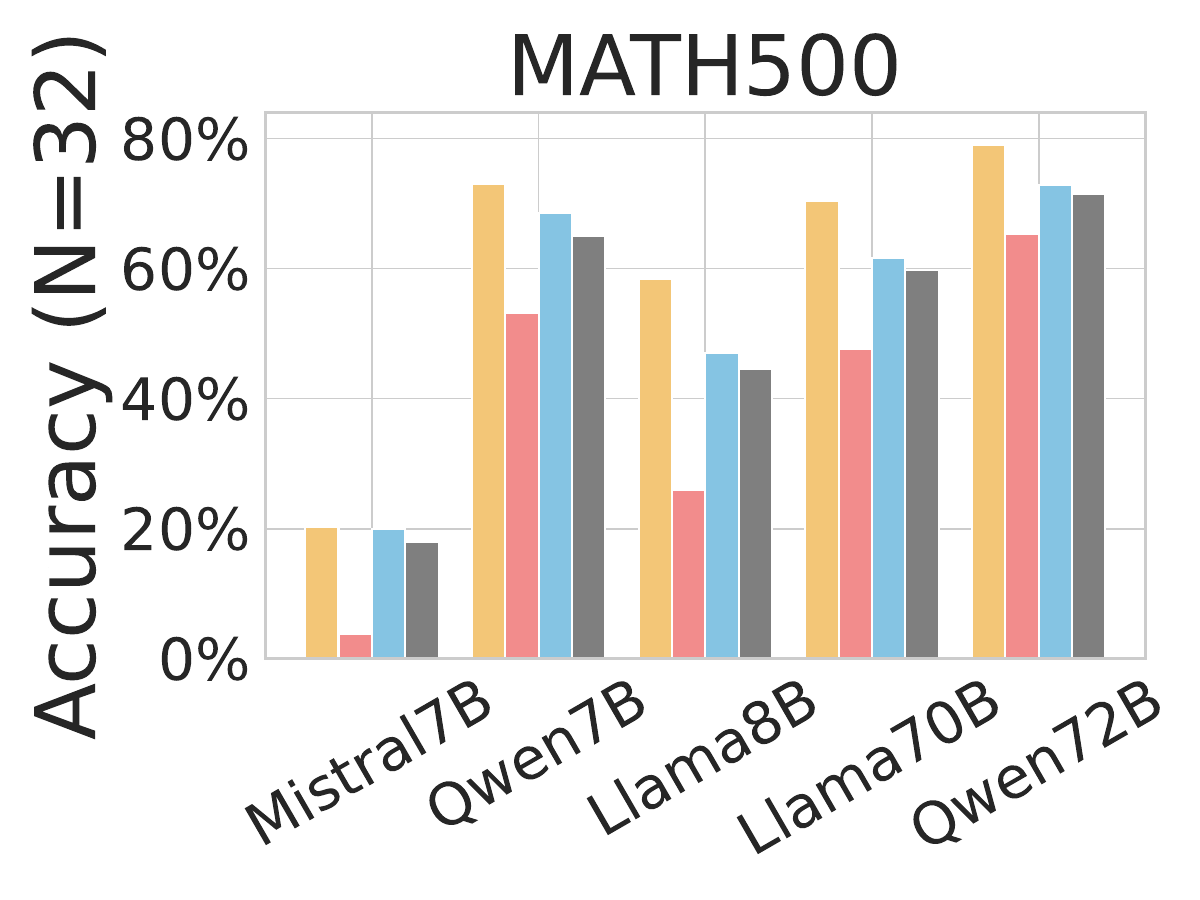}
}
\vspace{-0.15in}\\  
\subfigure{
\includegraphics[width=0.225\textwidth]{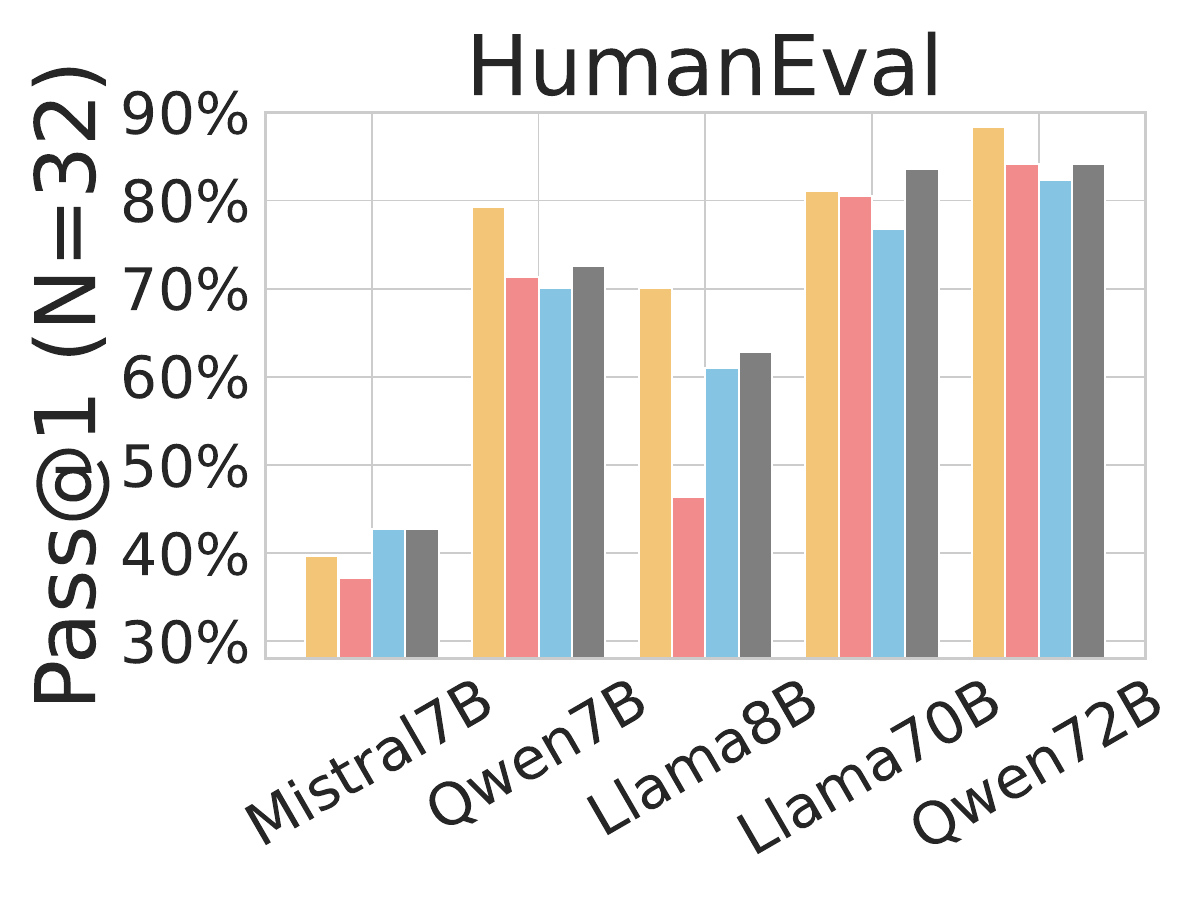}
}
\subfigure{
\includegraphics[width=0.225\textwidth]{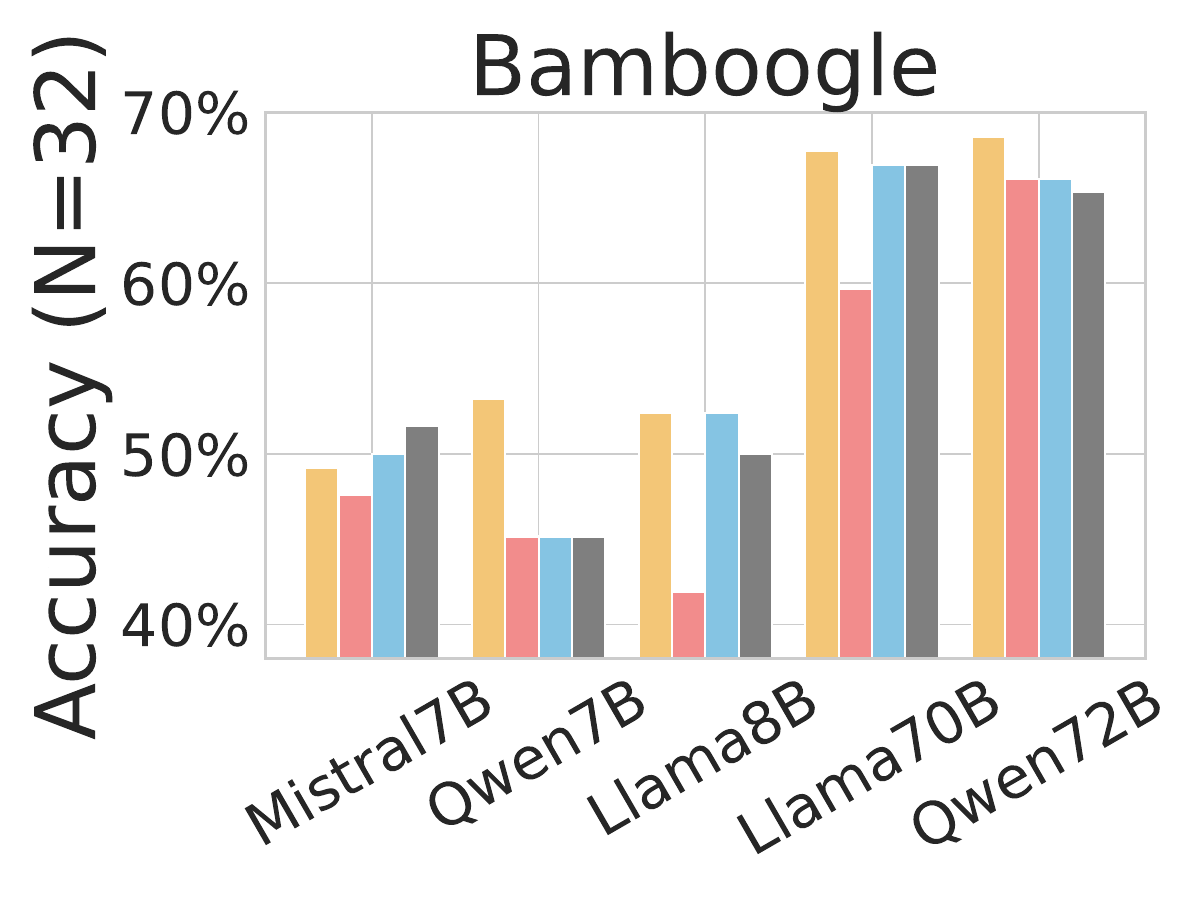}
}
\vspace{-0.2in}
\caption{Comparison of different reward models  across benchmarks showcasing their impact on accuracy and pass@1 for multiple LLMs.}~\label{fig:trick_evalaution_type}
\vspace{-0.2in}
\end{figure}

\textbf{Reward Type.} 
Recently developed reward models (e.g., reward models, critic models, process reward models) have become key tools in enhancing the reasoning capabilities of LLMs during the inference-time phase~\cite{processbench, prmlessons}. We investigate various reward types, including RLHF Reward~\cite{cai2024internlm2}, Proof-Critical Reward, LLM-as-Judge, and majority voting. Figure~\ref{fig:trick_evalaution_type} illustrates the significant impact of different reward models on inference-time computation performance. Specifically, for knowledge-based reasoning, certain reward models can substantially improve performance. In contrast, for more complex reasoning tasks such as mathematics and code generation, the LLM-as-Judge process reward seems to provide a greater performance boost.
\begin{table*}[t]  
\centering  
\small
\caption{LLM Reasoning Performance under Inference-Time Computation with Fixed Token Budget. The table shows accuracy and token consumption across various reasoning tasks for Llama-3.1-8B and Qwen-2.5-7B models. }  ~\label{tab:main_result}  
\vspace{-0.2in}
  \begin{center}  
  \resizebox{\textwidth}{!}{
      \begin{tabular}{l||c|c|c|c|c|c|c|c||}
        \toprule
        \multirow{3}{*}{Method}& 
        \multicolumn{3}{c|}{\textbf{Knowledge-based Reasoning}} & \multicolumn{5}{c||}{\textbf{Complex Reasoning}} 
        \\
        &\multicolumn{3}{c|}{Accuracy (\%) / \#Tokens}& 
        \multicolumn{1}{c|}{Pass@1 / \#Tokens}&
        \multicolumn{4}{c||}{Accuracy (\%)  / \#Tokens} 
        \\\cmidrule{2-9}
        &\multicolumn{1}{c}{Bamboogle}&
        \multicolumn{1}{c}{HotPotQA}&
        \multicolumn{1}{c|}{Fever}&
        \multicolumn{1}{c}{HumanEval}&
        \multicolumn{1}{c}{GSM8K}&
        \multicolumn{1}{c}{GSM-HARD}&
        \multicolumn{1}{c}{MATH}&
        \multicolumn{1}{c||}{PrOntoQA}\\
        \midrule
        \multicolumn{9}{l}{\textit{Llama-3.1-8B}}\\
        \midrule
        \textbf{Best-of-N}
        & 48.4 / 1077 & \textbf{48.6} / 876 & 61.1 / 1191 & \textbf{60.4} / 1244 & 83.1 / 976 & \textbf{33.6} / 1174 & 44.6 / 1412 & \textbf{94.0} / 969 \\
        \textbf{Step-Level Best-of-N}
        & 52.4 / 1119 & 45.7 / 632 & 60.3 / 581 & 55.5 / 921 & 83.0 / 845 & 32.9 / 1013 & 39.8 / 708 & 93.8 / 881 \\
        \textbf{Beam Search}
        & 33.1 / 1801 & 31.8 / 2091 & 57.3 / 917 & 50.6 / 1019 & 78.7 / 1066 & 25.1 / 1328 & 43.2 / 1014 & 89.6 / 1241 \\
        \textbf{MCTS}
        & 49.2 / 896 & 41.3 / 956 & 59.9 / 843 & 48.7 / 1114 & 81.3 / 582 & 31.6 / 627 & 31.0 / 1898 & 93.0 / 1511 \\
        \textbf{Self-Consistency}
        & \textbf{54.8} / 1077 & 48.3 / 876 & \textbf{61.9} / 1191 & 60.3 / 1261 & \textbf{84.3} / 976 & 33.2 / 1174 & \textbf{45.0} / 1412 & 93.4 / 969 \\
        \textbf{Self-Refine}
        & 41.9 / 287 & 40.5 / 280 & 19.1 / 265 & 41.4 / 495 & 64.5 / 410 & 25.1 / 564 & 37.8 / 743 & 62.2 / 525 \\
        \midrule
        \multicolumn{9}{l}{\textit{Qwen-2.5-7B}}\\
        \midrule
        \textbf{Best-of-N}
        & 49.2 / 881 & \textbf{44.6} / 1189 & 55.6 / 830 & 78.0 / 1102 & 90.5 / 1462 & 50.1 / 1174 & 61.2 / 1827 & \textbf{96.2} / 1271 \\
        \textbf{Step-Level Best-of-N}
        & \textbf{51.6} / 863 & 43.8 / 1055 & 50.2 / 995 & 73.1 / 691 & 89.7 / 709 & \textbf{50.7} / 877 & 63.0 / 1162 & 92.8 / 625 \\
        \textbf{Beam Search}
        & 29.0 / 1552 & 44.3 / 1280 & \textbf{55.7} / 997 & 75.6 / 1019 & 88.3 / 1341 & 50.2 / 1417 & 49.2 / 836 & 90.2 / 1089 \\
        \textbf{MCTS}
        & 47.6 / 1467 & 42.6 / 2585 & 53.9 / 1442 & 69.8 / 724 & 66.2 / 1416 & 48.7 / 1114 &26.2/ 1805 & 70.6 / 2582 \\
        \textbf{Self-Consistency}
        & 49.2 / 881 & 44.5 / 1189 & 56.7 / 830 & \textbf{79.9} / 1271 & \textbf{90.9} / 1462 & 50.0 / 1174 & \textbf{70.2} / 1878 & 94.8 / 1271 \\
        \textbf{Self-Refine}
        & 48.4 / 232 & 42.3 / 316 & 16.8 / 639 & 41.4 / 496 & 86.1 / 601 & 43.4 / 982 & 55.8 / 947 & 58.4 / 1105 \\
        \bottomrule
      \end{tabular}
  }
  \end{center}  
\vspace{-0.3in}
\end{table*}

\begin{figure}[htb]
\vspace{-0.0cm}
\centering
\includegraphics[width=0.48\textwidth]{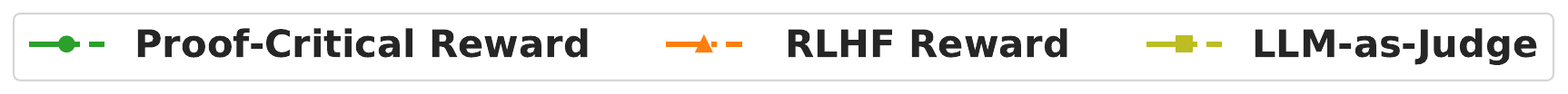}
\subfigure{
\includegraphics[width=0.225\textwidth]{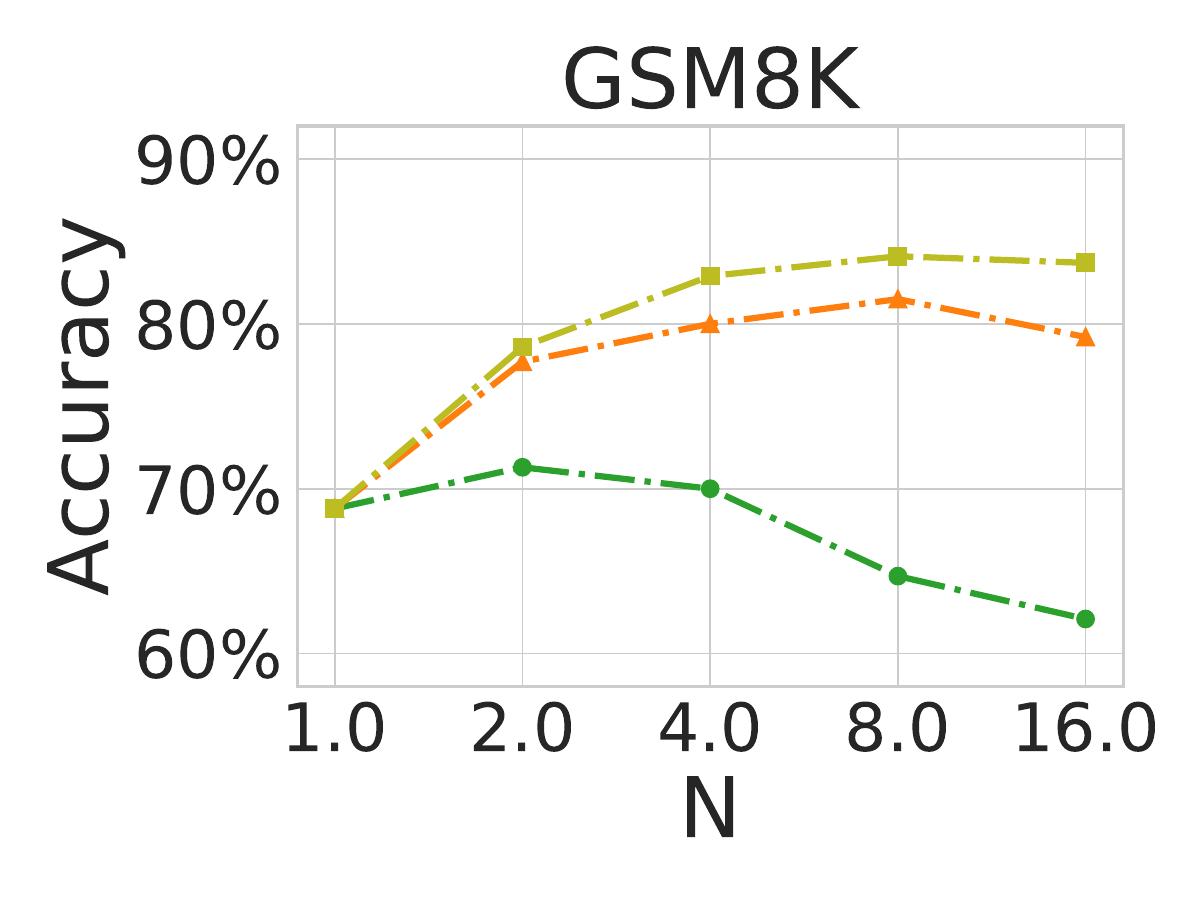}
}
\subfigure{
\includegraphics[width=0.225\textwidth]{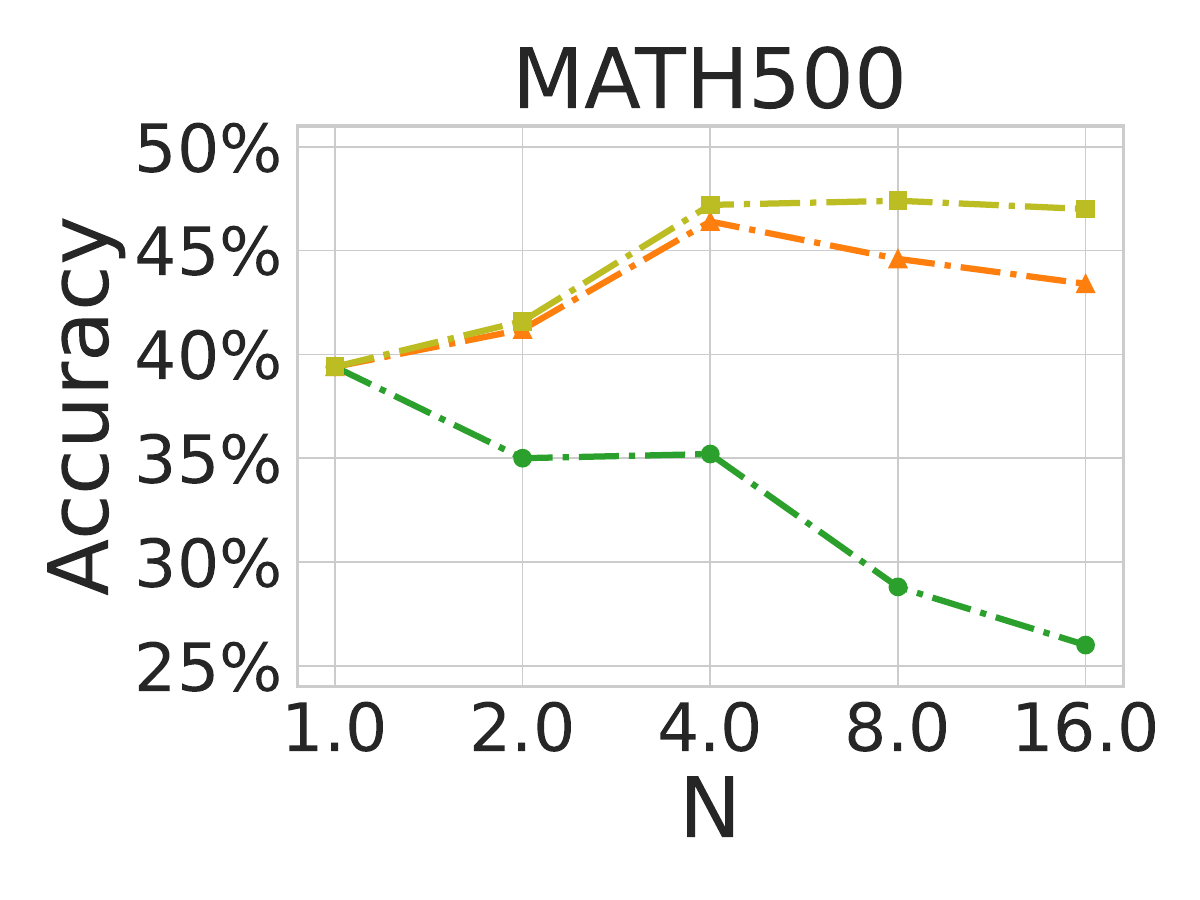}
}
\vspace{-0.2in} \\  
\subfigure{
\includegraphics[width=0.225\textwidth]{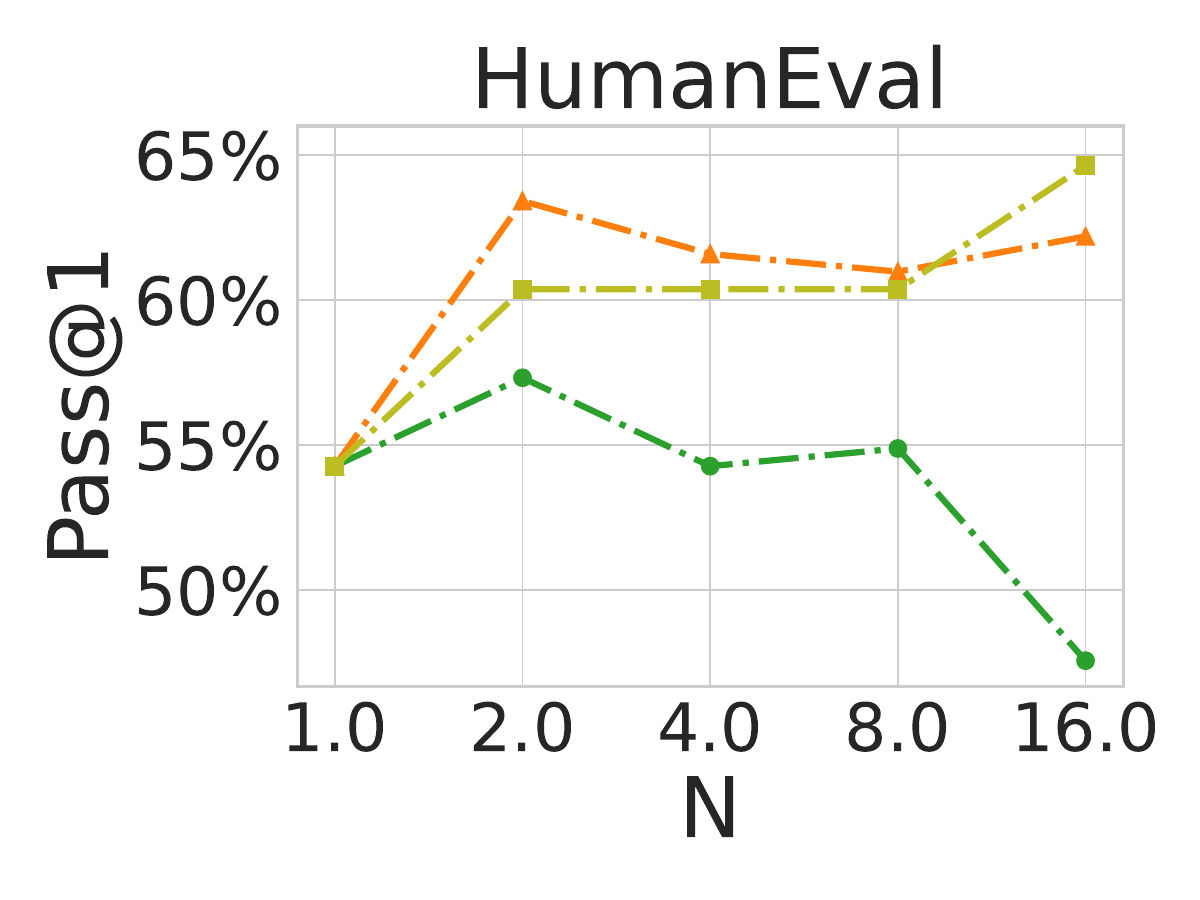}
}
\subfigure{
\includegraphics[width=0.225\textwidth]{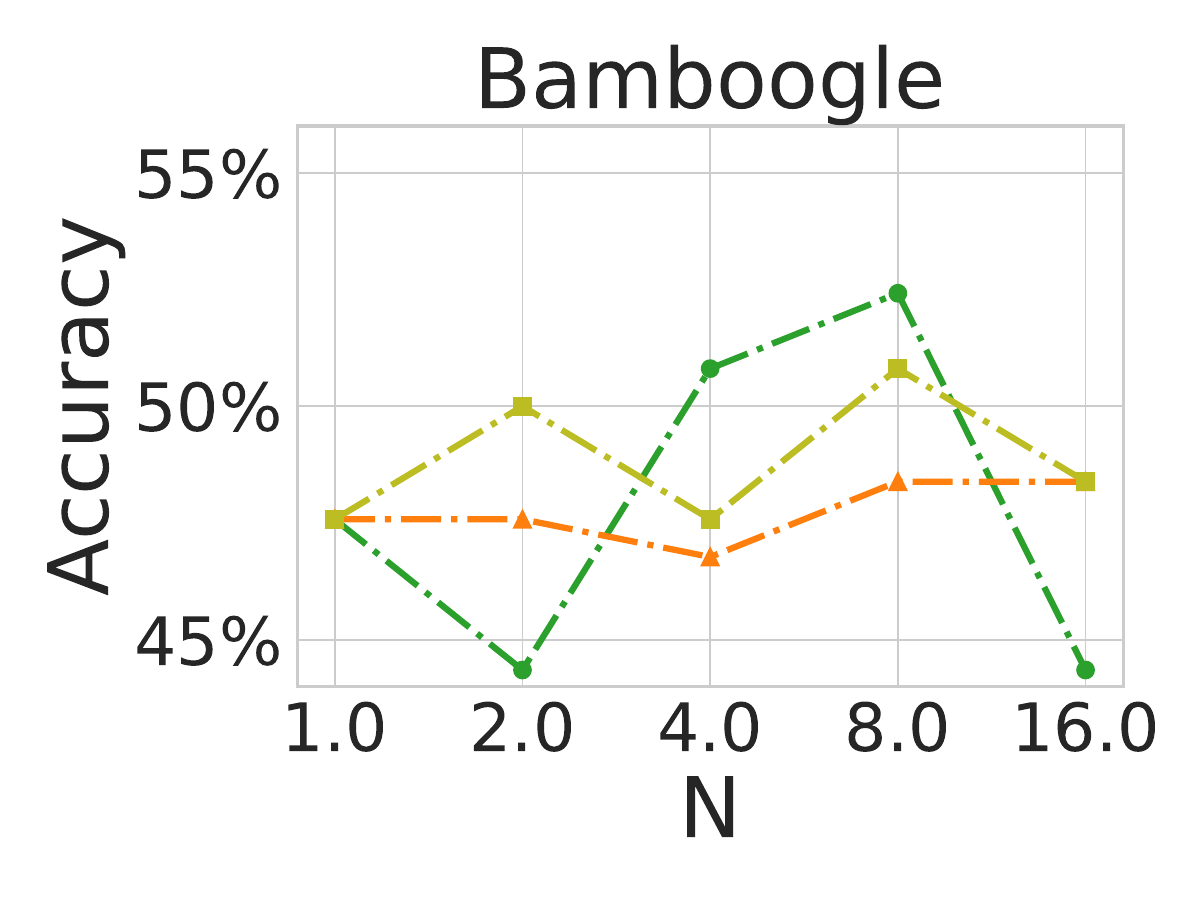}
}
\vspace{-0.2in}
\caption{Scaling test-time performance with reward models (Proof-Critical, RLHF, LLM-as-Judge) across benchmarks.
}
\vspace{-0.0in}
\label{fig:over-optimization-llama}
\end{figure}

\textbf{Performance Fluctuation with Reward Model.} Scaling test-time computation with reward models presents complexities. While one might expect LLM reasoning performance to improve steadily as the reward model is optimized, this is not always the case. Figure~\ref{fig:over-optimization-llama} reports that the reward model does not perform consistently across all cases during scaling. For example, in the challenging MATH task, the Proof-Critical Reward model can lead to a decrease in performance rather than a progressive improvement. This fluctuation of LLM reasoning performance can be attributed to the generalization issues of the reward model, as reported in~\cite{processbench, prmlessons}. Currently, the reward model does not perform well across all tasks.

\begin{table}[htb]  
    \centering  
    \small
    \caption{Combination of Inference-Time Tricks on Llama3.1-8b.  We highlight optimal trick combinations. }  
    \label{tab:tricks_com_llama}  
    \begin{center}  
    \resizebox{0.48\textwidth}{!}{
        \begin{tabular}{c|c|c|c|c|c}  
            \toprule
            \textbf{Prompt} & \textbf{Reward} & \textbf{Temp} & \textbf{Top-p} & \textbf{Bamboogle} & \textbf{MATH} \\
            \midrule
            IO & Majority & 0.7 & 0.9 & 42.7 & 11.8 \\
            CoT & Majority & 0.7 & 0.9 & 52.4 & 58.4 \\
            CoT & Majority & 0.8 & 0.9 & 54.8 & 57.0\\
            CoT & Majority & 0.6 & 0.9 & 53.2 & 58.0\\
            CoT & Majority & 0.7 & 1.0 & 54.0 &  55.0\\
            CoT & Majority & 0.7 & 0.8 & \textbf{56.4} & 57.8\\
            CoT & Random & 0.7 & 0.9 & 44.4& 40.6 \\
            CoT & Self-Process-Evaluation & 0.7 & 0.9 & 53.2& 40.6\\
            CoT & Self-Result-Evaluation & 0.7 & 0.9 & 45.1 &  39.2\\
             Reflect CoT & Majority & 0.7 & 0.9 & \textbf{56.4} & \textbf{58.6} \\\rowcolor{blue!20}
            Reflect CoT & Majority & 0.8 & 0.9 & 55.6 & 57.0 \\
            \bottomrule
        \end{tabular} 
    }
    \end{center} 
\vspace{-0.0in}
\end{table}

\vspace{-0.1in}
\subsubsection{Combination of Inference-Time Computation Tricks.}~\label{sec:combination_trick}
\vspace{-0.00in}
We further investigate the combination of selected useful techniques, including prompt type, temperature, top-p, and the reward model. As demonstrated in Table~\ref{tab:tricks_com_qwen}, the improvements are not always additive when combining different techniques, although methods such as prompt, temperature, and reward model can be effective. For example, as shown in Table~\ref{tab:tricks_com_llama}, using Reflect COT with a temperature of 0.8 and top-p of 0.9 does not result in an improvement for the Bamboogle and MATH tasks. These results suggest that careful selection and combination of techniques can enhance performance, though the impact may vary across different models.

\begin{figure}[t]
\vspace{-0.1cm}
\centering
\includegraphics[width=0.48\textwidth]{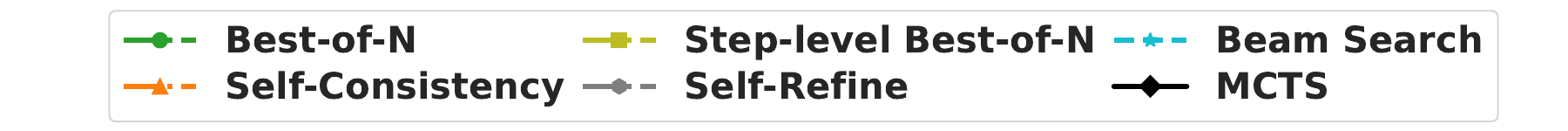}
\subfigure{
\includegraphics[width=0.225\textwidth]{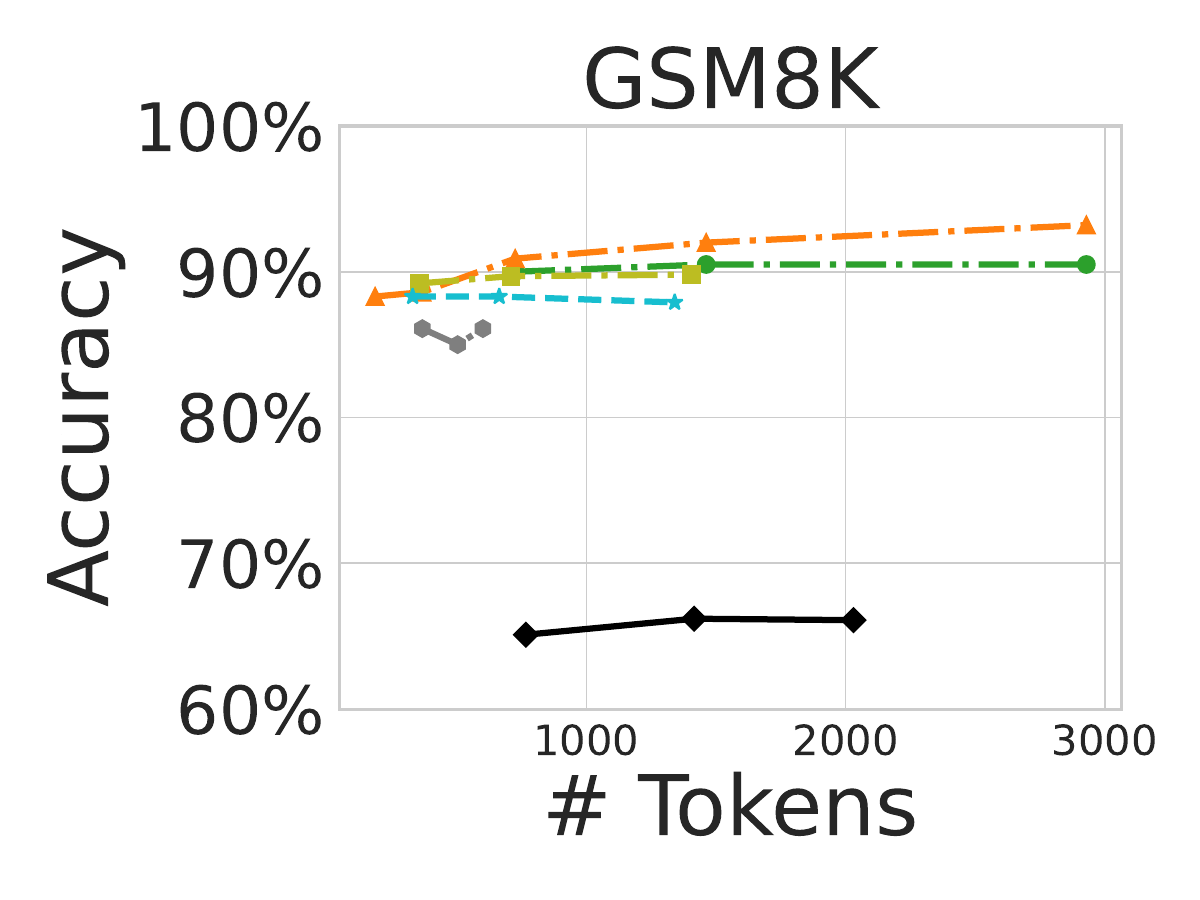}
}
\subfigure{
\includegraphics[width=0.225\textwidth]{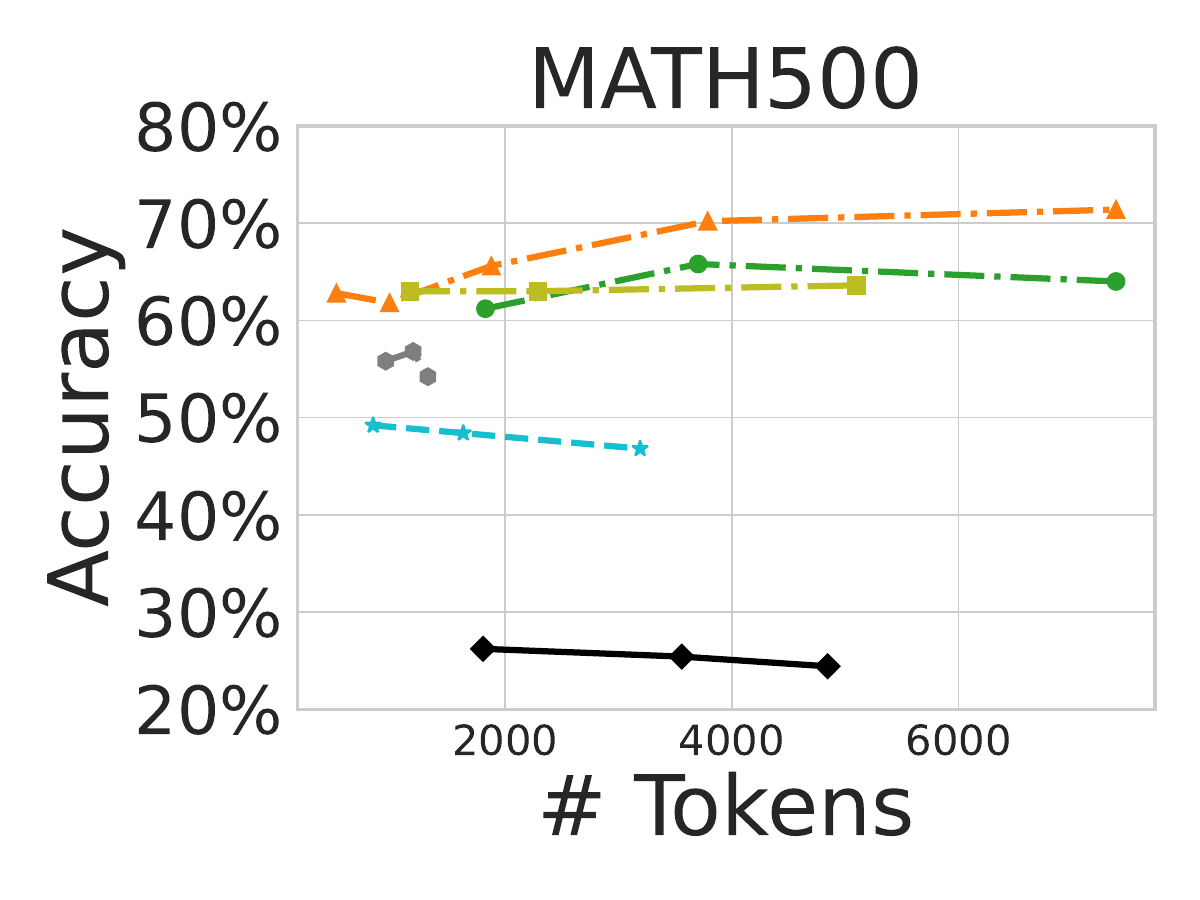}
}
\vspace{-0.2in} \\  
\subfigure{
\includegraphics[width=0.225\textwidth]{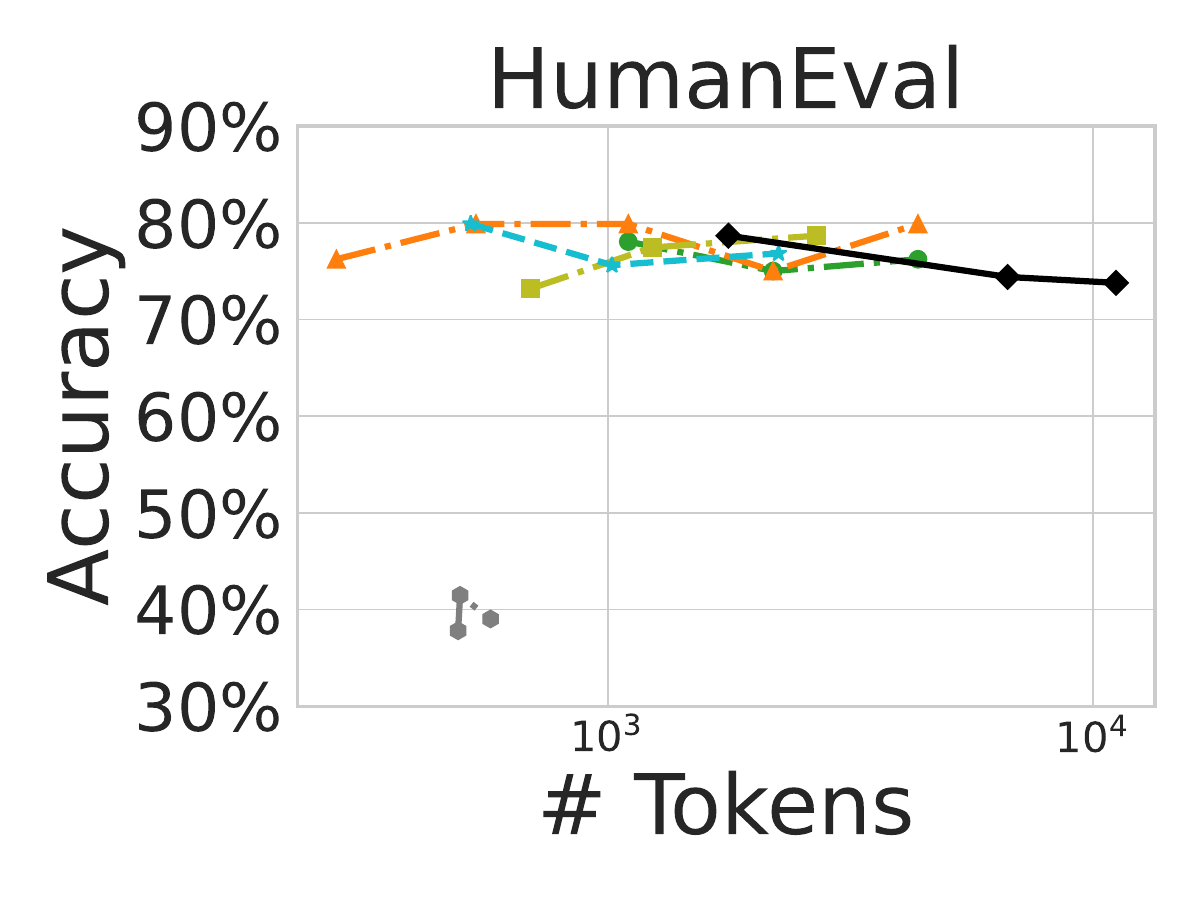}
}
\subfigure{
\includegraphics[width=0.225\textwidth]{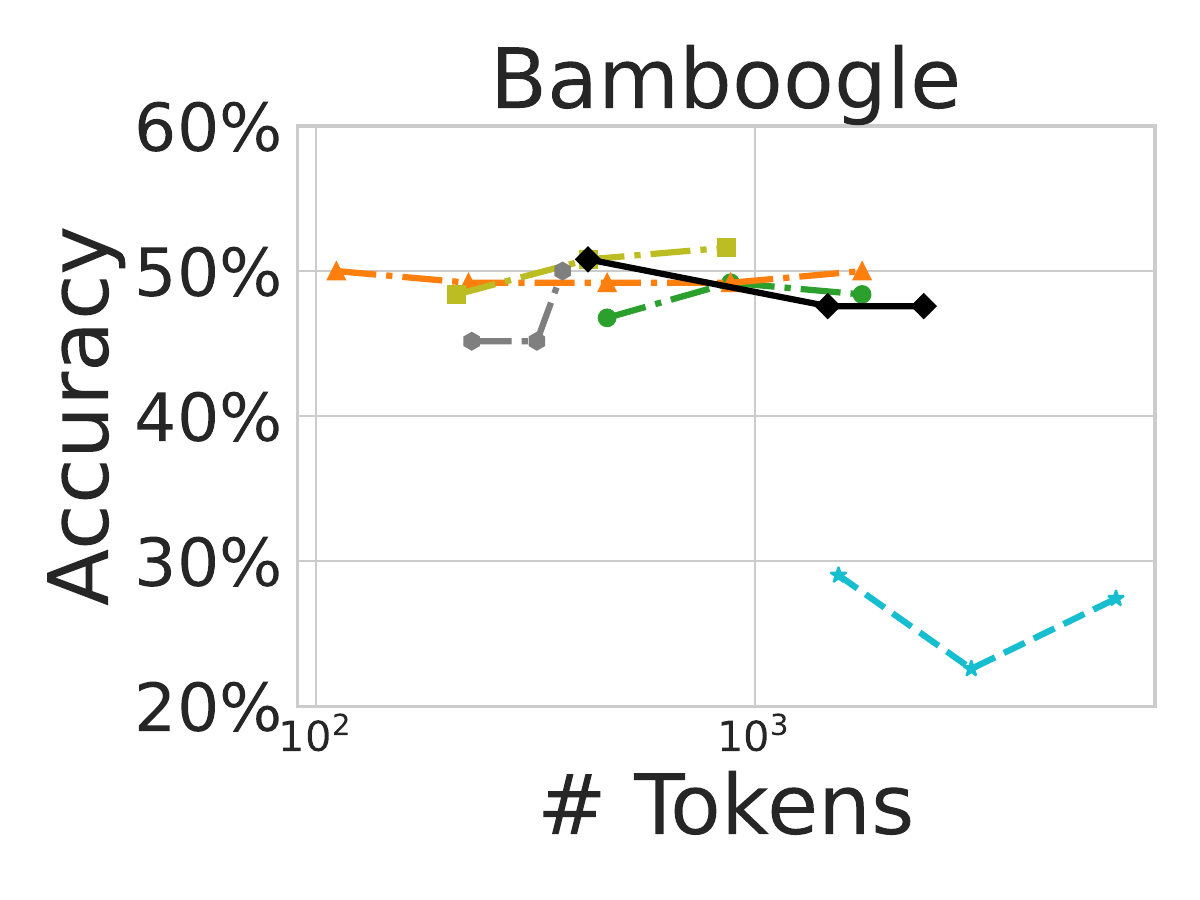}
}
\vspace{-0.2cm}
\caption{Performance versus token consumption across benchmarks for inference-time computation strategies.
}~\label{fig:qwen_token_comsumption}
\vspace{-0.2in}
\end{figure}


\begin{takeaway}[Takeaways]
(1) Instruction prompts significantly influence LLM reasoning, with self-correction yielding mixed results compared to CoT prompting;
(2) A recommended temperature of $\tau = 0.8$ balances diversity and confidence, optimizing reasoning performance, with Top-p performing best at 0.9;
(3) LLMs are not effective at self-evaluation;
(4) Reward models enhance inference-time computation, with process-based rewards excelling in complex tasks like mathematics and code generation;
(5) Reward models can inflate performance evaluations due to generalization issues, leading to inconsistent task effectiveness.
\end{takeaway}

\subsection{Benchmarking of Inference-time computation of LLMs reasoning under Computation Budgets}

 We further examine various inference-time computation methods, including Best-of-N, Step-level Best-of-N, Beam Search, MCTS,  Self-Consistency, and Self-Refine, under a fixed computation budget.

\textbf{Setup.} We evaluate six inference-time computation methods under a fixed computation budget (equal-token results to standardize computational consumption across tasks) on Qwen-2.5-7B and LLaMA-3.3-8B. To ensure fairness, consistent settings are applied to all methods. For knowledge-based reasoning tasks (e.g., Bamboogle, HotpotQA, Fever), we use the RLHF reward model. For complex reasoning tasks (e.g., MATH, code generation), the QwQ-32B model provides the process reward. Implementation details are available in Appendix~\ref{sec:exp_setup}.

Table~\ref{tab:main_result} summarizes the results of six inference-time computation methods across eight reasoning tasks. Key observations include:
(1)  Performance varies significantly across tasks. Best-of-N and Self-Consistency perform well in knowledge-based reasoning tasks, but for complex tasks like MATH and GSM-HARD, Qwen-2.5-7B outperforms Llama-3.1-8B.
(2)  Adding more completion tokens does not always improve accuracy. While Beam Search may leverage higher token counts, the additional tokens often fail to yield proportional accuracy gains.
(3) The Self-Refine method generally underperforms compared to Consistency, suggesting that overly restrictive processes may hinder optimal outputs for complex reasoning tasks.

Figure~\ref{fig:token_vary_llama} illustrates how performance varies with increased computational consumption. As token usage rises, Self-Consistency and Self-Refine achieve higher accuracy than other methods, while Step-level Best-of-N, and MCTS show slower improvements. In contrast, Beam Search exhibits minimal gains, indicating lower token efficiency. Notably, the GSM8K task demonstrates a sharper accuracy increase with higher token consumption compared to MATH500, underscoring the task-specific nature of each method's performance. Additionally, we observe some performance inflation with increased computational consumption, attributed to the generalization limitations of the reward model, which does not perform well on this task.

\vspace{-0.1in}
\section{Conclusion}
\vspace{-0.0in}
In this paper, we investigate the role of inference-time computation in enhancing the reasoning capabilities of LLMs.  Our extensive experimental evaluation reveals that seemingly simple yet overlooked tricks—such as sampling strategies and  reward mechanisms—can yield substantial improvements in reasoning performance. The results demonstrate that careful tuning of inference parameters, such as temperature settings, top-k sampling, reward models, plays a crucial role in optimizing the performance of LLMs during inference time.

\section*{Impact Statement}

This study explores the optimization of inference-time computation strategies to enhance the reasoning abilities of LLMs. By addressing the limitations of existing methods, our findings contribute to improving LLM performance across diverse reasoning tasks, including logical reasoning, code generation, and fact verification. Our work establishes a comprehensive benchmark and demonstrates that previously overlooked techniques—such as temperature and top-p sampling adjustments—can significantly enhance reasoning accuracy. These advancements provide a critical foundation for future research and practical applications, particularly in resource-constrained settings. Optimizing LLMs at inference time without extensive retraining opens new possibilities for improving AI systems' efficiency and reliability in real-world use cases.

\nocite{langley00}

\bibliography{example_paper}
\bibliographystyle{icml2025}


\clearpage
\appendix

\newpage
\onecolumn

\section{Experiments Setup}~\label{sec:exp_setup}
In this section, we provide a more detailed explanation of the experimental setup.
Our objective is to evaluate the impact of previously overlooked techniques on the performance of inference-time computation methods. These methods typically consist of two primary steps: (1) generating candidate solutions using specified parameters (e.g., prompt type, temperature, and top-p), and (2) selecting the optimal solution based on predefined reward signals (e.g., self-evaluation strategies, reward types, and processes).

\textbf{Default Configuration.}
In our default setup, we employ the Best-of-N inference-time computation, where: The number of candidate solutions N is set to 32.
The temperature ($\tau$) is set to 0.7, which introduces moderate stochasticity during candidate generation.
The nucleus sampling parameter, top-p, is configured at 0.9, ensuring diversity by sampling from the top 90\% cumulative probability distribution of the predicted tokens.
We utilize a Chain-of-Thought (CoT) instruction prompt type to facilitate step-by-step reasoning for more complex tasks. Unless explicitly modified, all experiments adhere to this baseline configuration.

\textbf{Reward Model.} Specifically, for process and result rewards, we employ a prompt-driven approach to guide QwQ-32B, a reasoning model, in evaluating candidate solutions. The prompts used are the process evaluation prompt and the result evaluation prompt, which refer to the evaluation of step-level solutions and the final results, respectively. For RLHF and proof-critical rewards, which are numerical reward models, scores are directly assigned to candidate solutions. RLHF Reward: Based on InternLM2-Chat-1.8B-SFT~\cite{cai2024internlm2}, trained on over 2.4 million preference samples. It balances performance, helpfulness, and alignment.
Proof-Critical Reward Derived from the InternLM2-5-Step-Prover-Critic model~\cite{wu2024internlm25stepproveradvancingautomatedtheorem}, which excels in multiple benchmarks.

\begin{tcolorbox}[
    enhanced,
    title=Process-Evaluation.,
    fonttitle=\bfseries,
    coltitle=white, 
    colbacktitle=black!25!gray, 
    colback=white, 
    colframe=black, 
    boxrule=0.2mm, 
    arc=3mm, 
    toptitle=0.3mm, 
    bottomtitle=0.3mm, 
    left=3mm, 
    right=3mm 
]

Evaluate whether the language model can decompose the question into relevant sub-questions and assess whether this decomposition aids in answering the original question, either partially or directly. The evaluation result will be classified as "Sure," "Likely," or "Impossible" based on the effectiveness of the decomposition.

Evaluation Process: 1. Relevance of Sub-Questions: Determine if the sub-questions are directly related to solving the original question.
2. Effectiveness of Decomposition: Assess whether the sub-questions, when answered, lead to a comprehensive response to the original question.

Evaluation Outcomes: Sure: The model successfully decomposes the question into relevant sub-questions, each structured to contribute to an accurate final answer.Likely: The model decomposes the question into relevant sub-questions, but minor improvements in structure or relevance may enhance the response. Impossible: The model fails to decompose the question effectively or the sub-questions are not relevant to the original question.

\end{tcolorbox}

\begin{tcolorbox}[
    enhanced,
    title=Result-Evaluation.,
    fonttitle=\bfseries,
    coltitle=white, 
    colbacktitle=black!25!gray, 
    colback=white, 
    colframe=black, 
    boxrule=0.2mm, 
    arc=3mm, 
    toptitle=0.3mm, 
    bottomtitle=0.3mm, 
    left=3mm, 
    right=3mm 
]

The Result-Evaluation prompt evaluates the final outcome of the language model's reasoning process based on the accuracy, clarity, and completeness of its answer. Each evaluation is categorized as "Sure," "Likely," or "Impossible." The result is judged by verifying whether the language model's final answer is both directly relevant and valid in response to the question posed. The process involves reviewing whether the model’s conclusion is definitive and supported by logical reasoning. If the answer is clear and unambiguous, it is marked as "Sure." If there are minor ambiguities, it is categorized as "Likely." If the answer is incorrect or irrelevant, it is deemed "Impossible."
\end{tcolorbox}

\textbf{Tasks.}  
Our research focuses on the following reasoning tasks:
(1) \textbf{Arithmetic Reasoning:}  
\textbf{GSM8K:} A dataset with grade school math word problems requiring 2 to 8 calculation steps using basic arithmetic operations. 
\textbf{GSM-Hard:} A harder variant of GSM8K with larger, less common numbers, increasing the challenge of arithmetic reasoning.
(2) \textbf{Complex Mathematical Reasoning:}  
\textbf{MATH Dataset:} A dataset covering advanced topics like algebra, geometry, and calculus. Each problem includes detailed solutions to evaluate both final answers and problem-solving processes. We use the MATH 500 to test the complex mathematical reasoning ability of LLM.
(3) \textbf{Logical Reasoning:}  
\textbf{ProntoQA:} Measures logical deduction and inference, requiring the application of logical principles to reach correct conclusions.
(4) \textbf{Code Generation:}  
\textbf{HumanEval:} A benchmark for generating functional code snippets, with programming problems that include prompts and test cases to verify correctness.
(5) \textbf{Question Answering:}  
\textbf{Bamboogle:} Evaluates diverse question-answering performance across various topics, testing comprehension and accurate responses.
(6) \textbf{Fact Verification:}  
\textbf{FEVER:} Assesses the ability to verify factual claims using a document corpus, promoting fact-checking system development.
(7) \textbf{Common Sense Reasoning:}  \textbf{HotpotQA:} Features multi-hop questions requiring reasoning across multiple facts, testing common sense knowledge and the ability to link disparate information.

\textbf{Instruction Prompt Type.}
Different instruction prompts can guide an LLM to generate distinct reasoning paths. Specifically, Input-Output (IO) prompts directly provide the answer, whereas Chain-of-Thought (CoT) prompts encourage the LLM to reason step by step. Recent research \cite{huang2023large} suggests that self-correction or self-reflection mechanisms are often ineffective when LLMs operate without external feedback under certain prompt types. We further explores the impact of various prompt types, including IO prompts, standard Chain-of-Thought (CoT) prompts, and reflection-based CoT. 
\begin{tcolorbox}[
    enhanced,
    title=Input-Output (IO) Prompt.,
    fonttitle=\bfseries,
    coltitle=white, 
    colbacktitle=black!25!gray, 
    colback=white, 
    colframe=black, 
    boxrule=0.2mm, 
    arc=3mm, 
    toptitle=0.3mm, 
    bottomtitle=0.3mm, 
    left=3mm, 
    right=3mm 
]

The Input-Output (IO) Prompt directly answers a given question without intermediate reasoning steps. It is designed to generate concise and accurate responses, ending with the phrase \textit{"so the final answer is:"}. This prompt structure is particularly suitable for straightforward queries where minimal context or explanation is required.

\end{tcolorbox}

\begin{tcolorbox}[
    enhanced,
    title=Chain-of-Thought (CoT) Prompt.,
    fonttitle=\bfseries,
    coltitle=white, 
    colbacktitle=black!25!gray, 
    colback=white, 
    colframe=black, 
    boxrule=0.2mm, 
    arc=3mm, 
    toptitle=0.3mm, 
    bottomtitle=0.3mm, 
    left=3mm, 
    right=3mm 
]

The Chain-of-Thought (CoT) Prompt guides the model to answer questions step-by-step, breaking down the reasoning process into intermediate steps before concluding with the final answer. For instance, when comparing the lifespans of Theodor Haecker and Harry Vaughan Watkins, the CoT prompt explicitly calculates their ages step-by-step before determining the longer-lived individual. This structured approach enhances reasoning transparency and aligns the response with logical steps.

\end{tcolorbox}

\begin{tcolorbox}[
    enhanced,
    title=Reflect CoT.,
    fonttitle=\bfseries,
    coltitle=white, 
    colbacktitle=black!25!gray, 
    colback=white, 
    colframe=black, 
    boxrule=0.2mm, 
    arc=3mm, 
    toptitle=0.3mm, 
    bottomtitle=0.3mm, 
    left=3mm, 
    right=3mm 
]

The Reflect Chain-of-Thought (Reflect CoT) prompt introduces a structured reasoning approach where each step in answering a question is followed by a reflection to verify its accuracy and reliability. For example, when comparing the lifespans of Theodor Haecker and Harry Vaughan Watkins, the process involves step-by-step reasoning to establish each individual’s age at death. After each step, a "Reflection" line is used to ensure the validity of the information, such as verifying directly provided ages or confirming the consistency of the comparison. The final conclusion, supported by reflections, ensures a reliable and transparent reasoning process.

\end{tcolorbox}

\section{Other Experiments}~\label{sec:other_exp}
\begin{figure}[t]
\vspace{-0.1cm}
\centering
\includegraphics[width=0.4\textwidth]{figure/abalation/figure/self_prompt/legend.pdf}

\subfigure{
\vspace{-0.8cm}
\includegraphics[width=0.225\textwidth]{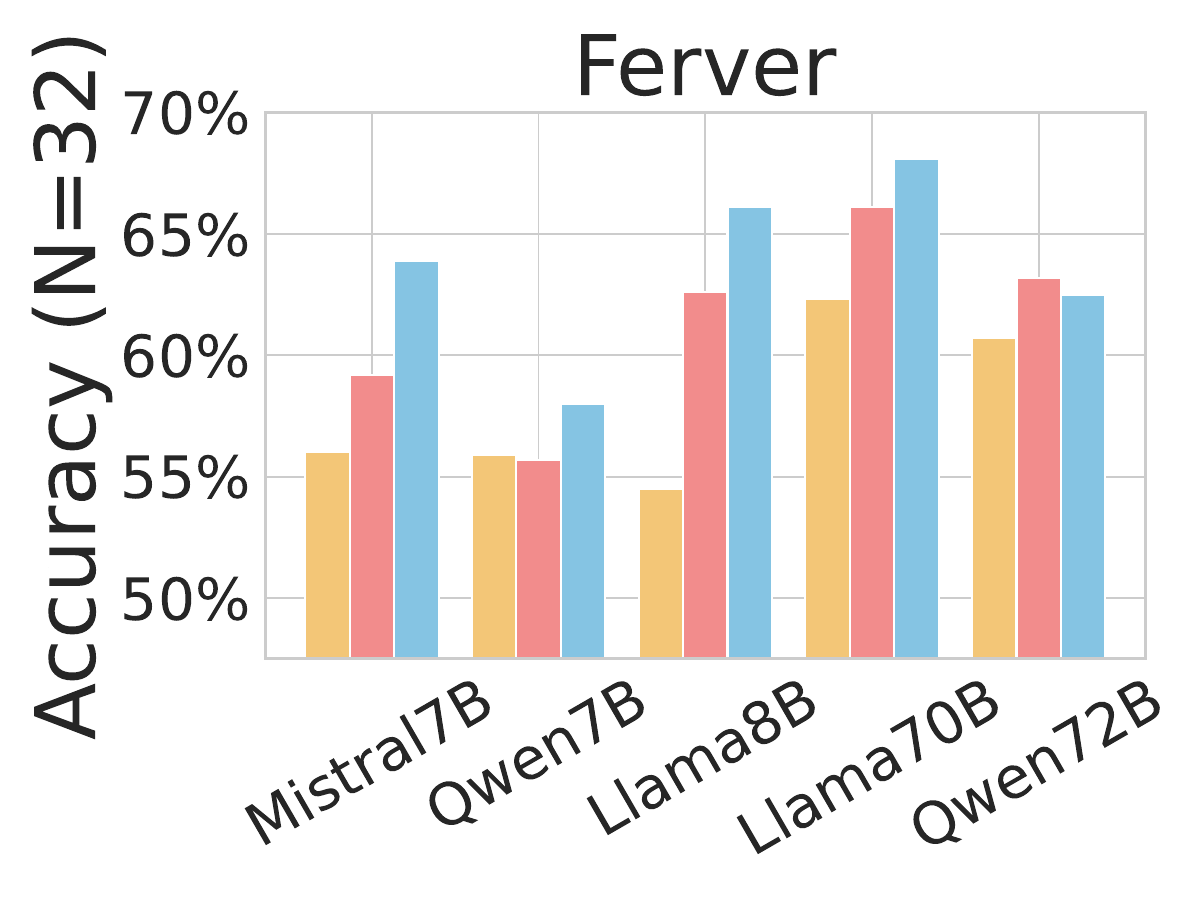}
}
\subfigure{
\includegraphics[width=0.225\textwidth]{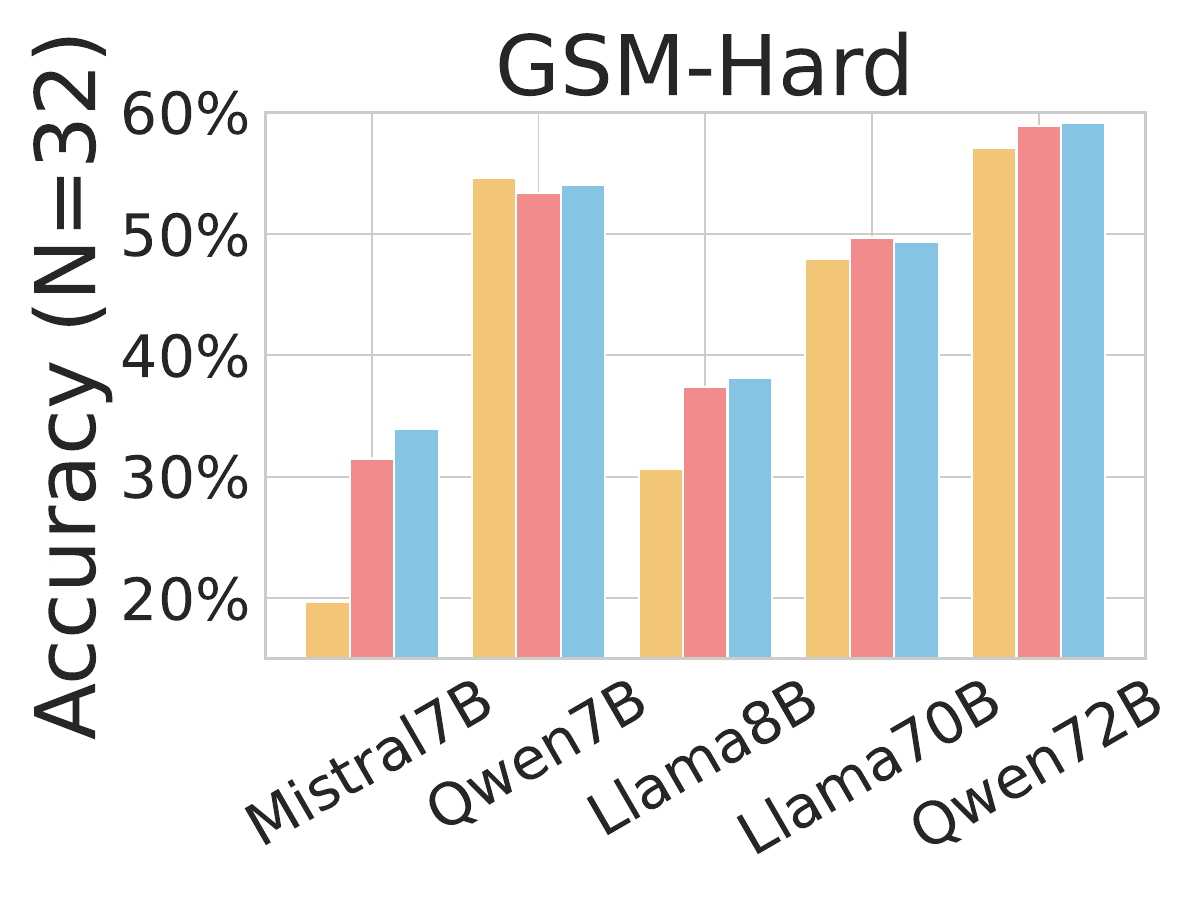}
}
\vspace{-0.4cm}  
\subfigure{
\includegraphics[width=0.225\textwidth]{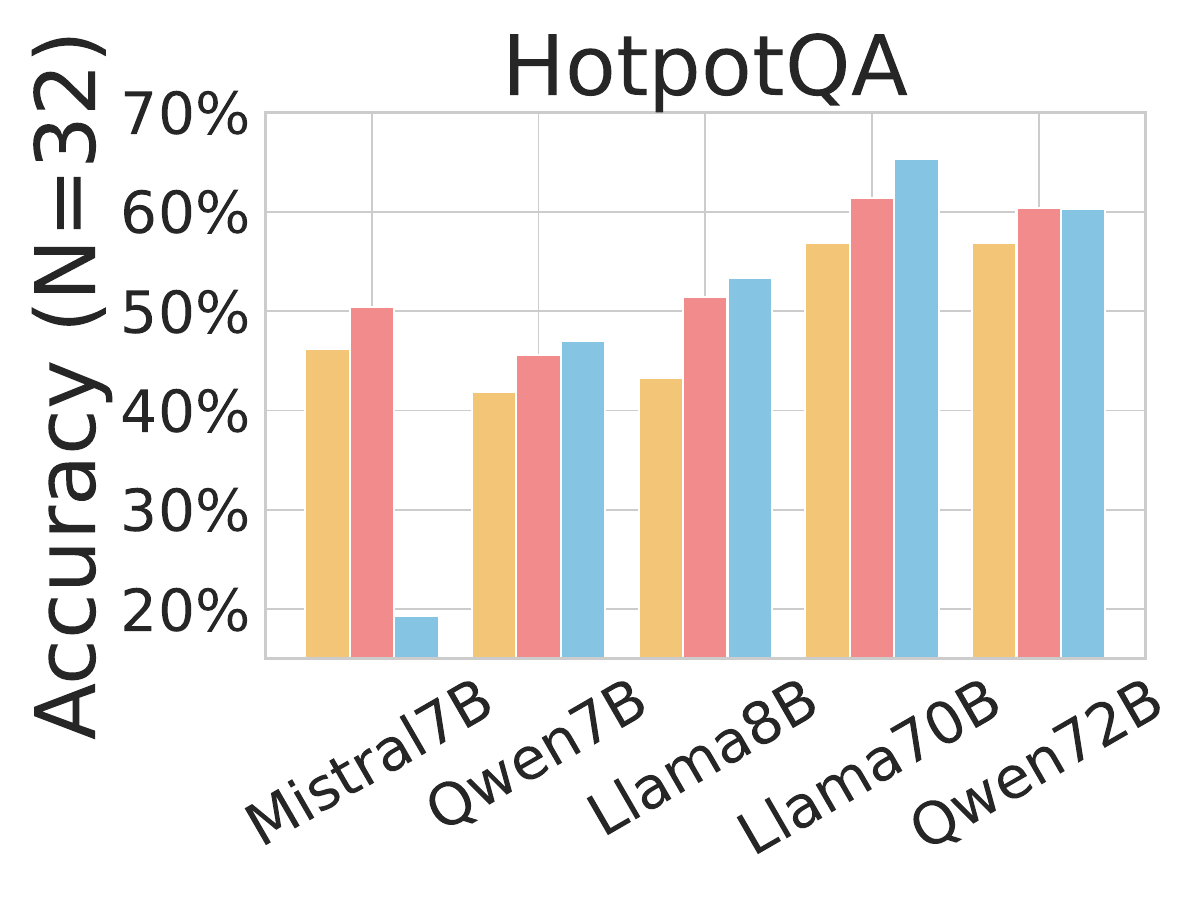}
}
\subfigure{
\includegraphics[width=0.225\textwidth]{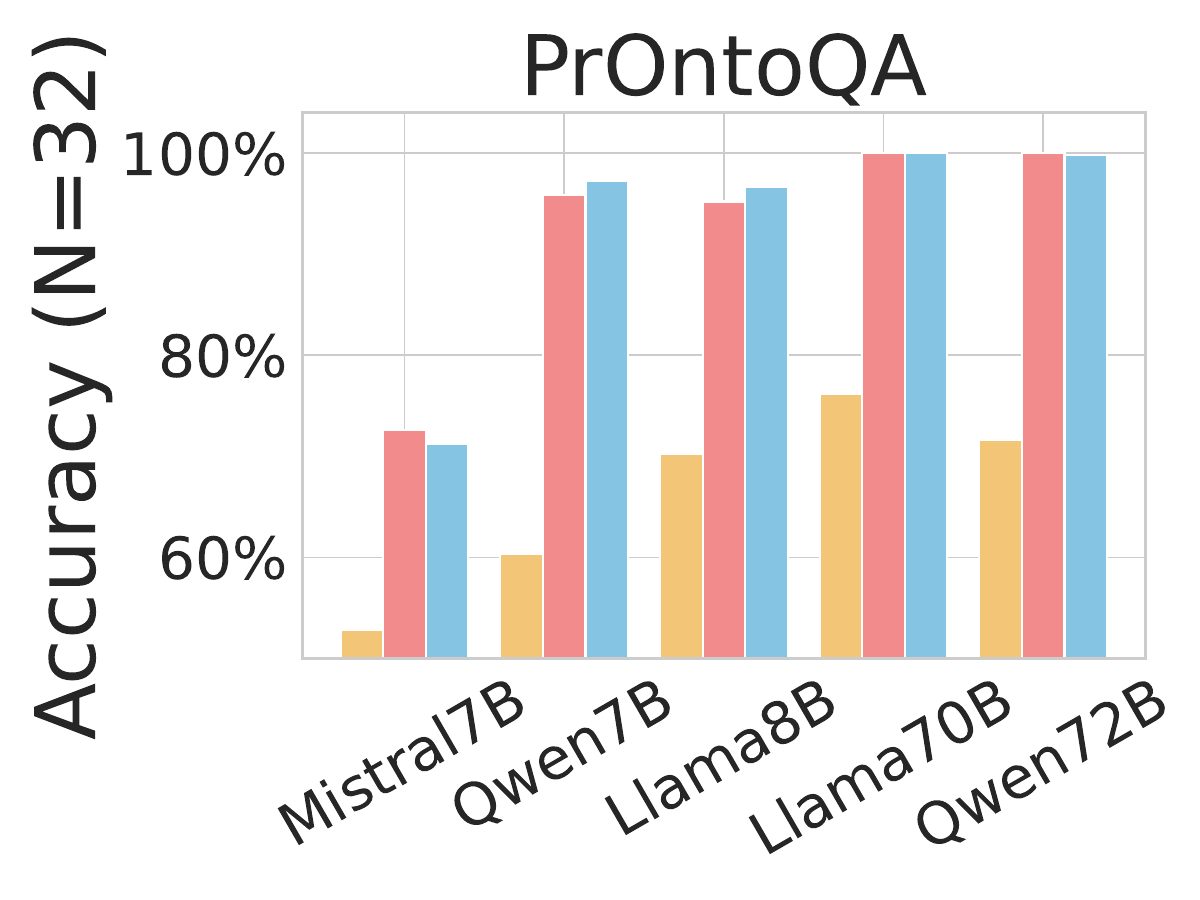}
}
\vspace{-0.2cm}
\caption{ Comparative accuracy of different models (Mistral7B, Qwen7B, Llama88, Llama70B, and Qwen72B) across four datasets (Ferver, GSM-Hard, HotpotQA, and PrOntoQA) using three instruction prompts: Input-Output (IO), Chain-of-Thought (CoT), and Reflect Chain-of-Thought (Reflect CoT).
}
\vspace{-0.3cm}
\label{fig:prompt_other_data}
\end{figure}

\textbf{Instruction Prompt Type.}
The experimental results on additional datasets reveal a consistent finding: the type of instruction prompt significantly influences the inference-time computation of LLM reasoning, as illustrated in Figure~\ref{fig:prompt_other_data}.

\begin{figure}[t]
\vspace{-0.1cm}
\centering
\includegraphics[width=0.6\textwidth]{figure/abalation/figure/temp/legend.pdf}

\subfigure{
\vspace{-0.8cm}
\includegraphics[width=0.225\textwidth]{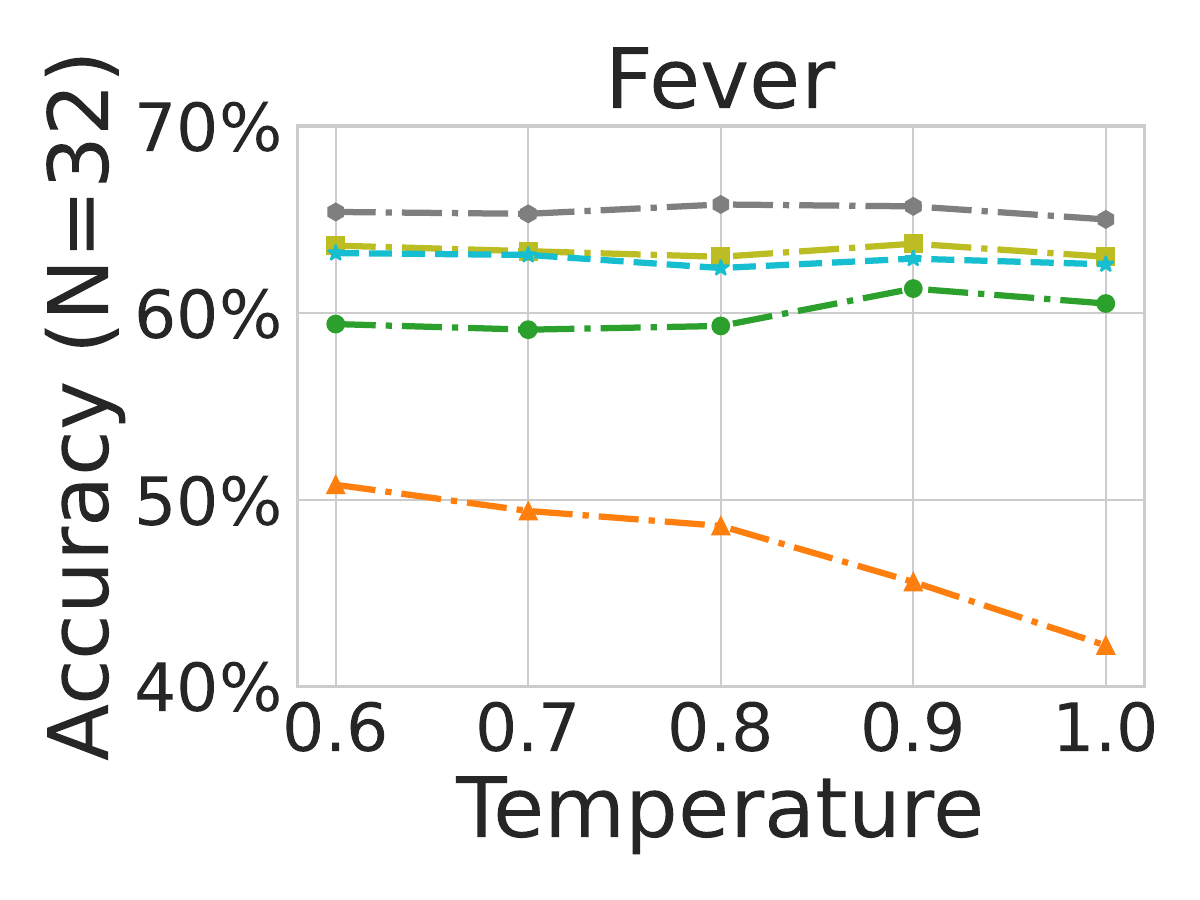}
}
\subfigure{
\includegraphics[width=0.225\textwidth]{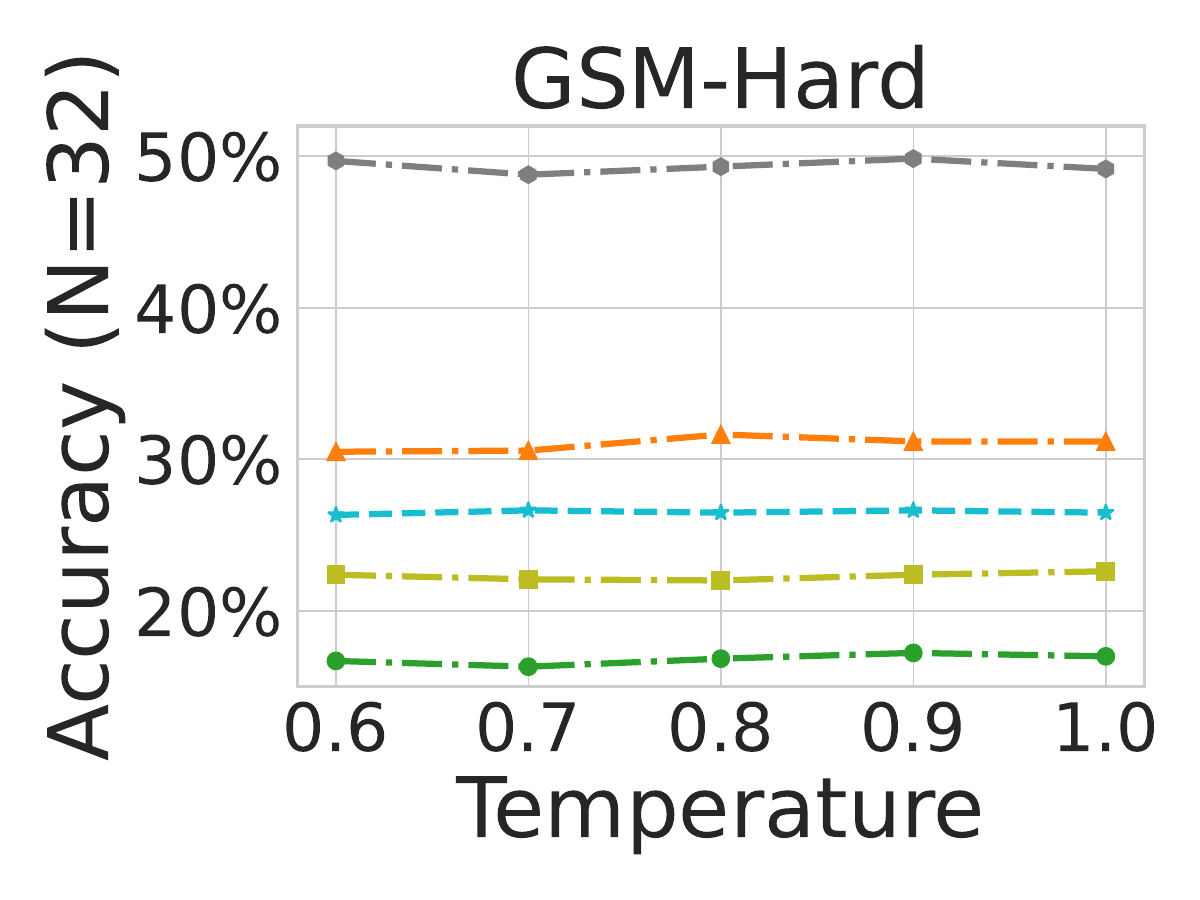}
}
\vspace{-0.4cm}   
\subfigure{
\includegraphics[width=0.225\textwidth]{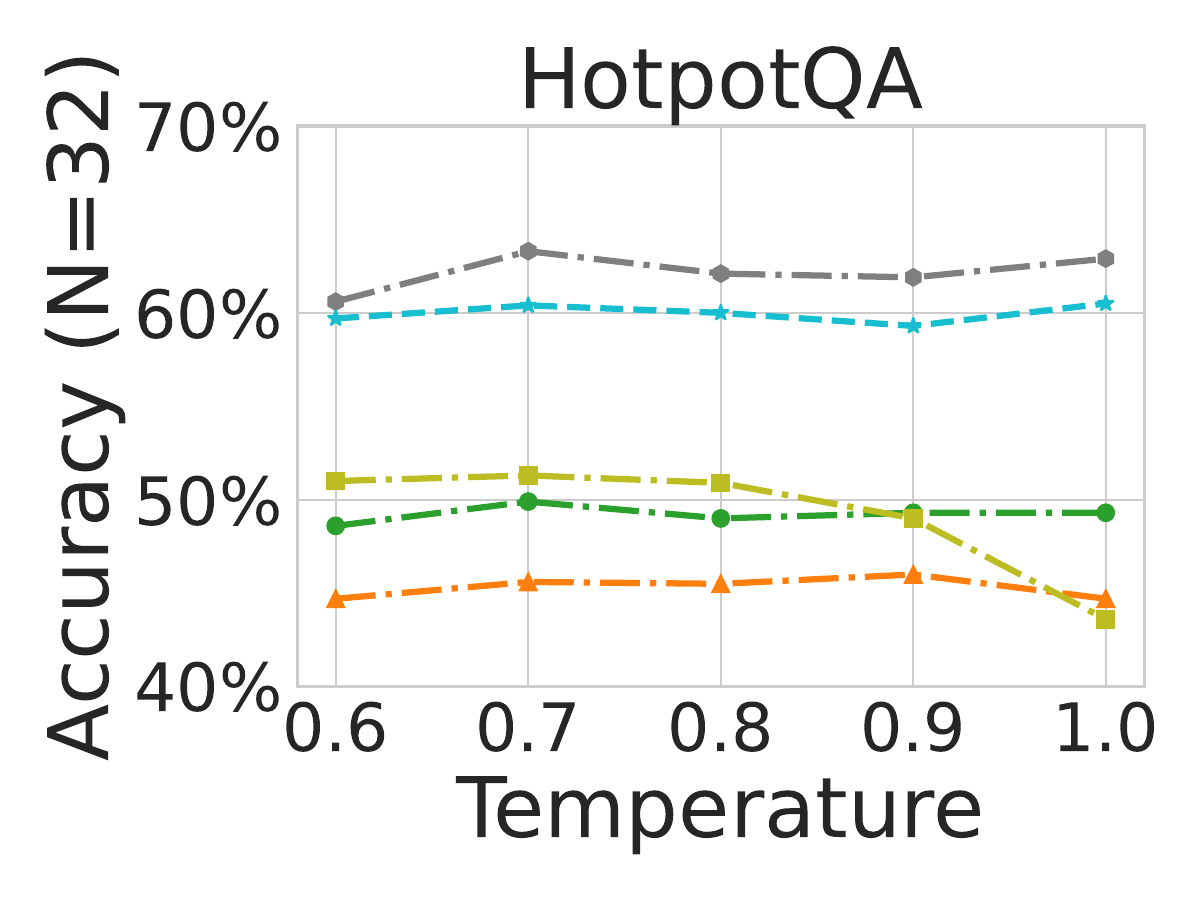}
}
\subfigure{
\includegraphics[width=0.225\textwidth]{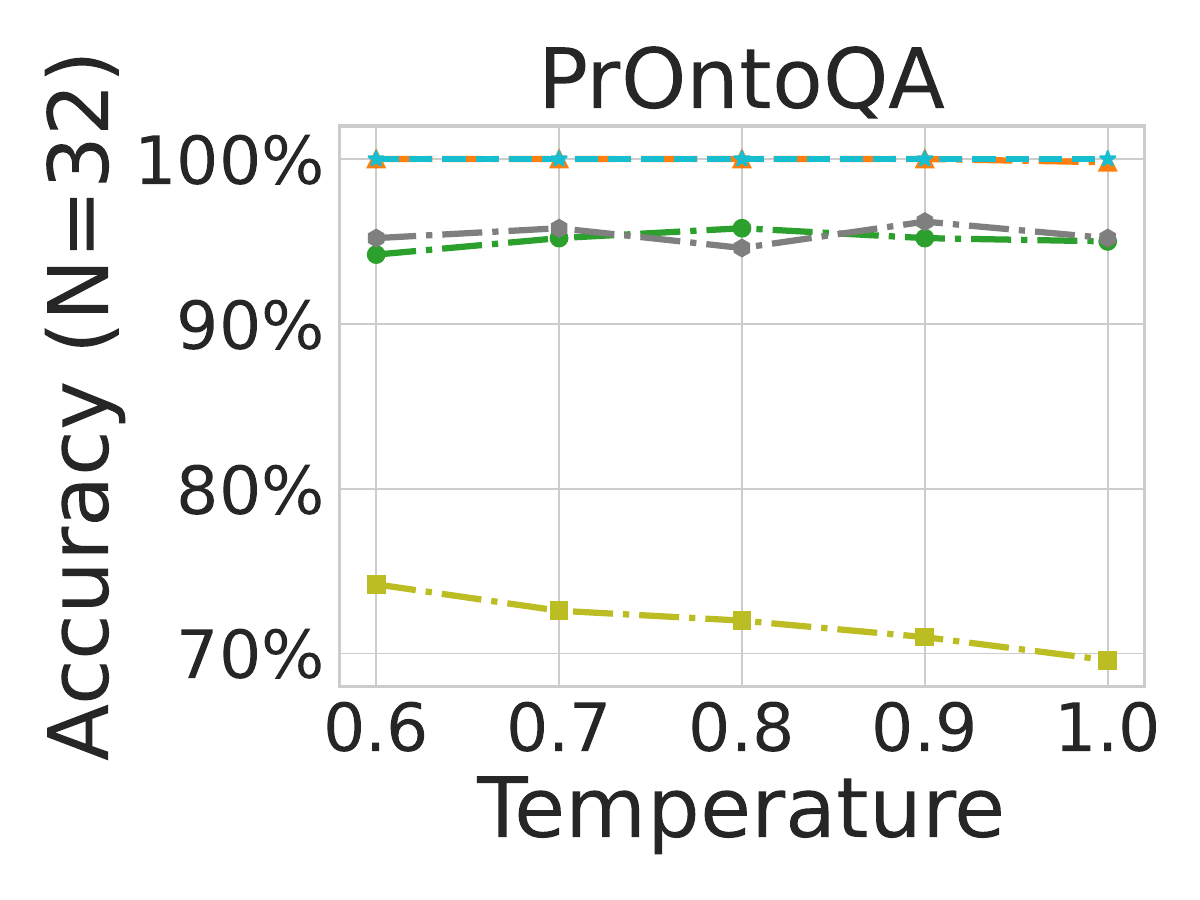}
}
\vspace{-0.2cm}
\caption{ Accuracy vs. temperature for different models (Mistral7B, Qwen7B, Llama88, Llama70B, and Qwen72B) across four datasets (Ferver, GSM-Hard, HotpotQA, and PrOntoQA), with temperature settings ranging from 0.6 to 1.0.
}
\vspace{-0.3cm}
\label{fig:temp_other_data}
\end{figure}

\textbf{Tempature.}
The experimental results on additional datasets reveal a consistent finding: the type of instruction prompt significantly influences the inference-time computation of LLM reasoning, as illustrated in Figure~\ref{fig:temp_other_data}.

\begin{figure}[t]
\vspace{-0.1cm}
\centering
\includegraphics[width=0.6\textwidth]{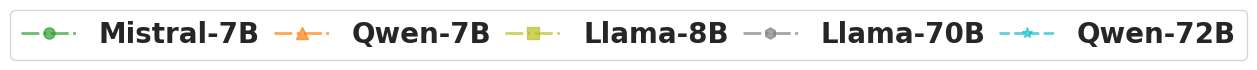}

\subfigure{
\vspace{-0.8cm}
\includegraphics[width=0.225\textwidth]{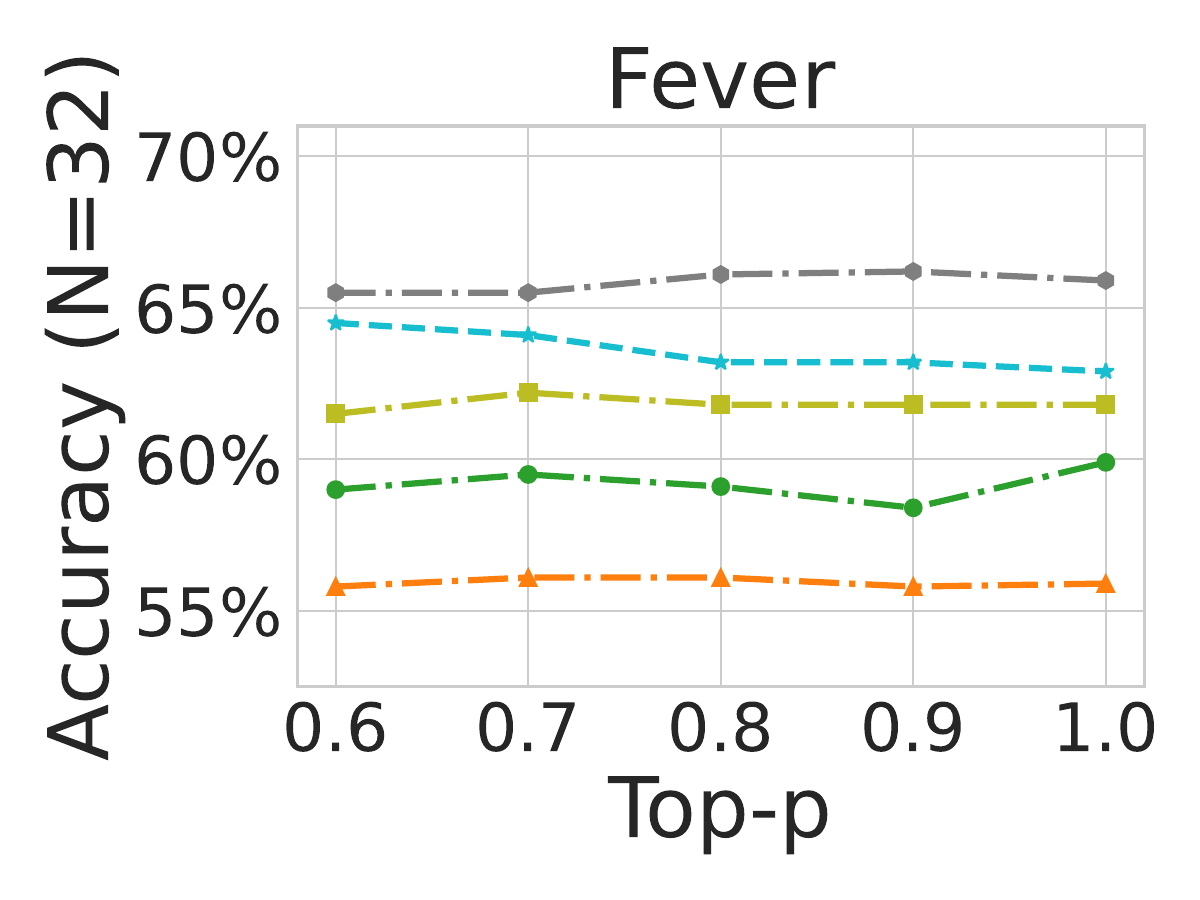}
}
\subfigure{
\includegraphics[width=0.225\textwidth]{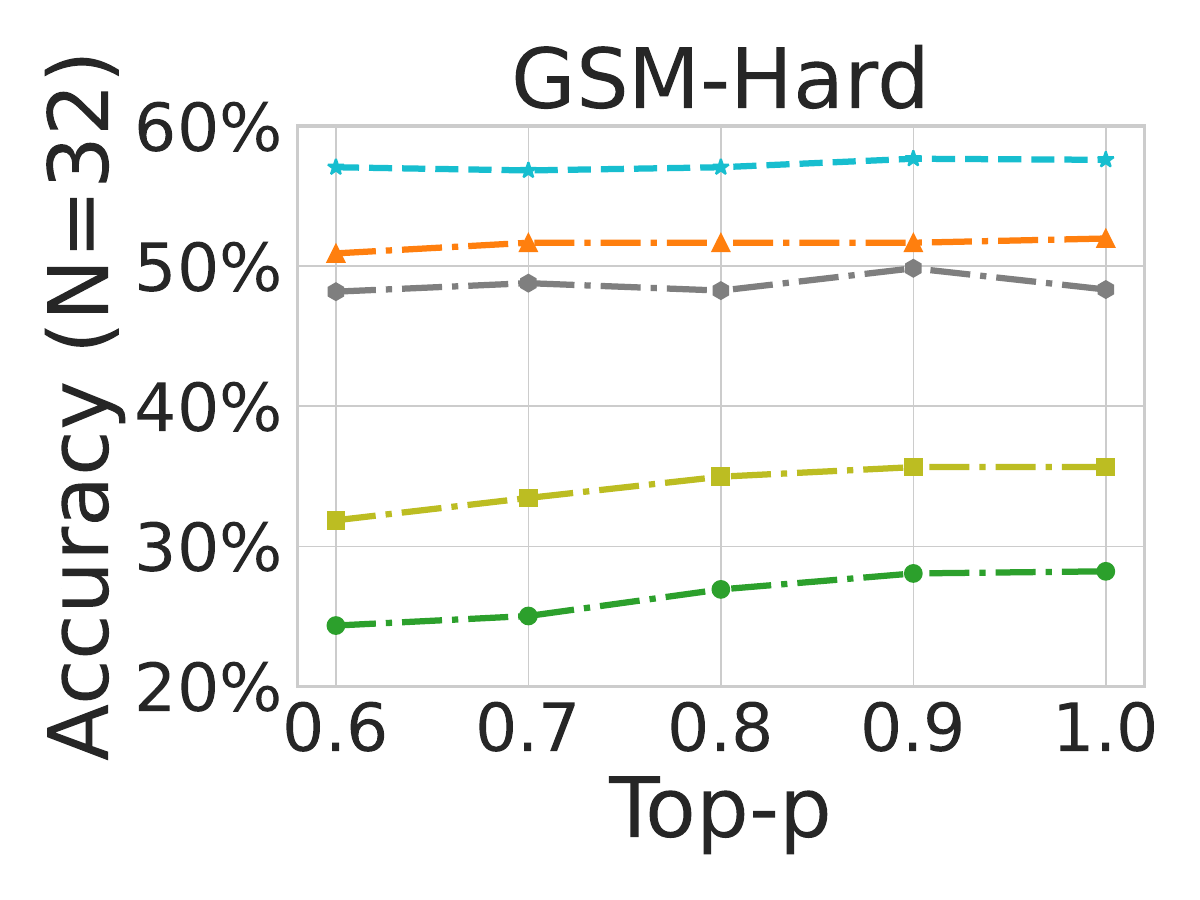}
}
\vspace{-0.4cm}  
\subfigure{
\includegraphics[width=0.225\textwidth]{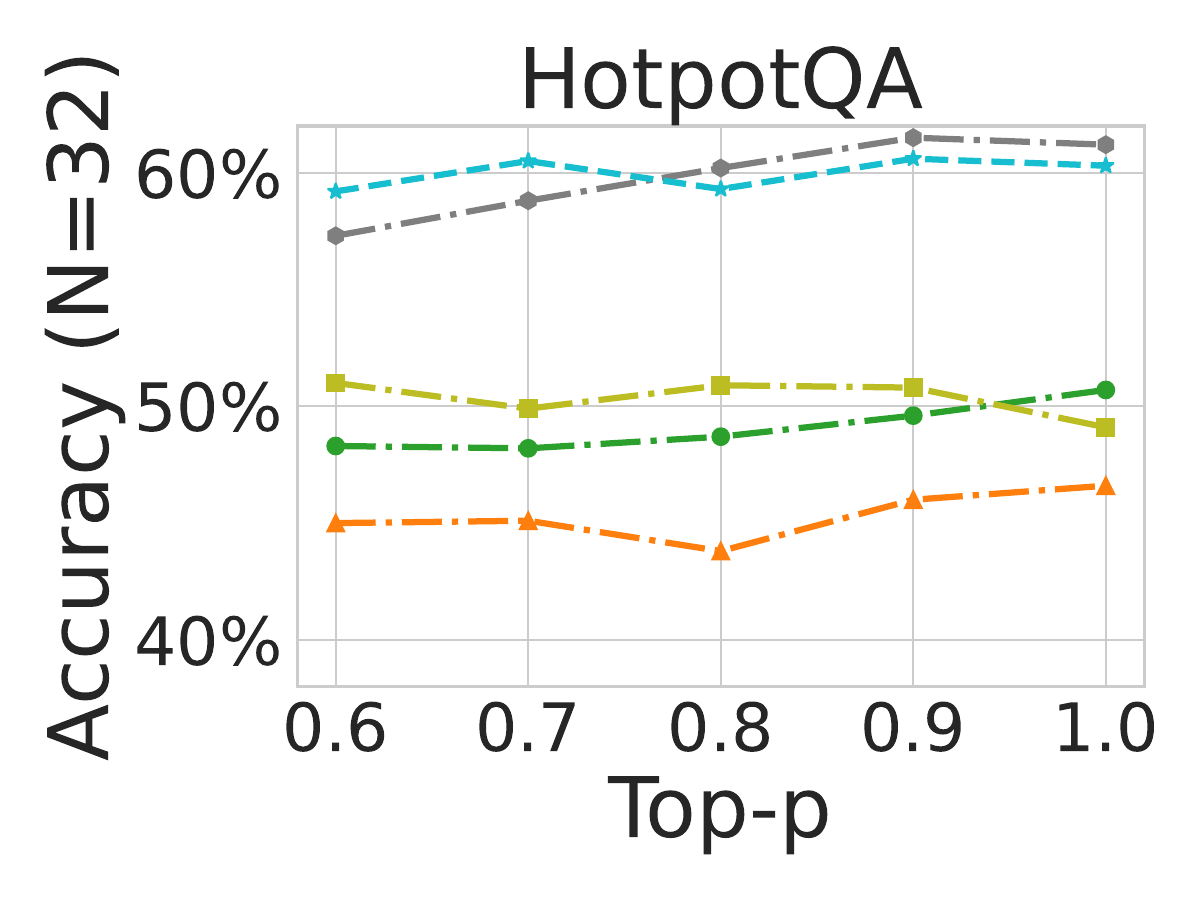}
}
\subfigure{
\includegraphics[width=0.225\textwidth]{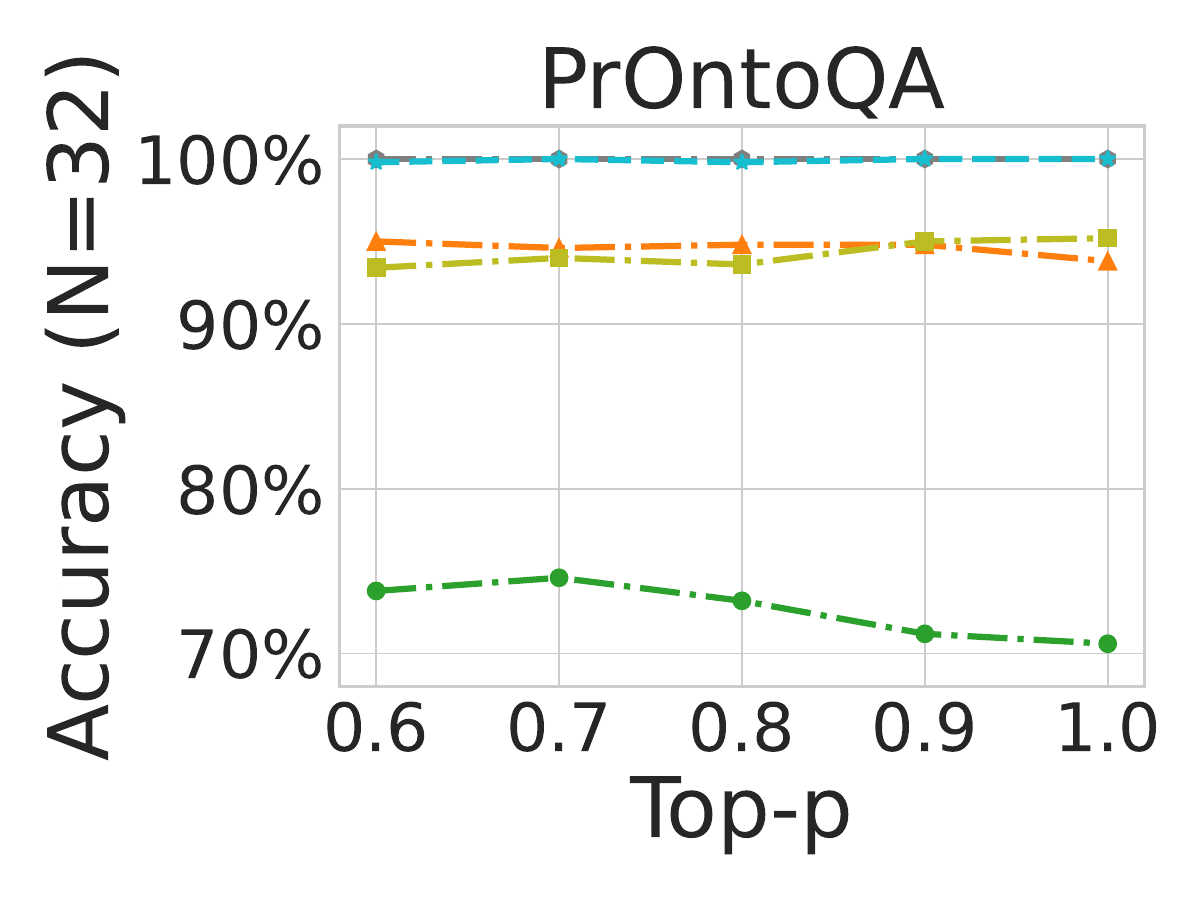}
}
\vspace{-0.2cm}
\caption{ Accuracy vs. Top-p for different models (Mistral7B, Qwen7B, Llama88, Llama70B, and Qwen72B) across four datasets (Ferver, GSM-Hard, HotpotQA, and PrOntoQA), with Top-p values ranging from 0.6 to 1.0. 
}
\vspace{-0.3cm}
\label{fig:top_other_data}
\end{figure}

\textbf{Top-p.} We further present the ablation study on top-p applied to other reasoning tasks. The experimental results are shown in Figure~\ref{fig:top_other_data}. The impact of top-p is significant; generally, as top-p increases, the LLM's reasoning performance also improves, with optimal performance observed at 0.9 for most reasoning tasks.

\begin{figure}[t]
\centering
\includegraphics[width=0.6\textwidth]{figure/abalation/figure/evaluation_type/legend.pdf}
\\
\subfigure{
\includegraphics[width=0.225\textwidth]{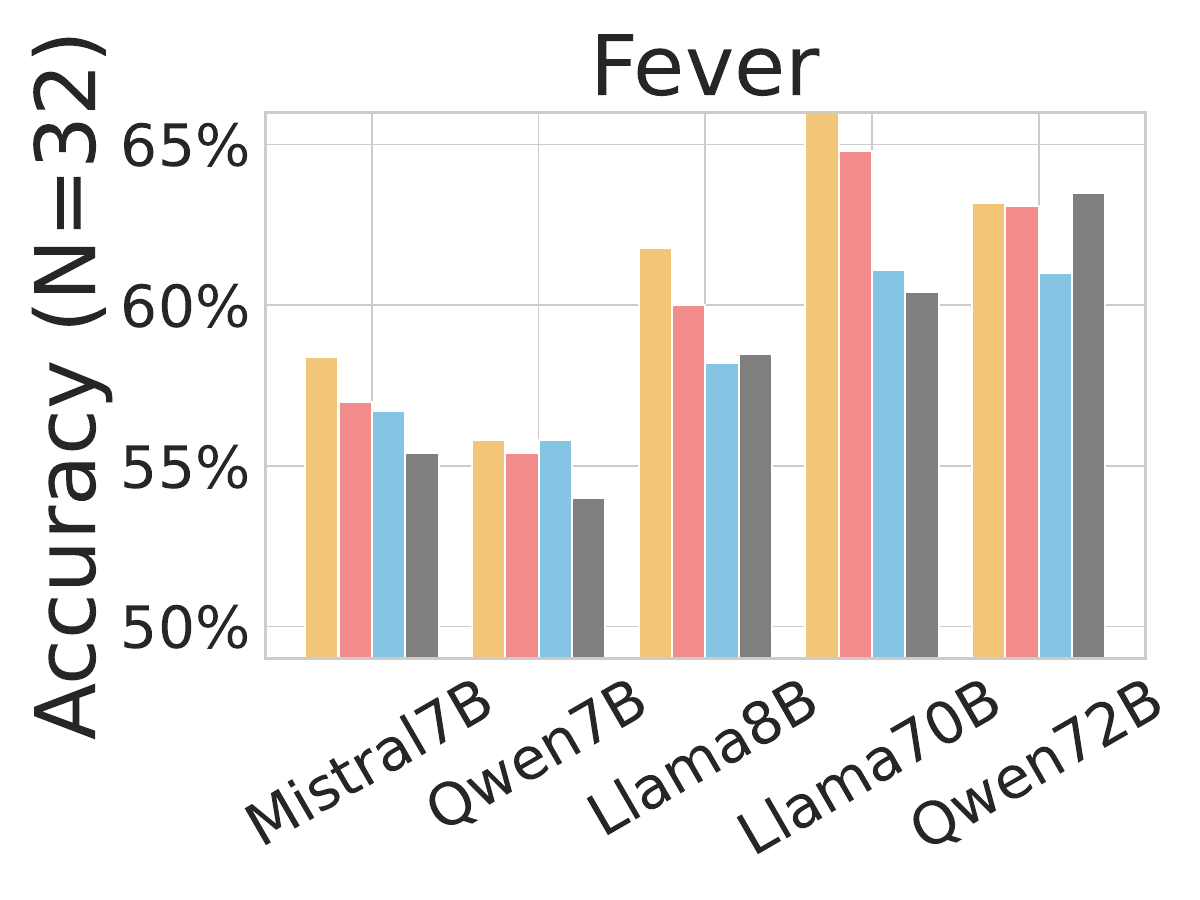}
}
\subfigure{
\includegraphics[width=0.225\textwidth]{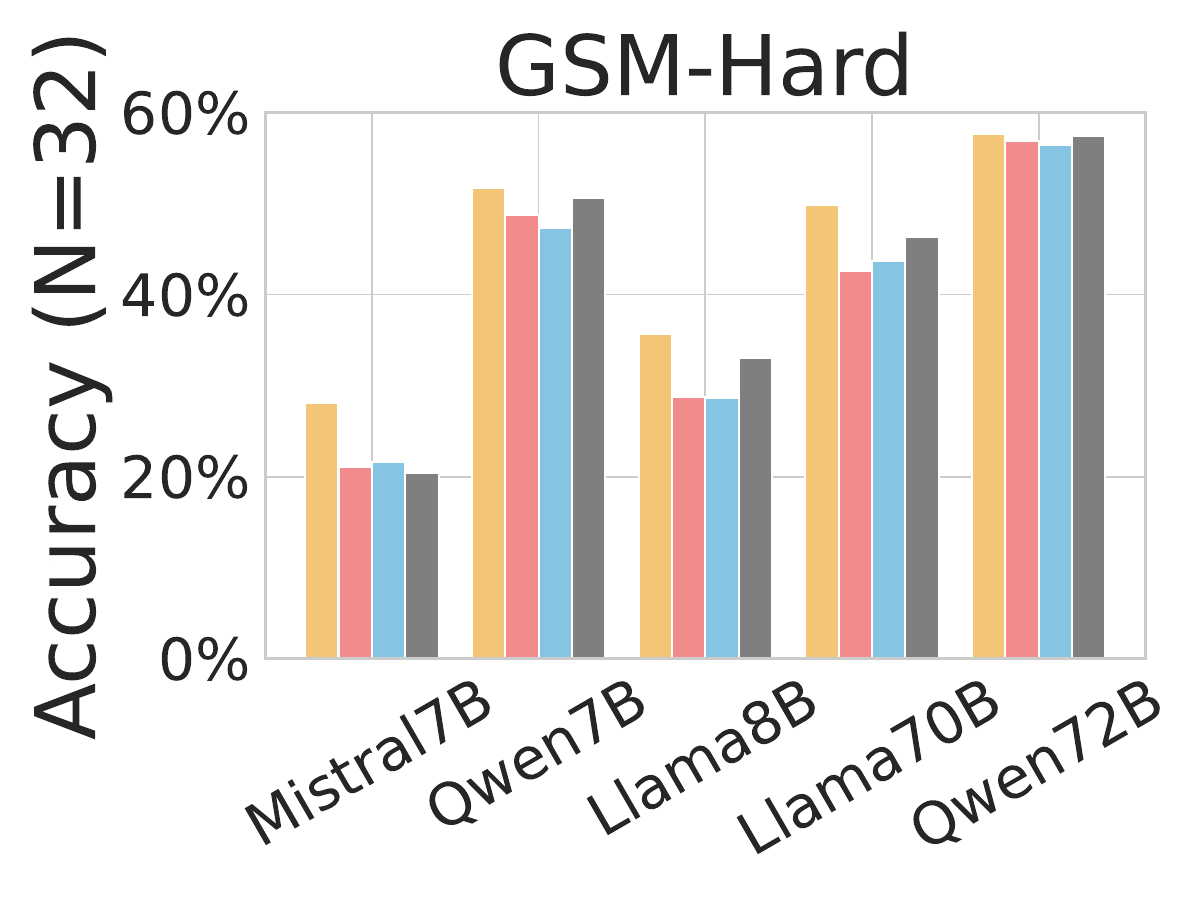}
}   
\subfigure{
\includegraphics[width=0.225\textwidth]{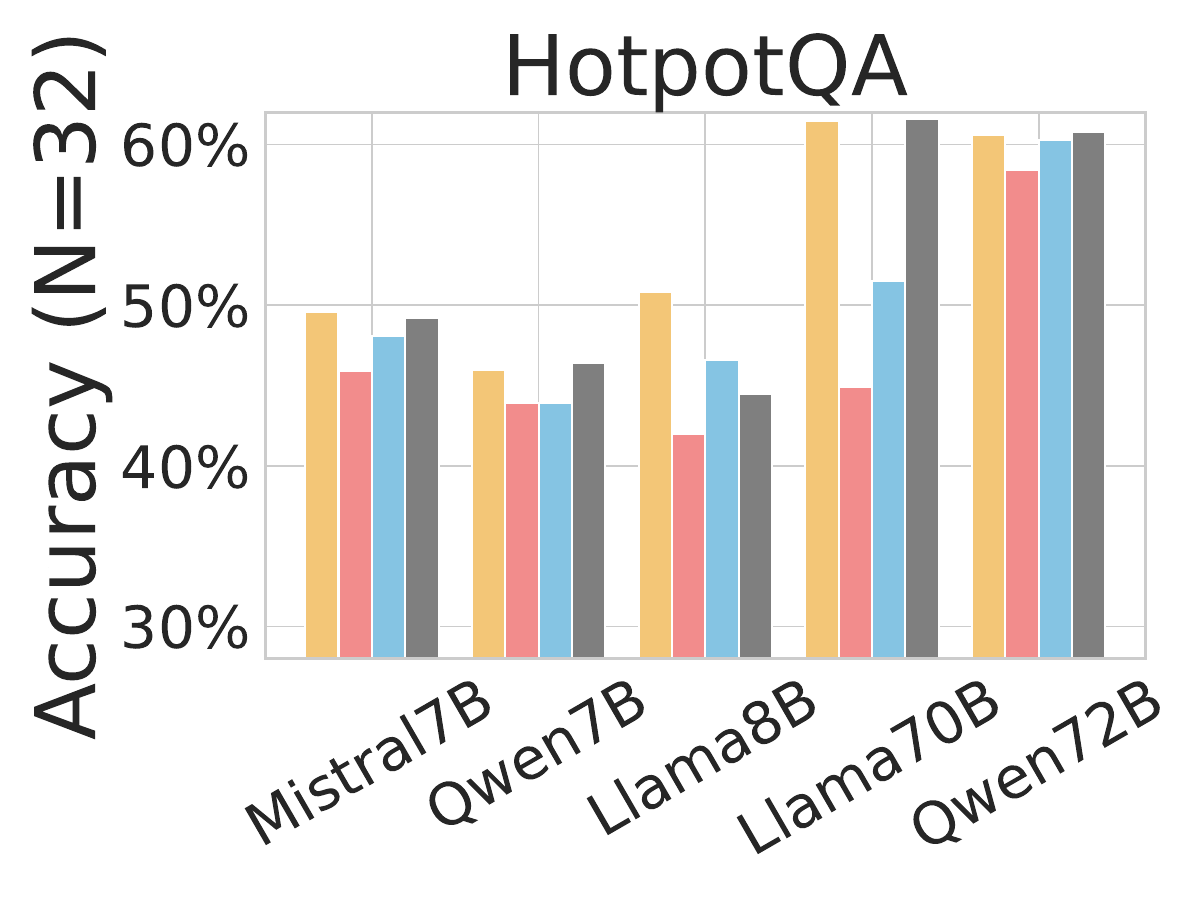}
}
\subfigure{
\includegraphics[width=0.225\textwidth]{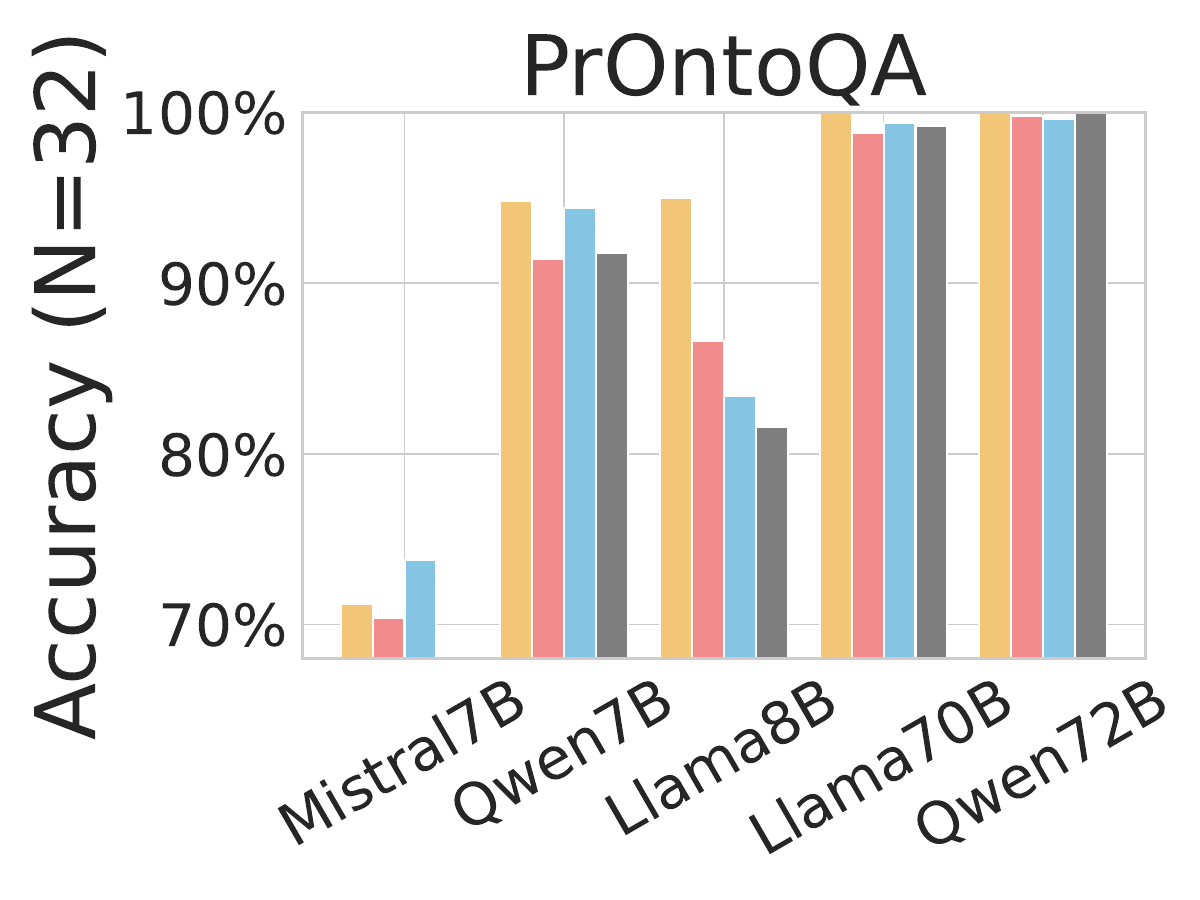}
}
\caption{ Accuracy (\%) across four benchmark tasks  with different evaluation strategies  The results shows that Self-evaluation often fails to assess solution quality effectively.
}~\label{fig:trick_self_evalaution_othe_data}
\vspace{-0.3in}
\end{figure}

\textbf{Self-Evaluation.} We further present the self-evaluation conducted on additional datasets. Figure~\ref{fig:trick_self_evalaution_othe_data} illustrates the experimental results, which reveal that the LLM still cannot effectively distinguish the correct solution.

\begin{figure}[t]
\centering
\includegraphics[width=0.6\textwidth]{figure/abalation/figure/reward_type/legend.pdf}\\
\subfigure{
\includegraphics[width=0.225\textwidth]{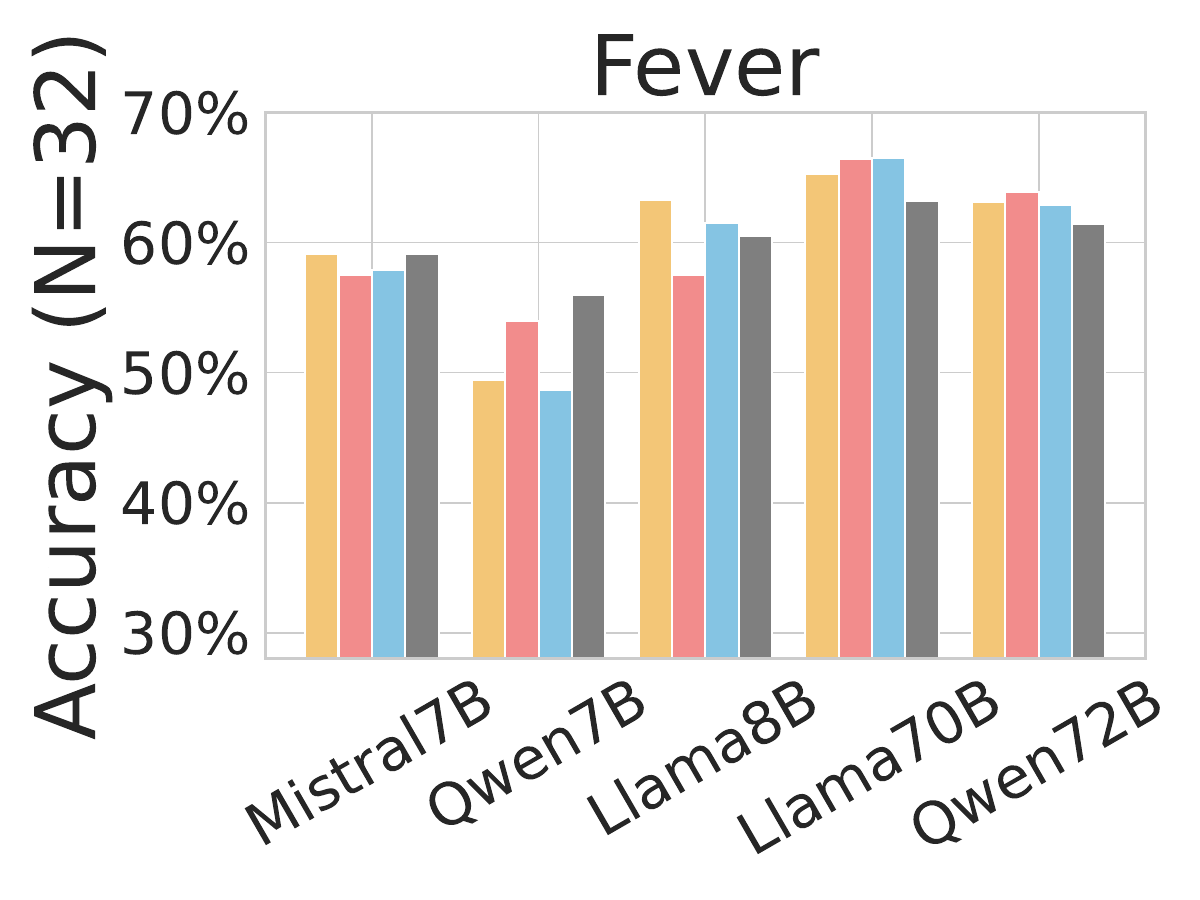}
}
\subfigure{
\includegraphics[width=0.225\textwidth]{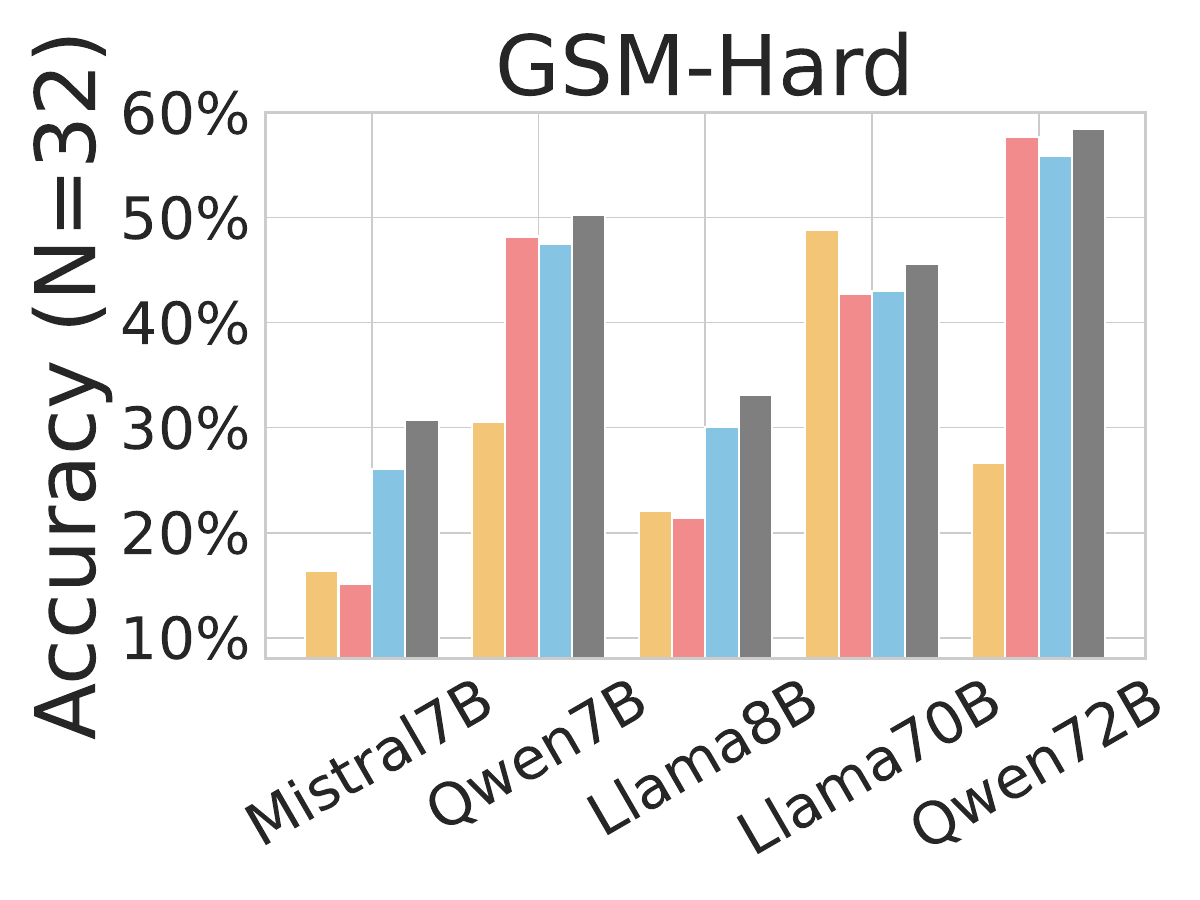}
}   
\subfigure{
\includegraphics[width=0.225\textwidth]{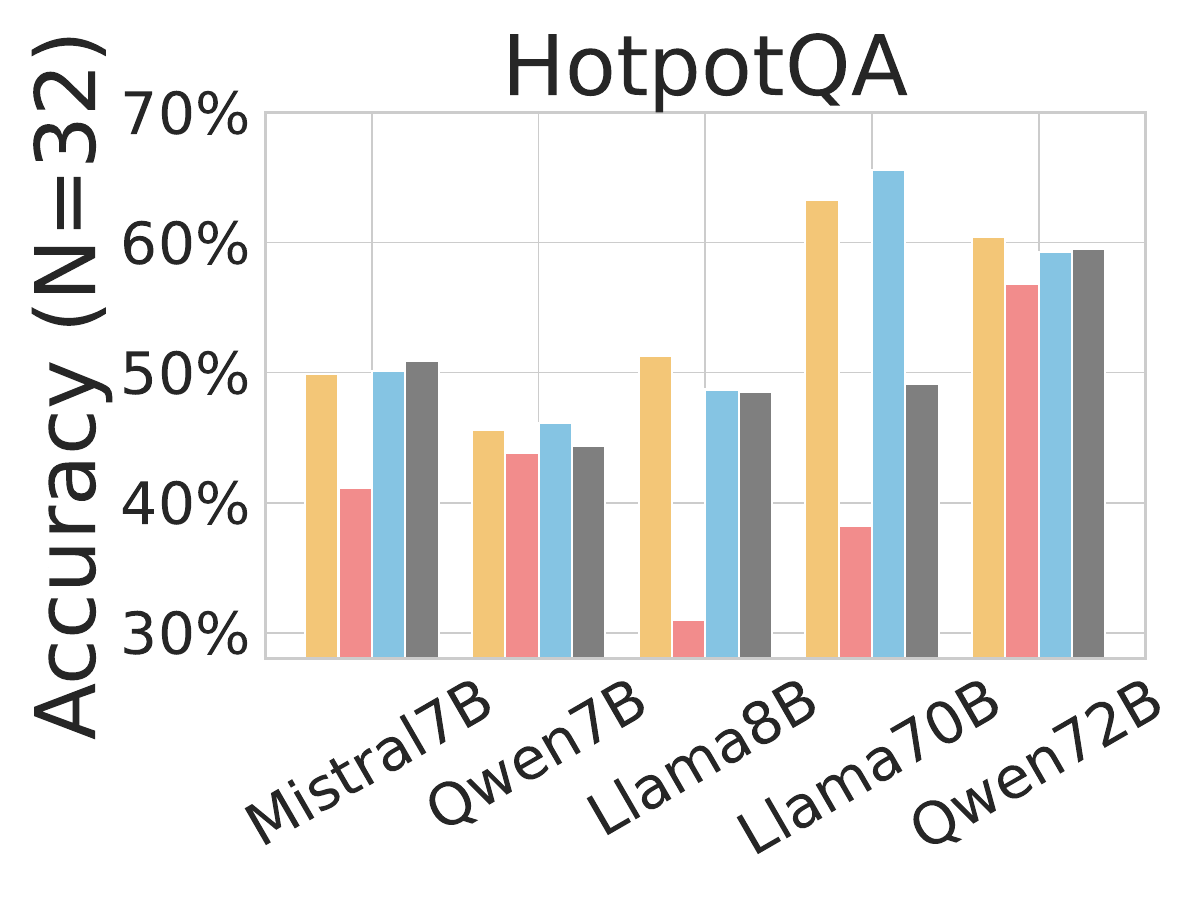}
}
\subfigure{
\includegraphics[width=0.225\textwidth]{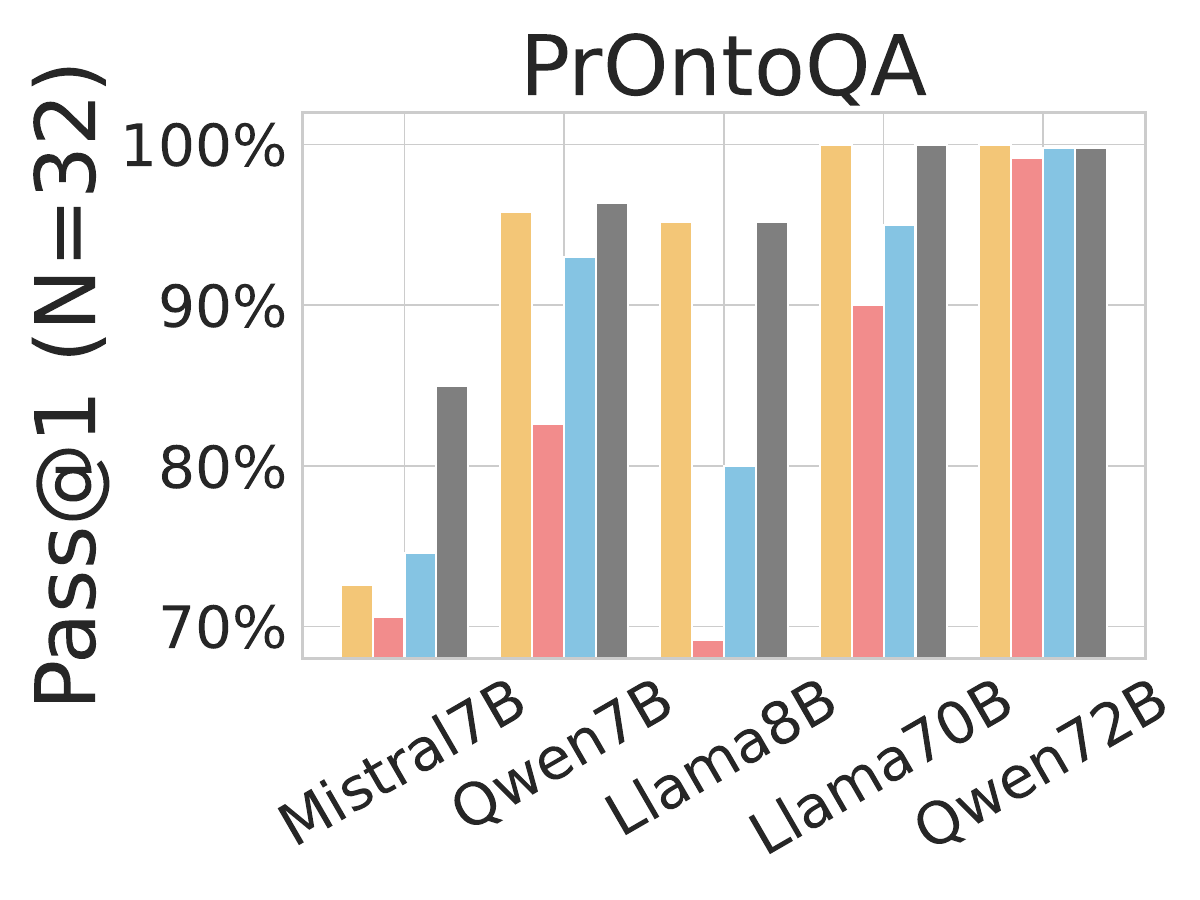}
}
\caption{Comparison of different reward models across benchmarks showcasing their impact on accuracy.}~\label{fig:trick_evalaution_type_other_data}
\vspace{-0.1in}
\end{figure}

\textbf{Reward Type.} Figure~\ref{fig:trick_evalaution_type_other_data} further reports the impact of reward types on other datasets, confirming the same findings. Specifically, for knowledge-based reasoning, certain reward models can substantially improve performance. In contrast, for more complex reasoning tasks, such as PrOntoQA, the LLM-as-Judge process reward provides a more significant performance boost.

\begin{figure}[t]
\vspace{-0.5in}
\centering
\includegraphics[width=0.6\textwidth]{figure/abalation/figure/overoptimize/legend.pdf}\\
\subfigure{
\includegraphics[width=0.225\textwidth]{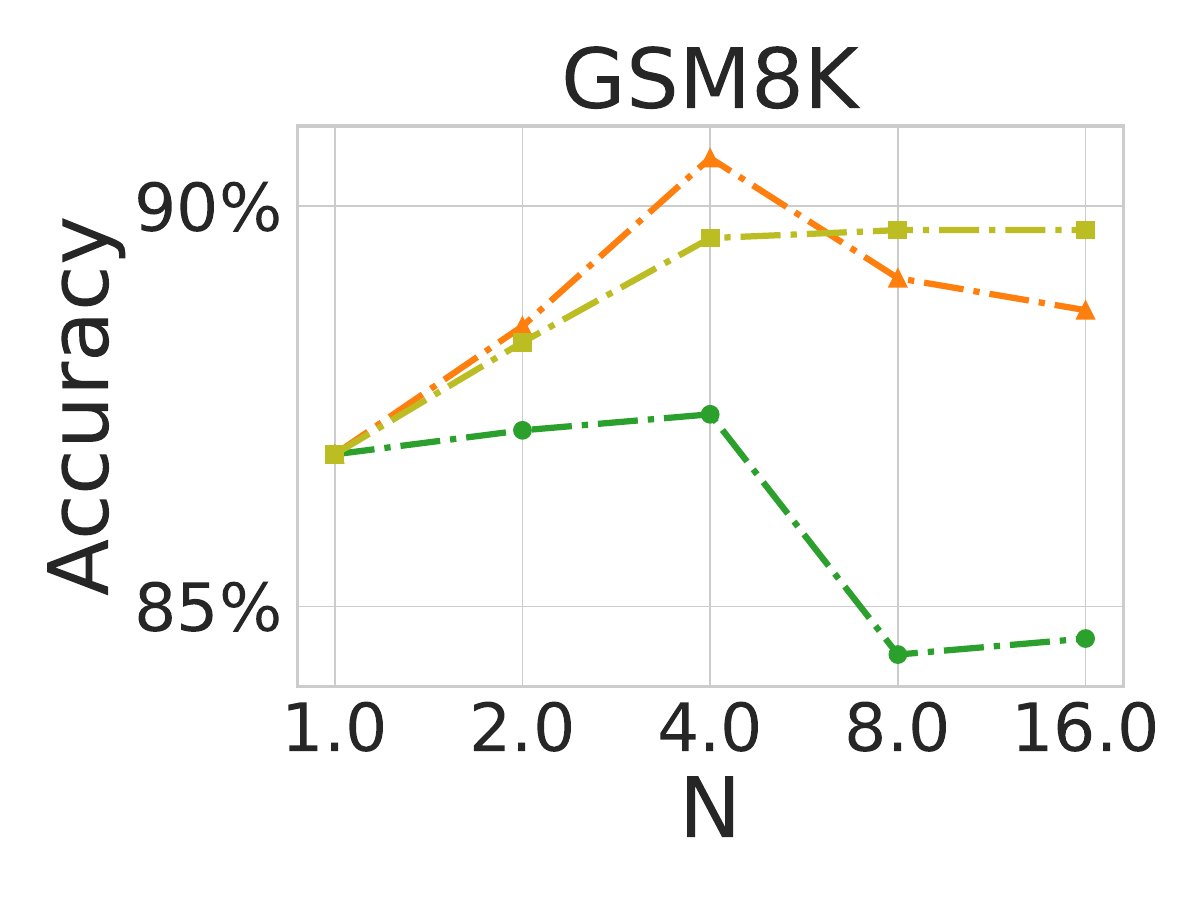}
}
\subfigure{
\includegraphics[width=0.225\textwidth]{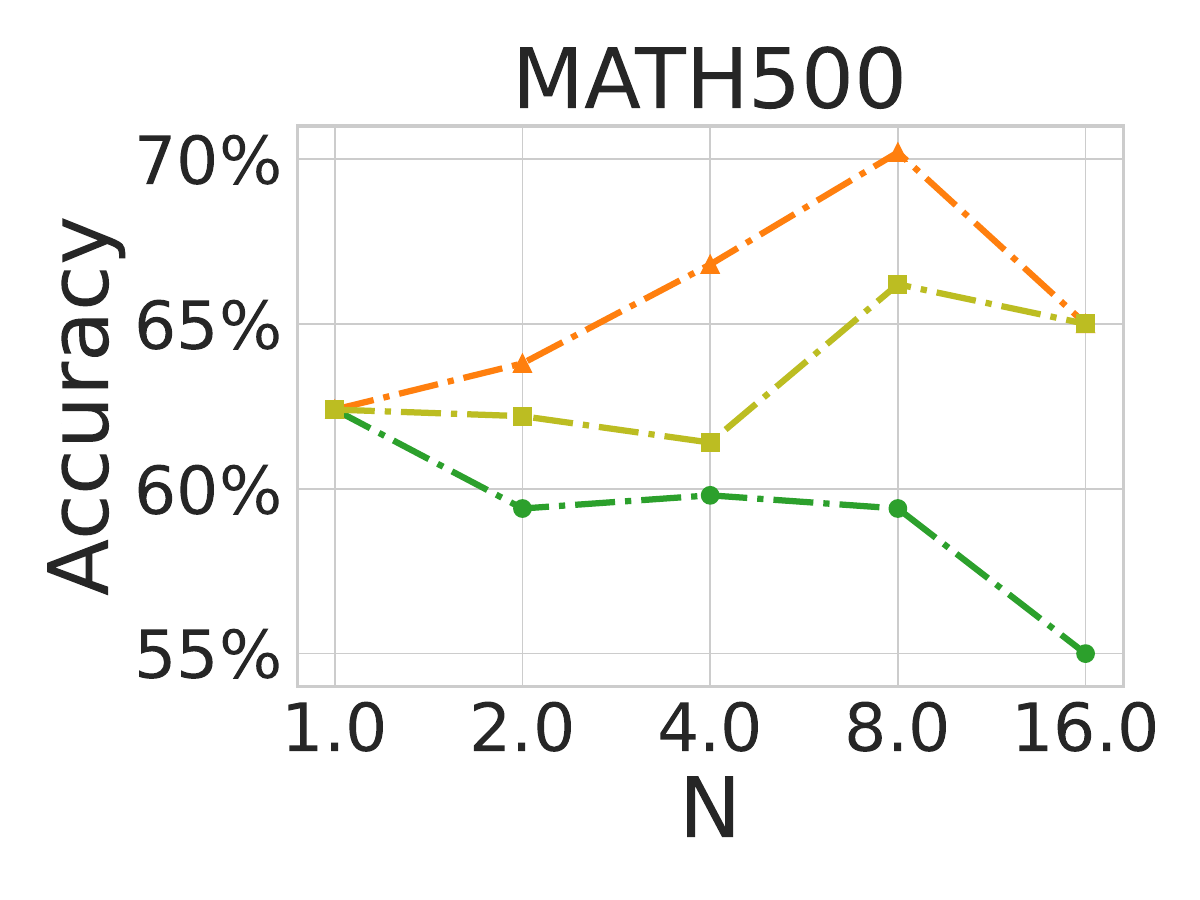}
}
\subfigure{
\includegraphics[width=0.225\textwidth]{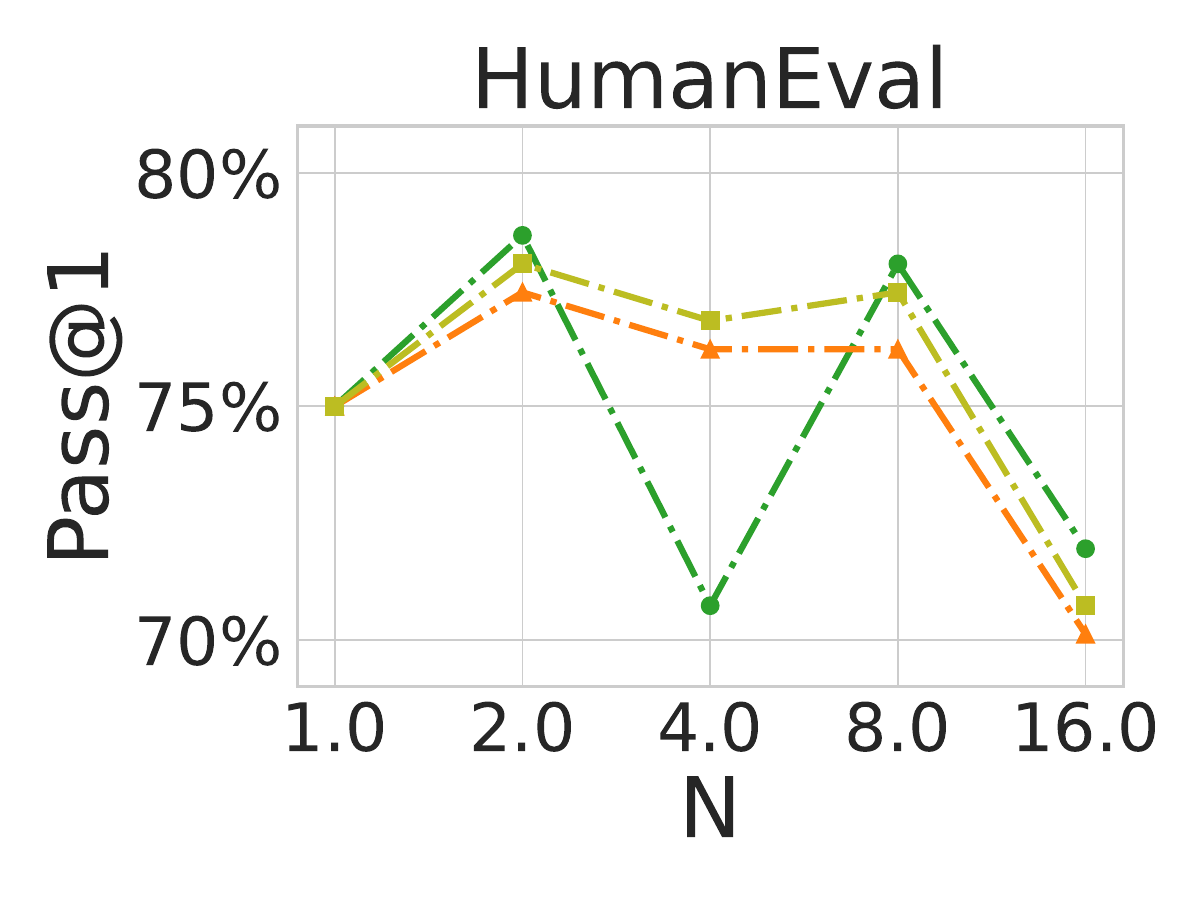}
}
\subfigure{
\includegraphics[width=0.225\textwidth]{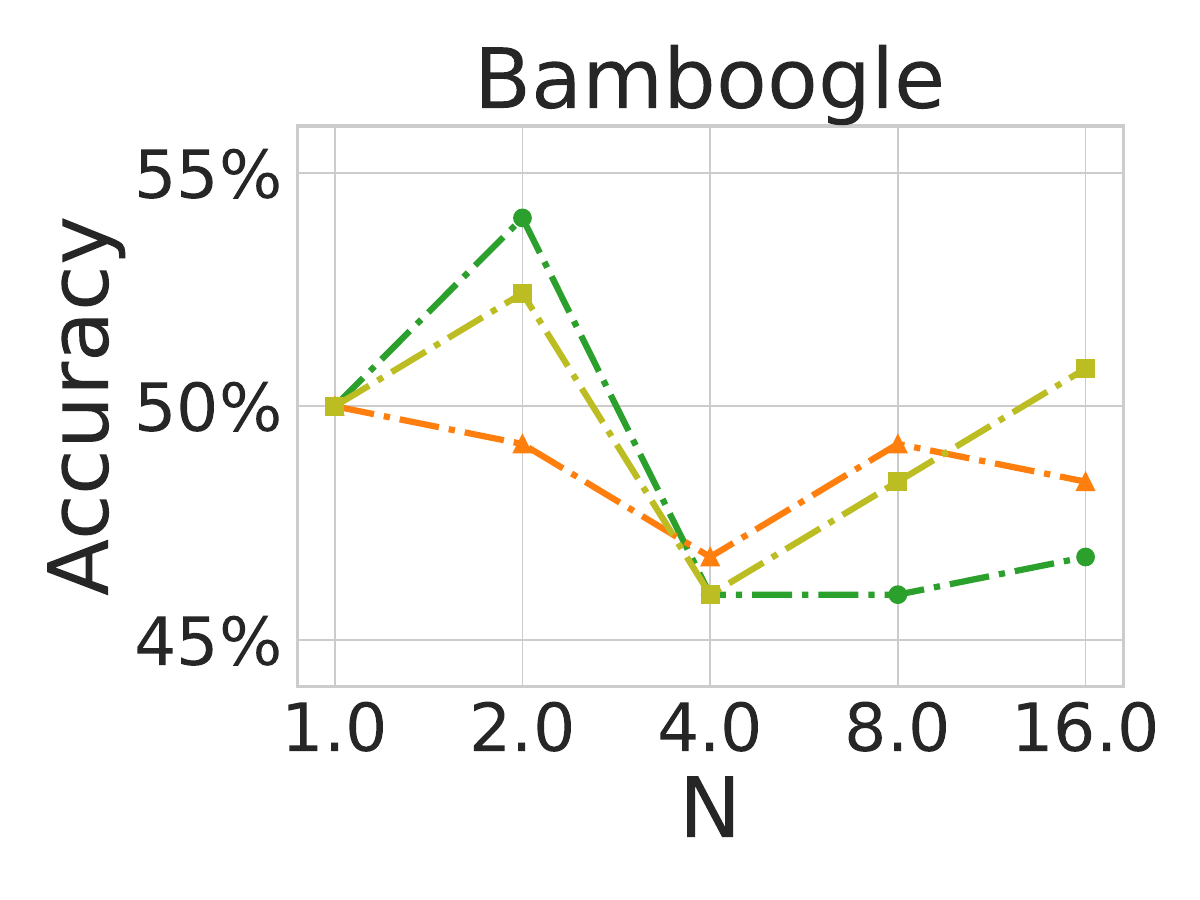}
}
\caption{Scaling test-time performance with reward models (Proof-Critical, RLHF, LLM-as-Judge) across benchmarks, highlighting inconsistent trends and generalization issues, with notable declines in challenging tasks like MATH500 On Qwen-2.5-7B.
}
\vspace{-0.3in}
\label{fig:over-optimization_qwen}
\end{figure}

\textbf{Performance Inflation with Reward Model.} Scaling test-time computation with reward models presents complexities. Figure~\ref{fig:over-optimization_qwen} reports that the reward model does not perform consistently across all cases during scaling on other datasets. For example, in the challenging MATH task, the proof-critical reward model can lead to a decrease in performance rather than a progressive improvement. This inflation of LLM reasoning performance can be attributed to the generalization issues of the reward model, as reported in~\cite{processbench, prmlessons}. Currently, the reward model does not perform well across all tasks.

\textbf{Combination of Tricks.}
Table~\ref{tab:tricks_com_qwen} further investigates the combination of selected useful techniques, including prompt type, temperature, top-p, and the reward model on Qwen2.5-7B. As demonstrated in Table~\ref{tab:tricks_com_qwen}, the improvements are not always additive when combining different techniques, although methods such as prompt type, temperature, and reward models can be individually effective. For example, as shown in Tables~\ref{tab:tricks_com_qwen} using Reflect CoT with a temperature of 0.8 and a top-p of 0.9 does not lead to improvements for the Bamboogle and MATH tasks. These results suggest that the careful selection and combination of techniques can enhance performance, though the impact may vary across different models and tasks.
\begin{table}[htb]  
    \centering  
    \small
    \caption{Combination of Inference-Time Tricks on Qwen2.5-7B. We highlight optimal trick combinations. }  
    ~\label{tab:tricks_com_qwen}  
    \begin{center}  
    \resizebox{0.48\textwidth}{!}{
        \begin{tabular}{c|c|c|c|c|c}  
            \toprule
            \textbf{Prompt} & \textbf{Reward} & \textbf{Temp} & \textbf{Top-p} & \textbf{Bamboogle} & \textbf{MATH} \\
            \midrule
            IO & Majority & 0.7 & 0.9 & 42.7 & 23.8 \\
            CoT & Majority & 0.7 & 0.9 & \textbf{53.2} & 72.4 \\
            CoT & Majority & 0.8 & 0.9 & 53.2 & 73.8\\
            CoT & Majority & 0.6 & 0.9 & 51.6 & 74.0\\
            CoT & Majority & 0.7 & 1.0 & 52.4 & 73.4 \\
            CoT & Majority & 0.7 & 0.8 & 50.8 & 73.8\\
            CoT & Random & 0.7 & 0.9 & 45.9 & 63.2 \\
            CoT & Self-Process-Evaluation & 0.7 & 0.9 & 45.2 & 62.8 \\
            CoT & Self-Result-Evaluation & 0.7 & 0.9 & 46.7 & 66.4 \\
            Reflect CoT & Majority & 0.7 & 0.9 & 49.2 & \textbf{74.6} \\
            Reflect CoT & Majority & 0.8 & 0.9 & 49.2 & 69.2 \\
            \rowcolor{blue!20}  
           Reflect CoT & Majority & 0.8 & 0.9 & 49.2 & 69.2 \\
            \bottomrule
        \end{tabular} 
    }
    \end{center} 
\vspace{-0.1in}
\end{table}

\begin{figure}[t]
\vspace{-0.5in}
\centering
\includegraphics[width=0.6\textwidth]{figure/tokentoacc/legend.pdf}\\
\subfigure{
\includegraphics[width=0.225\textwidth]{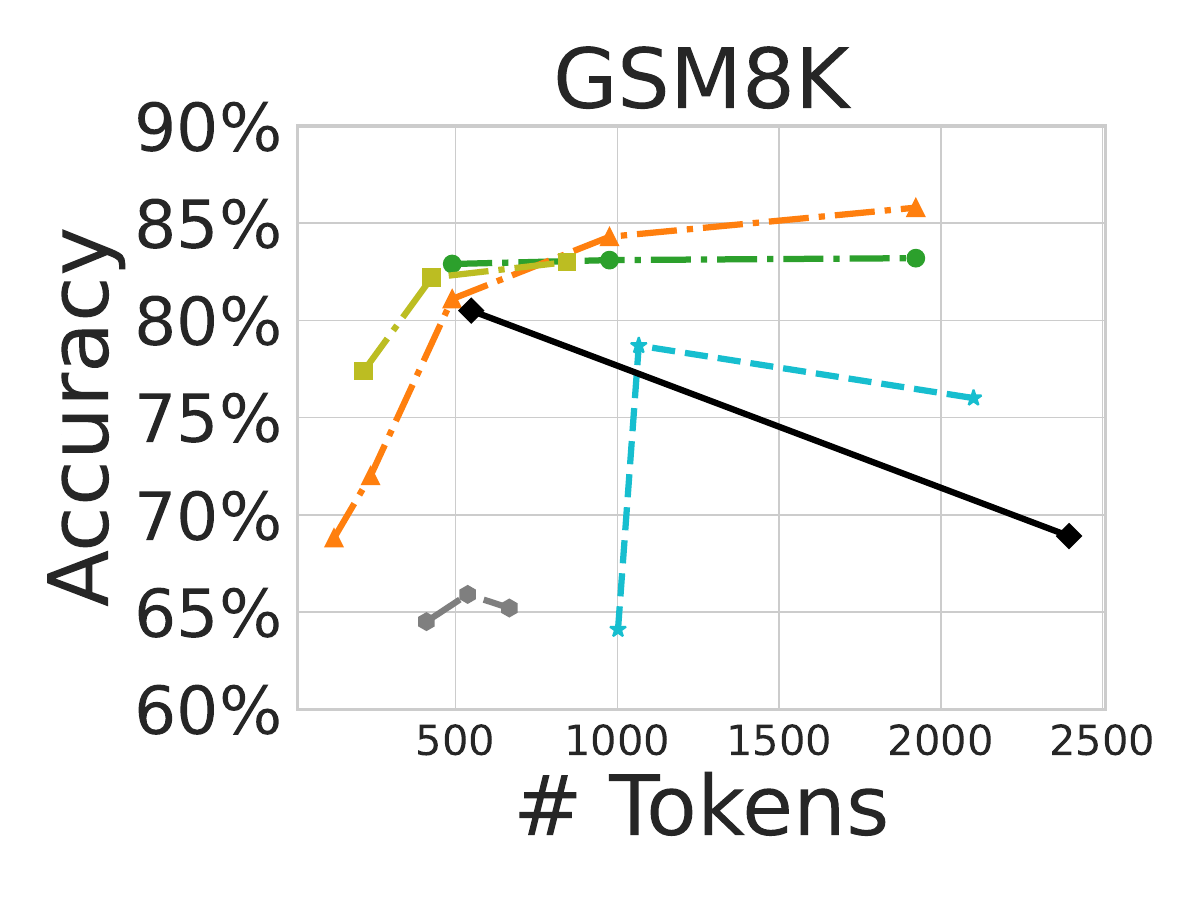}
}
\subfigure{
\includegraphics[width=0.225\textwidth]{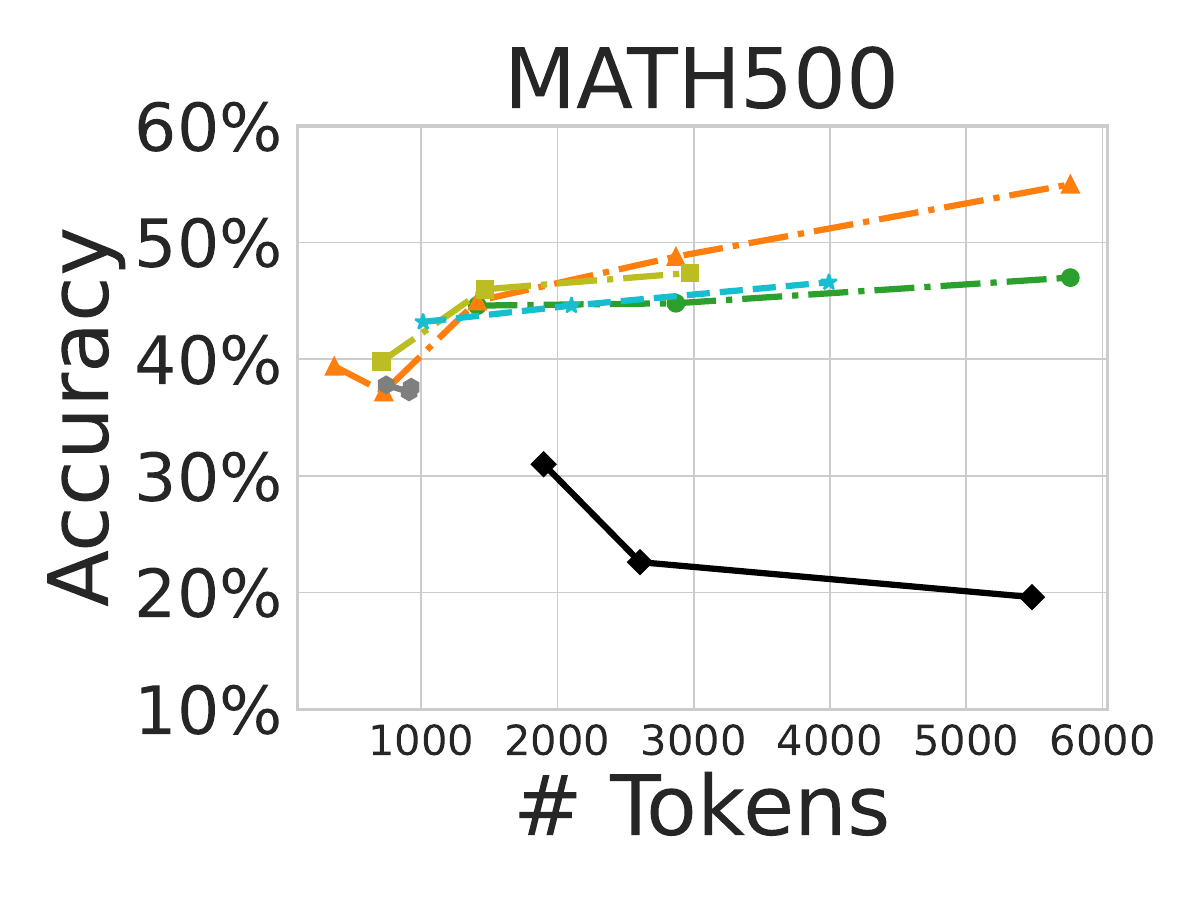}
}
\subfigure{
\includegraphics[width=0.225\textwidth]{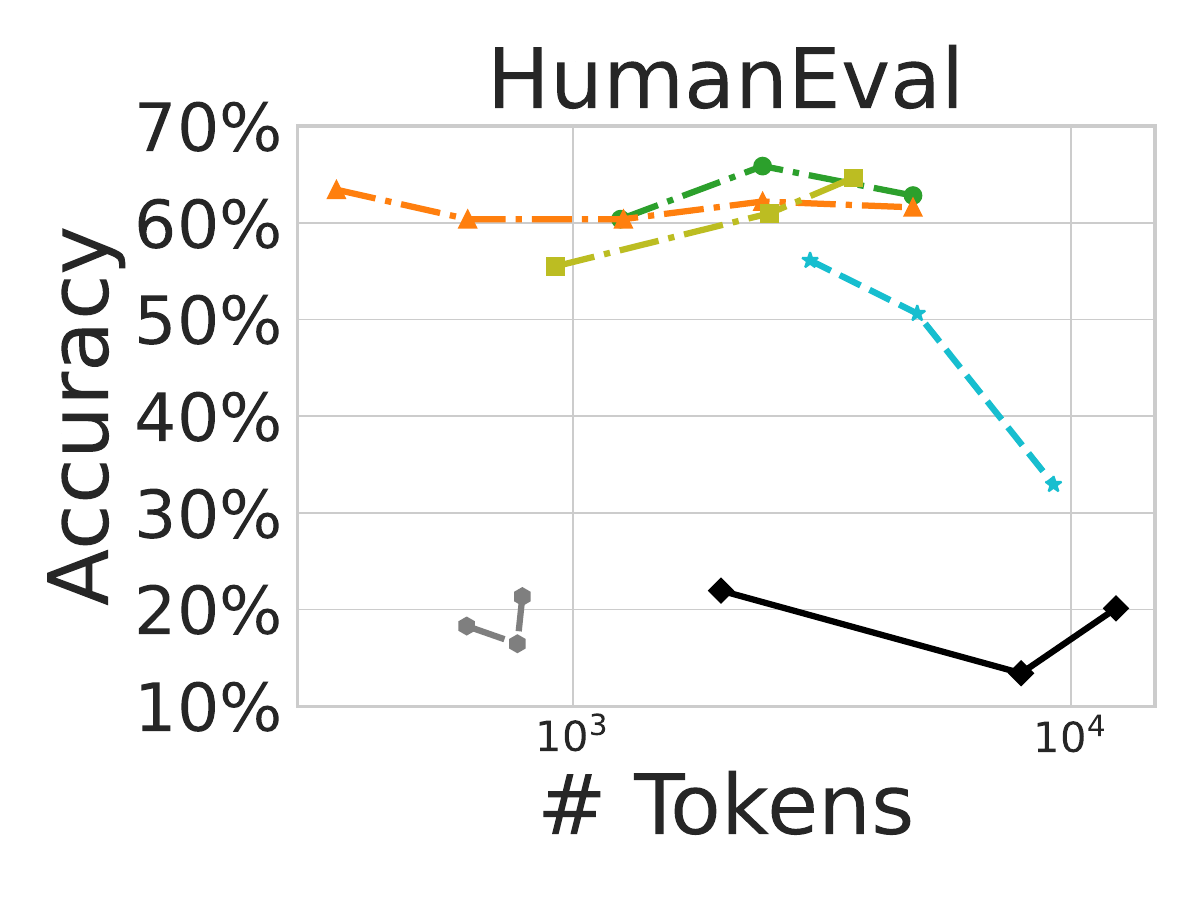}
}
\subfigure{
\includegraphics[width=0.225\textwidth]{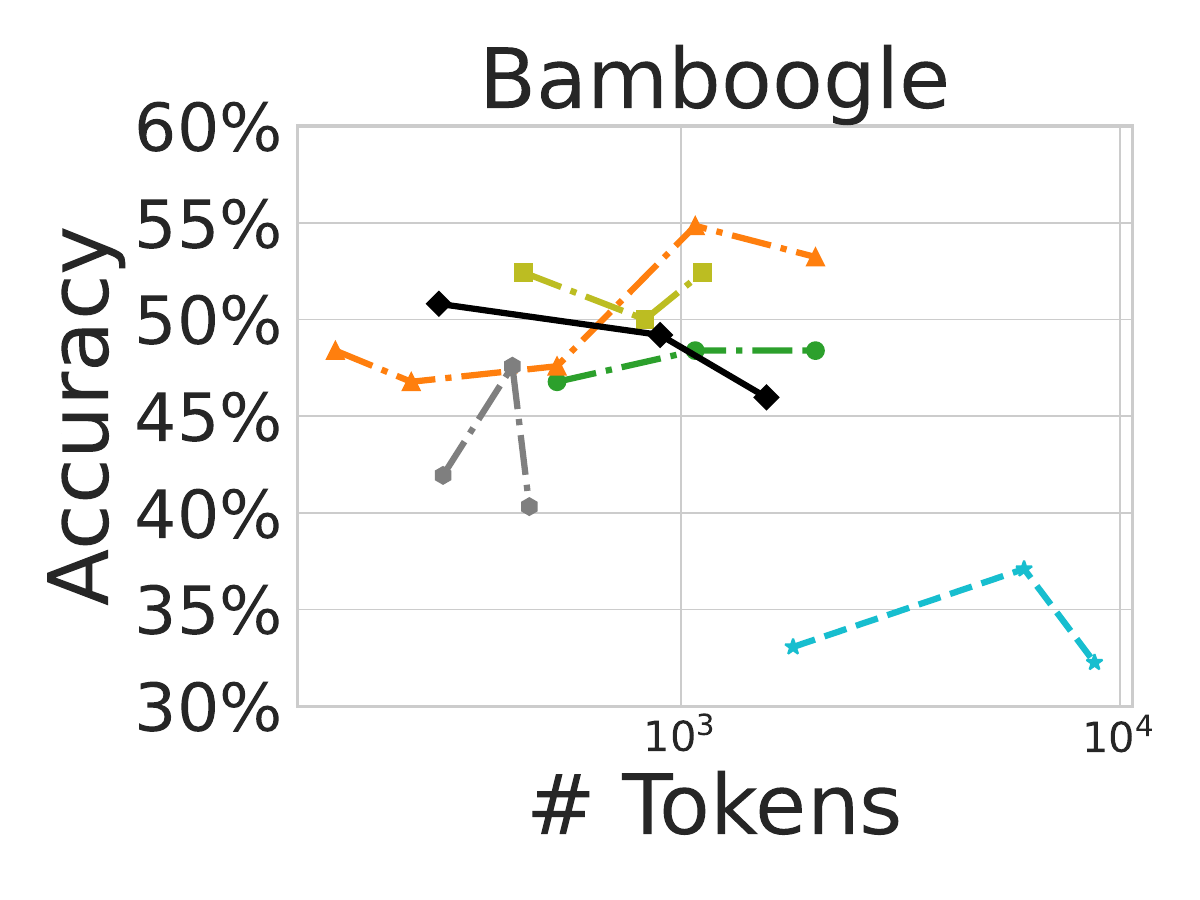}
}
\caption{ Performance versus token consumption across benchmarks  for decoding strategies (Self-Consistency, Self-Refine, Best-of-N, Greedy, Beam Search, MCTS), highlighting task-specific trends and token efficiency disparities, with GSM8K showing sharp accuracy gains and MCTS lagging in improvement
}~\label{fig:token_vary_llama}
\vspace{-0.4in}
\end{figure}

\textbf{Main experiments of Inference-time computation methods.}
Figure~\ref{fig:token_vary_llama} illustrates how performance varies with increased computational consumption. As token usage rises, Self-Consistency and Self-Refine achieve higher accuracy than other methods, while Step-level Best-of-N, and MCTS show slower improvements. In contrast, Beam Search exhibits minimal gains, indicating lower token efficiency. Notably, the GSM8K task demonstrates a sharper accuracy increase with higher token consumption compared to MATH500, underscoring the task-specific nature of each method's performance. Additionally, we observe some performance inflation with increased computational consumption, attributed to the generalization limitations of the reward model, which does not perform well on this task.

\end{document}